\documentclass[10pt,twocolumn,letterpaper]{article}

\usepackage{cvpr}              % To produce the CAMERA-READY version
\usepackage{times}
\usepackage{epsfig}
\usepackage{graphicx}
\usepackage{amsmath}
\usepackage{amssymb}
\usepackage{booktabs}
\usepackage{algorithm}
\usepackage{algorithmic}
\usepackage{multirow}
\usepackage{amsthm}
\usepackage{url}
\usepackage{rotate}
\usepackage{bm}
\usepackage{setspace}
\usepackage{color}
\usepackage{enumerate}
\usepackage{cases}

\usepackage{cuted}
\usepackage{capt-of}
\usepackage[pagebackref,breaklinks,colorlinks]{hyperref}

\usepackage[capitalize]{cleveref}
\crefname{section}{Sec.}{Secs.}
\Crefname{section}{Section}{Sections}
\Crefname{table}{Table}{Tables}
\crefname{table}{Tab.}{Tabs.}

\def\cvprPaperID{4550} % *** Enter the CVPR Paper ID here
\def\confName{CVPR}
\def\confYear{2022}

\begin{document}

%\title{S3C: Self-Supervised Sharing-Compensation Learning for Brightening Images}
%\title{Rain Removal in Video from Synthetic to Real: Large-Scale Datasets, General Framework, and Learning Strategy}
\title{Toward Fast, Flexible, and Robust Low-Light Image Enhancement}
\author{Long Ma$^\dag$, Tengyu Ma$^\dag$, Risheng Liu$^{\ddag}$\thanks{Corresponding author.},  Xin Fan$^{\ddag}$, Zhongxuan Luo$^{\dag}$\\	
	% To save space, use either the email address or home page, not both \normalsize
	\normalsize$^\dag$School of Software Technology, Dalian University of Technology\\
	\normalsize $^\ddag$International School of Information Science \& Engineering, Dalian University of Technology\\
	{\tt \small \{rsliu, xin.fan, zxluo\}@dlut.edu.cn, malone94319@gmail.com, matengyu@mail.dlut.edu.cn}\
}

\maketitle
\thispagestyle{empty}
\begin{strip}
	%\begin{figure*}[t]
	\vspace{-1.6cm}
	\centering
	\begin{tabular}{c@{\extracolsep{0.75em}}c@{\extracolsep{0.35em}}c} 
		\includegraphics[width=0.435\textwidth]{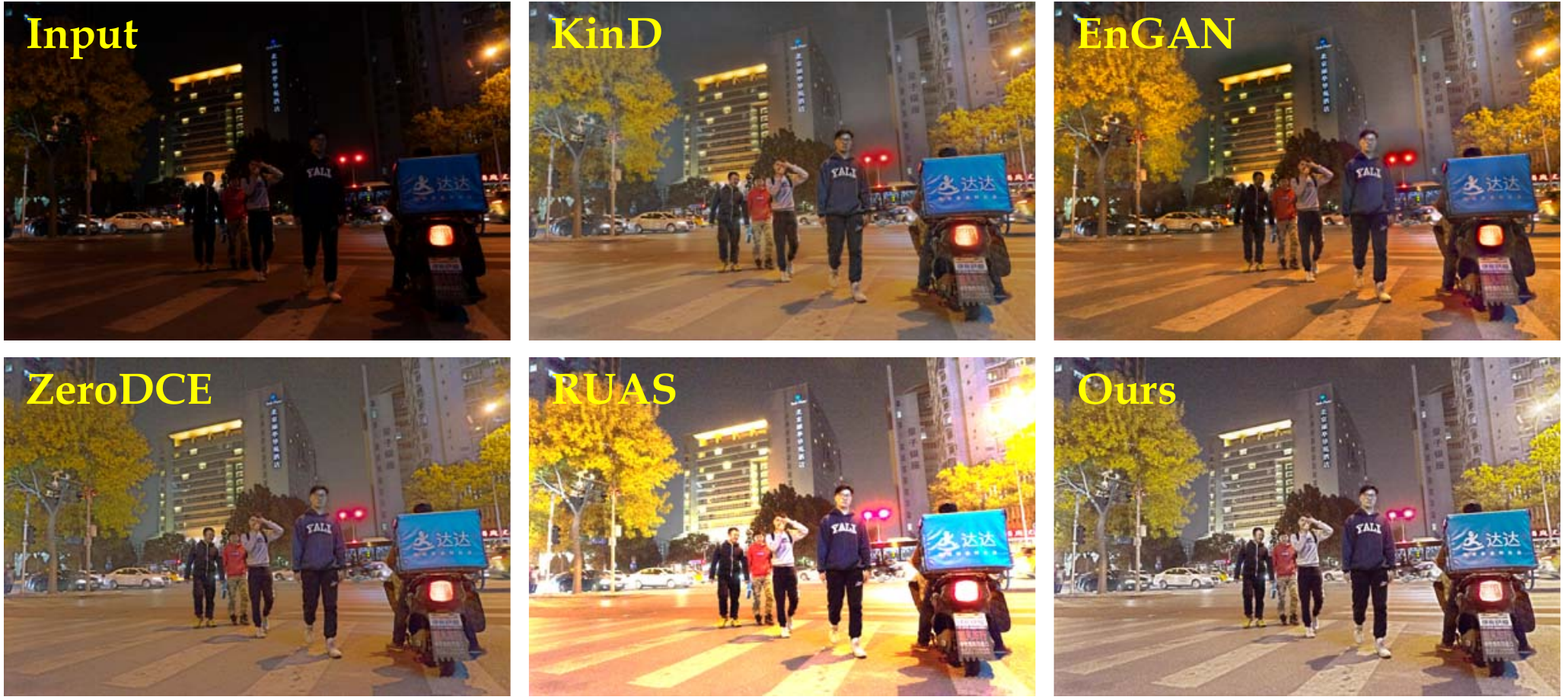}&
		\includegraphics[width=0.292\textwidth]{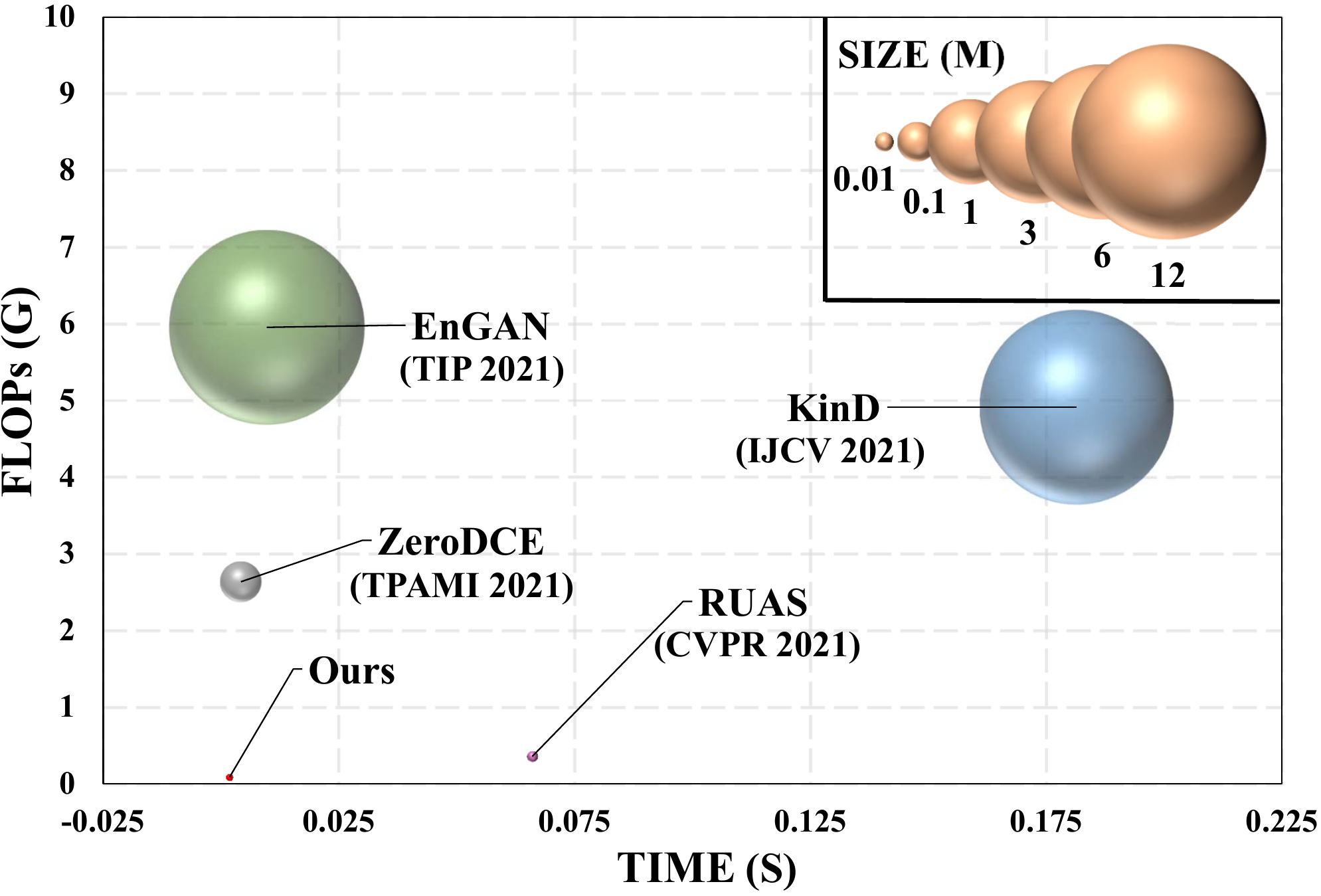}&
		\includegraphics[width=0.231\textwidth]{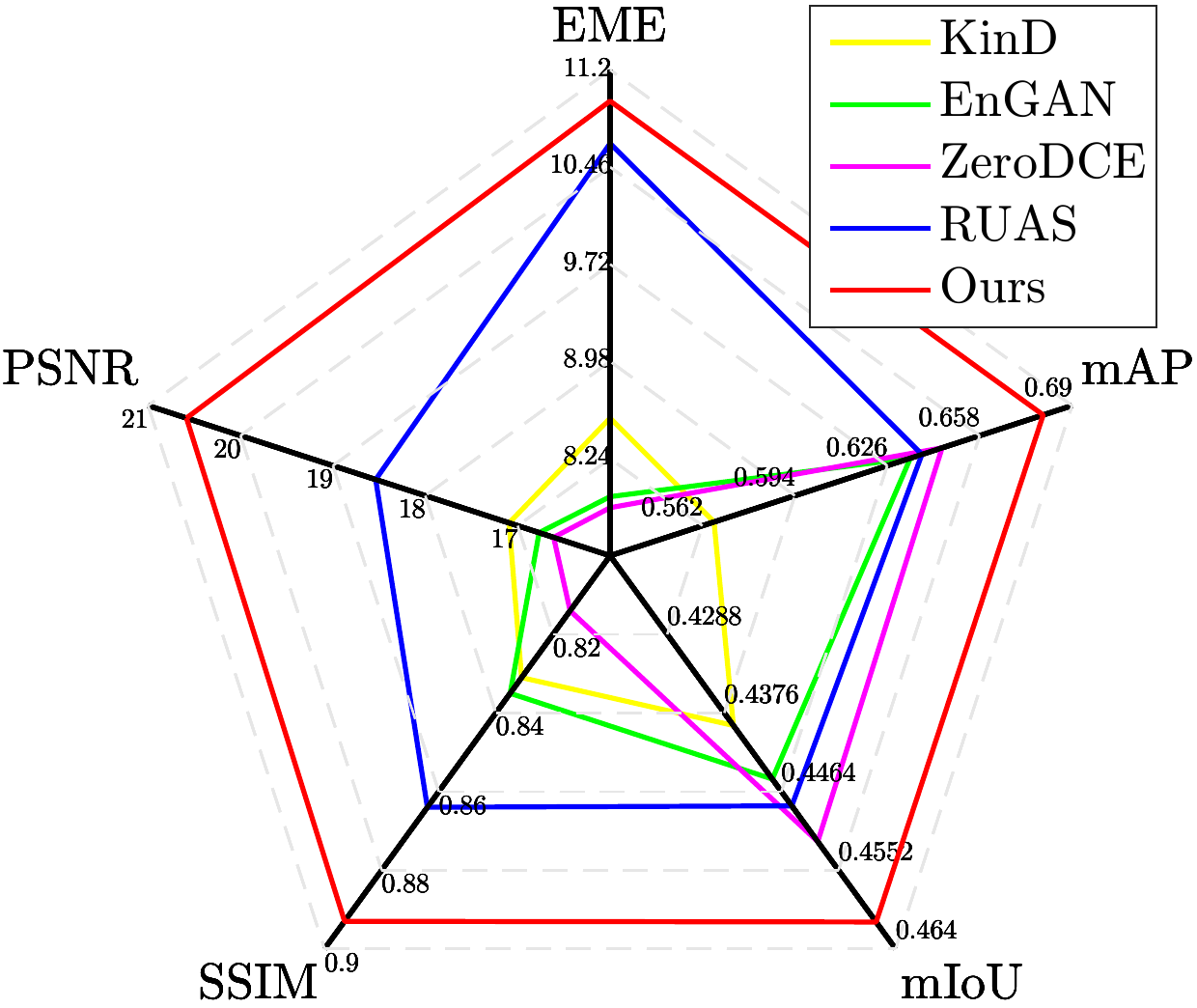}\\
		\footnotesize (a) Visual Quality&	\footnotesize (b) Computational Efficiency&	\footnotesize (c) Numerical Scores\\
	\end{tabular}
%	\vspace{-0.3cm}
	\captionof{figure}{Comparison among recent state-of-the-art methods and our method. KinD~\cite{zhang2019kindling} is a representative paired supervised method. EnGAN~\cite{jiang2019EnGAN} considers the unpaired supervised learning. ZeroDCE~\cite{guo2020zero} and RUAS~\cite{liu2021retinex} introduce unsupervised learning. Our method (just contains three convolutions with the size of $3\times 3$) also belongs to unsupervised learning. As shown in the zoomed-in regions, these compared methods appear incorrect exposure, color distortion, and insufficient structure to degrade visual quality. In contrast, our result presents a vivid color and sharp outline. Further, we report the computational efficiency (SIZE, FLOPs, and TIME) in (b) and numerical scores for five types of measurement metrics among three tasks including enhancement (PSNR, SSIM, and EME), detection (mAP), and segmentation (mIoU) in (c), it can be easily observed that our method is remarkably superior to others.}
	\label{fig:FirstFigure}
%	\vspace{-0.2cm}
	%	\end{figure*}
\end{strip}

%%%%%%%%% ABSTRACT
\begin{abstract}
Existing low-light image enhancement techniques are mostly not only difficult to deal with both visual quality and computational efficiency but also commonly invalid in unknown complex scenarios. 
In this paper, we develop a new Self-Calibrated Illumination (SCI) learning framework for fast, flexible, and robust brightening images in real-world low-light scenarios. To be specific, we establish a cascaded illumination learning process with weight sharing to handle this task. Considering the computational burden of the cascaded pattern, we construct the self-calibrated module which realizes the convergence between results of each stage, producing the gains that only use the single basic block for inference (yet has not been exploited in previous works), which drastically diminishes computation cost. We then define the unsupervised training loss to elevate the model capability that can adapt general scenes. 
Further, we make comprehensive explorations to excavate SCI's inherent properties (lacking in existing works) including operation-insensitive adaptability (acquiring stable performance under the settings of different simple operations) and model-irrelevant generality (can be applied to illumination-based existing works to improve performance). Finally, plenty of experiments and ablation studies fully indicate our superiority in both quality and efficiency. Applications on low-light face detection and nighttime semantic segmentation fully reveal the latent practical values for SCI. The source code is available at \url{https://github.com/vis-opt-group/SCI}.

\end{abstract}

%%%%%%%%% BODY TEXT

\section{Introduction}

Low-light image enhancement aims at making information hidden in the dark visible to improve image quality, it has drawn much attention in multiple emerging computer vision areas~\cite{wang2021hla,wu2021dannet,sakaridis2019guided} recently. 
%Although many advanced techniques have been developed in the past few years, how to construct an efficient and effective model for complex and diverse scenarios is still in suspense. 
In the following, we sort out the development process of two related topics. Further, we describe our main contributions. 

%\subsection{Related Works}
\textbf{Model-based Methods.} Generally, Retinex theory~\cite{rahman2004retinex} depicts the basic physical law for low-light image enhancement, that is, low-light observation can be decomposed into illumination and reflectance (i.e., clear image). Benefiting from the convenient solution of $\ell_2$-norm, Fu~\emph{et al.}~\cite{fu2015probabilistic,fu2016weighted} firstly utilized the $\ell_2$-norm to constrain the illumination. Further, Guo~\emph{et al.}~\cite{guo2017lime} adopted the relative total variation~\cite{tsmoothing2012} as the constraint of the illumination. However, its fatal defect exists in the overexposure appearance. Li~\emph{et al.}~\cite{li2018structure} modeled the noise removal and low-light enhancement in a unified optimization goal. The work in~\cite{hao2020low} proposed a semi-decoupled decomposition model for simultaneously improving the brightness and suppressing noises. Some works (e.g., LEACRM~\cite{ren2018lecarm}) also utilized the response characteristics of cameras for enhancement. 
Limited to the defined regularizations, they mostly generate unsatisfying results and need to manually adjust lots of parameters towards real-world scenarios. 

\textbf{Network-based Methods.} By adjusting the exposure time, the work in~\cite{Chen2018Retinex} built a new dataset, called LOL dataset. This work also designed the RetinexNet which tended to produce unnatural enhanced results. KinD~\cite{zhang2019kindling} ameliorated issues that appeared in RetinexNet by introducing some training losses and tuned up the network architecture. 
DeepUPE~\cite{wang2019underexposed} defined an illumination estimation network for enhancing the low-light inputs. The work in~\cite{yang2020fidelity} proposed a recursive band network and trained it by a semi-supervised strategy.
EnGAN~\cite{jiang2019EnGAN} designed a generator with attention for enhancement under the unpaired supervision.  
SSIENet~\cite{zhang2020self} built a decomposition-type architecture to simultaneously estimate the illumination and reflectance. ZeroDCE~\cite{guo2020zero} heuristically built a quadratic curve with learned parameters. Very recently, Liu~\emph{et al.}~\cite{liu2021retinex} built a Retinex-inspired unrolling framework with architecture search. Undeniably, these deep networks are well-designed. However, they are not stable, and hard to realize consistently superior performance, especially in unknown real-world scenarios, unclear details and inappropriate exposure are ubiquitous.

\textbf{Our Contributions.} To settle the above issues, we develop a novel Self-Calibrated Illumination (SCI) learning framework for fast, flexible and robust low-light image enhancement. By redeveloping the intermediate output of the illumination learning process, we construct a self-calibrated module to endow the stronger representation to the single basic block and convergence between results of each stage to realize acceleration. More concretely, our main contributions can be concluded as:
\begin{itemize}
	\item We develop a self-calibrated module for the illumination learning with weight sharing to confer the convergence between results of each stage, improving the exposure stability and reduce the computational burden by a wide margin. To the best of our knowledge, it is the first work to accelerate the low-light image enhancement algorithm by exploiting learning process.  
	\item We define the unsupervised training loss to constrain the output of each stage under the effects of self-calibrated module, endowing the adaptation ability towards diverse scenes. The attribute analysis shows that SCI possesses the operation-insensitive adaptability and model-irrelevant generality, which have not been found in existing works. 
	\item Extensive experiments are conducted to illustrate our superiority against other state-of-the-art methods. Applications on dark face detection and nighttime semantic segmentation are further performed to reveal our practical values. In nutshell, SCI redefines the peak-point in visual quality, computational efficiency, and performance on downstream tasks in the field of network-based low-light image enhancement. 
\end{itemize}

%\section{Related Work}
%In this section, we comb the development process of two related topics most relevant to this work including low-light image enhancement and unfolding-type techniques.

\section{The Proposed Method}
In this section, we firstly introduce the illumination learning with weight sharing, then we build the self-calibrated module. Next the unsupervised training loss is presented. Finally, we make a comprehensive discussion about our constructed SCI.   

\subsection{Illumination Learning with Weight Sharing}~\label{sec:illumination}
According to the Retinex theory, there is a connection existing between the low-light observation $\mathbf{y}$ and the desired clear image $\mathbf{z}$: $\mathbf{y}=\mathbf{z}\otimes\mathbf{x}$, where $\mathbf{x}$ represents the illumination component. Generally, illumination is viewed as the core component that needs to be mainly optimized for low-light image enhancement. The enhanced output can be further acquired by removing the estimated illumination according to the Retinex theory. Here, inspired by the stage-wise optimization process for illumination presented in the works~\cite{guo2017lime,liu2021retinex}, by introducing a mapping $\mathcal{H}_{\bm{\theta}}$ with parameters $\bm{\theta}$ to learn the illumination, we provide a progressive perspective to model this task, the basic unit is written as
\begin{equation}\label{eq:UL}
\mathcal{F}(\mathbf{x}^{t}):\left\{
\begin{aligned}
&\mathbf{u}^{t} = \mathcal{H}_{\bm{\theta}}\big(\mathbf{x}^{t}\big), \mathbf{x}^{0}=\mathbf{y},\\
&\mathbf{x}^{t+1}=\mathbf{x}^{t}+\mathbf{u}^{t},\\
\end{aligned}
\right.
\end{equation}
where $\mathbf{u}^{t}$ and $\mathbf{x}^{t}$ represent the residual term and illumination at $t$-th stage ($t=0, ..., T-1$), respectively. 
It should be noted that we do not mark the stage number in $\mathcal{H}_{\bm{\theta}}$ because we adopt the weight sharing mechanism, i.e., using the same architecture $\mathcal{H}$ and weights $\bm{\theta}$ in each stage. 

In fact, the parameterized operator $\mathcal{H}_{\bm{\theta}}$\footnote{The architecture for $\mathcal{H}_{\bm{\theta}}$ will be explored in Sec.~\ref{sec:flexibility}.} learns a simple residual representation $\mathbf{u}^{t}$ between the illumination and low-light observation. This process is inspired by a consensus, i.e., the illumination and low-light observation are similar or existing linear connections in most areas. 
Compared with adopting a direct mapping between the low-light observation and illumination (a commonly-used pattern in existing works, e.g.,~\cite{wang2019underexposed,liu2021retinex}), learning a residual representation substantially reduces the computational difficulty to both guarantee performance and improve the steadiness, especially for the exposure control\footnote{Please refer to the ablation study in Sec.~\ref{sec:ablation} to confirm it.}. 

Indeed, we can directly utilize the above-built process with the given training loss and data to acquire the enhanced model. But it is noticeable that the cascaded mechanism with multiple weight sharing blocks inevitably gives a rise to foreseeable inference cost.
\textit{Revisiting this sharing process, each shared block expects to output a result that is close to the desired goal as far as possible. Going a step further,, the ideal circumstance is that the first block can output the desired result, which satisfies task demands.  Meanwhile, the latter blocks output the similar, even the completely same results as the first block does. In this way, in testing phase, we just need a single block to accelerate the inference speed.} Next, we will explore how to realize it.

\begin{figure}[t]
	\centering
	\begin{tabular}{c} 
		\includegraphics[width=0.46\textwidth]{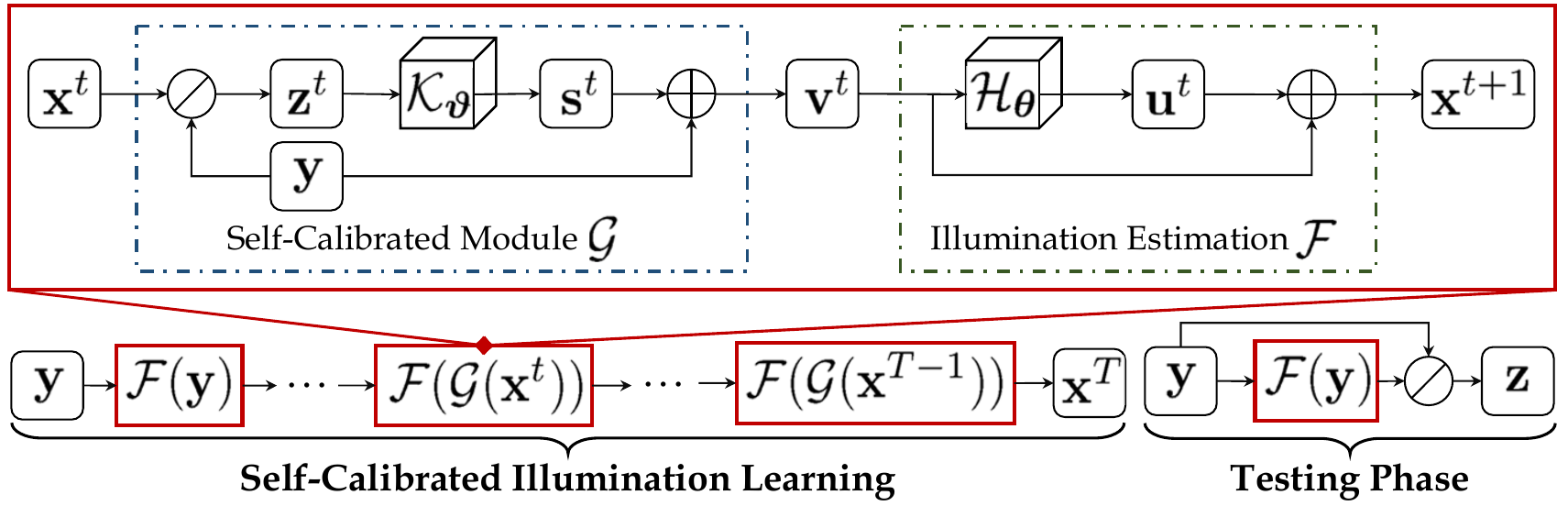}\\
	\end{tabular}
	\vspace{-0.2cm}
	\caption{The entire framework of SCI. In the training phase, our SCI is composed of the illumination estimation and self-calibrated module. The self-calibrated module map is added to the original low-light input as the input of the illumination estimation at the next stage. Note that these two modules are respectively shared parameters in the whole training procedure. In the testing phase, we just utilize a single illumination estimation module. }
	\label{fig:Flow}
	\vspace{-0.2cm}
\end{figure}

\subsection{Self-Calibrated Module}

\begin{figure}[t]
	\centering
	\begin{tabular}{c} 
		\includegraphics[width=0.45\textwidth]{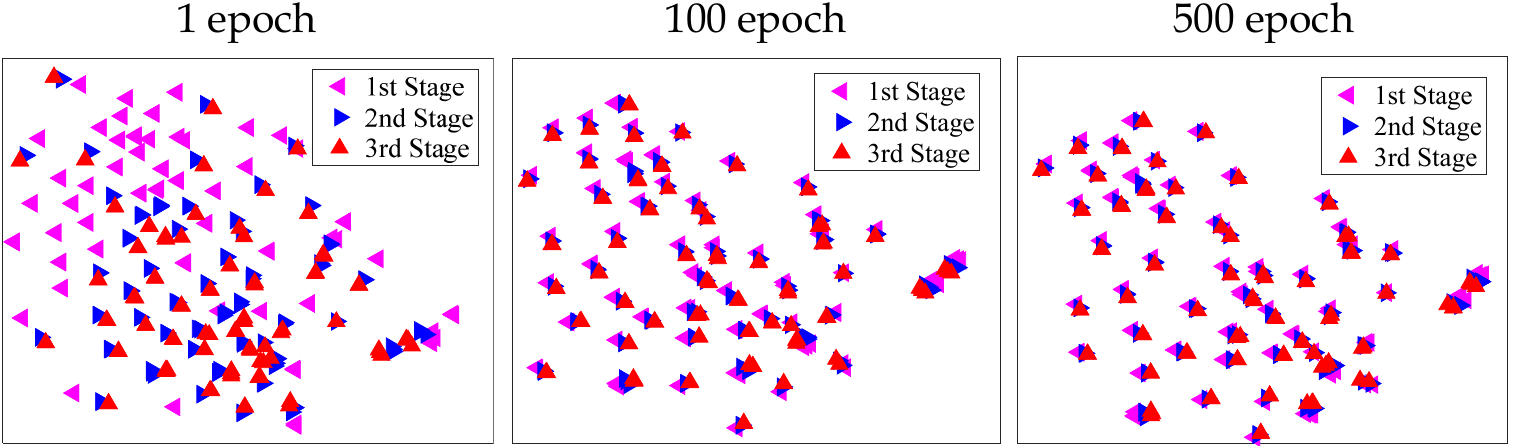}\\
		\footnotesize (a) w/o self-calibrated module\\
		\includegraphics[width=0.45\textwidth]{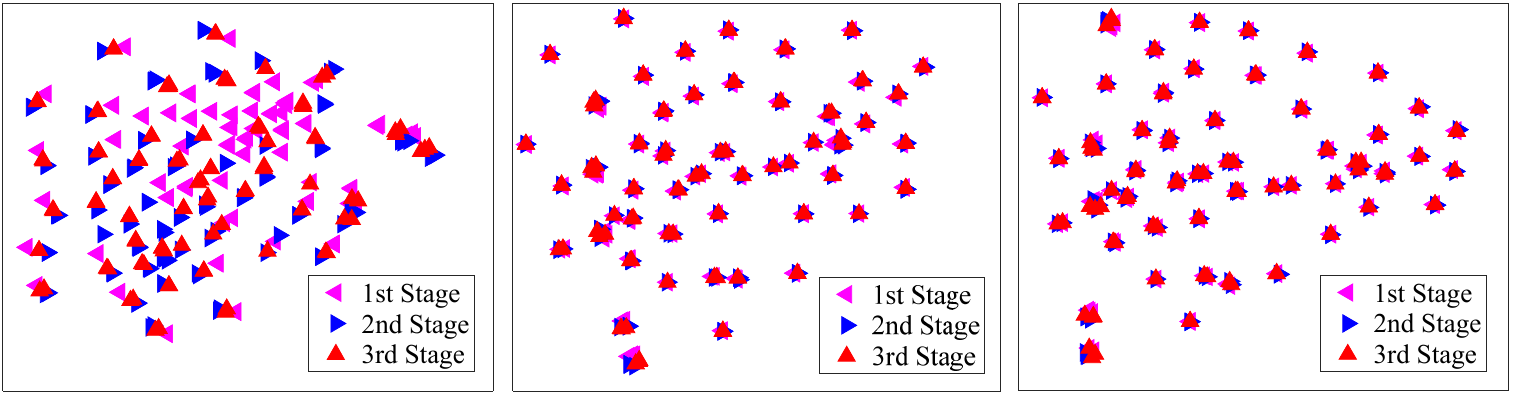}\\
		\footnotesize (b) w/ self-calibrated module\\
	\end{tabular}
	\vspace{-0.2cm}
	\caption{Comparing t-SNE~\cite{van2008visualizing} distributions in terms of the results of each stage on whether using self-calibrated module. It exhibits why we can use a single stage for testing, that is, the results of each stage in SCI can rapidly converge to the same value, but w/o self-calibrated module cannot realize it all the time. }
	\label{fig:tSNE}
	\vspace{-0.2cm}
\end{figure}

Here, we aim at defining a module to make results of each stage convergent to the same one state. 
We know that the input of each stage stems from the previous stage and the input of the first stage is definitely defined as the low-light observation. An intuitive idea is that whether we can bridge the input of each stage (except the first stage) and the low-light observation (i.e., the input of the first stage) to indirectly explore the convergence behavior between each stage. To this end, we introduce a self-calibrated map $\mathbf{s}$ and add it to the low-light observation to present the difference between the input in each stage and the first stage. Specifically, the self-calibrated module can be presented as
%\begin{subequations}\label{eq:SCI}
%	\begin{numcases}{}
%	\mathbf{s}^{t-1}=\mathcal{K}_{\bm{\vartheta}}(\mathbf{z}^{t-1}),\label{eq:SCI1}\\
%	\mathbf{v}^{t-1} = {\mathcal{G}_{\bm{\vartheta}}^{t-1}}(\mathbf{z}^{t-1})=\mathbf{y}+\mathbf{s}^{t-1}, \label{eq:SCI2}\\
%	\mathbf{u}^{t} = \mathcal{H}_{\bm{\theta}}\big(\mathbf{v}^{t-1}\big), \label{eq:SCI3}\\
%	\mathbf{x}^{t}={\mathcal{F}_{\bm{\theta}}^{t}}(\mathbf{v}^{t-1})
%	=\mathbf{v}^{t-1}+\mathbf{u}^{t},\label{eq:SCI4}\\
%	\mathbf{z}^{t} = \mathbf{y}\oslash\mathbf{x}^{t},\label{eq:SCI5}
%	\end{numcases}
%\end{subequations}
\begin{equation}\label{eq:SCI}
{\mathcal{G}}(\mathbf{x}^{t}):\left\{
	\begin{aligned}
	&\mathbf{z}^{t}= \mathbf{y}\oslash\mathbf{x}^{t},\\
	&\mathbf{s}^{t}=\mathcal{K}_{\bm{\vartheta}}(\mathbf{z}^{t}),\\
	&\mathbf{v}^{t}=\mathbf{y}+\mathbf{s}^{t},\\
	\end{aligned}
	\right.
\end{equation}
where $t\geq 1$, $\mathbf{v}^{t}$ is the converted input for each stage, and $\mathcal{K}_{\bm{\vartheta}}$\footnote{The architecture for $\mathcal{K}_{\bm{\vartheta}}$ will be explored in Supplemental Materials.} is the introduced parameterized operator with the learnable parameters $\bm{\vartheta}$. Then the conversion for the basic unit in $t$-th stage ($t\geq 1$) can be written as 
\begin{equation}
	\mathcal{F}(\mathbf{x}^t)\rightarrow\mathcal{F}(\mathcal{G}(\mathbf{x}^t)).
\end{equation}
Actually, our constructed self-calibrated module gradually corrects the input of each stage by integrating the physical principle to indirectly influence the output of each stage. To evaluate the effects of the self-calibrated module on the convergence, we plot tSNE distributions among results of each stage in Fig.~\ref{fig:tSNE}, and we can easily observe that the results of each stage indeed converge to the same value. But this phenomenon cannot be found in the case without the self-calibrated module. Additionally, the above conclusion also reflects that we indeed accomplish the intention as described in the last paragraph of Sec.~\ref{sec:illumination}, i.e., training multiple cascaded blocks with the weight sharing pattern but only using the single block for testing. 

We also provide the overall flowchart in Fig.~\ref{fig:Flow} for understanding our established SCI framework.

\subsection{Unsupervised Training Loss}\label{sec:ssl}
Considering the inaccuracy of existing paired data, we adopt the unsupervised learning to enlarge the network capability. We define the total loss as
$\mathcal{L}_{total}=\alpha\mathcal{L}_{f}+\beta\mathcal{L}_{s}$, where $\mathcal{L}_{f}$ and $\mathcal{L}_{s}$ represent the fidelity and smoothing loss, respectively. $\alpha$ and $\beta$ are two positive balancing parameters\footnote{Parameters analysis can be found in the Supplemental Materials.}. The fidelity loss is to guarantee the pixel-level consistency between the estimated illumination and the input of each stage, formulated as
\begin{equation}
	\mathcal{L}_{f}=\sum_{t=1}^{T}\|\mathbf{x}^{t}-(\mathbf{y}+\mathbf{s}^{t-1})\|^2,
\end{equation}
where $T$ is the total stage number. Actually, this function utilizes the redefined input $\mathbf{y}+\mathbf{s}^{t-1}$ to constrain the output illumination $\mathbf{x}^{t}$, rather than the hand-crafted ground truth or the plain low-light input. 

The smoothness property of the illumination is a broad consensus in this task~\cite{zhang2019kindling,guo2020zero}. Here we adopt a smoothness term with spatially-variant $\ell_1$ norm~\cite{fan2018image}, presented as
\begin{equation}
	\mathcal{L}_{s}= \sum_{i=1}^{N}\sum_{j\in\mathcal{N}(i)}w_{i,j}|\mathbf{x}^{t}_{i}-\mathbf{x}^{t}_{j}|,
\end{equation}
where $N$ is the total number of pixels. $i$ is the $i$-th pixel. $\mathcal{N}(i)$ denotes the adjacent pixels of $i$ in its $5\times5$ window. $w_{i,j}$ represents the weight, whose formulated form is  $w_{i,j}=\exp\Big(-\frac{\sum_{c}((\mathbf{y}_{i,c}+\mathbf{s}_{i,c}^{t-1})-(\mathbf{y}_{j,c}+\mathbf{s}_{j,c}^{t-1}))^2}{2\sigma^2}\Big)$, 
where $c$ denotes image channel in the YUV color space. $\sigma=0.1$ is the standard deviations for the Gaussian kernels.

\begin{table}[t]
	\renewcommand\arraystretch{1.1} 
	\setlength{\tabcolsep}{1.8mm}
	\centering
	\begin{tabular}{|cc|cc|cc|}
		\hline
		\multicolumn{2}{|c|}{\footnotesize  Setting for $\mathcal{H}_{\bm{\theta}}$}& \multicolumn{2}{c|}{\footnotesize Quality}& \multicolumn{2}{c|}{\footnotesize Efficiency}\\
		\hline
		\footnotesize Blocks&\footnotesize Channels&\footnotesize PSNR&\footnotesize NIQE&\footnotesize FLOPs (G)&\footnotesize TIME (S)\\
		\hline
		\footnotesize 1&\footnotesize 3-3&\footnotesize  20.6074&\footnotesize 4.0091&\footnotesize  0.0202&\footnotesize 0.0015\\
		\hline
		\footnotesize 2&\footnotesize 3-3-3&\footnotesize 20.5809&\footnotesize 4.0075&\footnotesize  0.0410&\footnotesize 0.0016\\
		\hline
		\footnotesize 3&\footnotesize 3-3-3-3&\footnotesize 20.4459&\footnotesize 3.9630&\footnotesize  0.0619&\footnotesize 0.0017\\
		\hline
		\footnotesize 3&\footnotesize 3-8-8-3&\footnotesize 20.5776&\footnotesize 3.9711&\footnotesize  0.2503&\footnotesize 0.0018\\
		\hline
		\footnotesize 3&\footnotesize 3-16-16-3&\footnotesize 20.5215&\footnotesize 4.0031&\footnotesize  0.7764&\footnotesize 0.0022\\
		\hline
	\end{tabular}
	\vspace{-0.2cm}
	\caption{Quantitative comparison among different settings for $\mathcal{H}_{\bm{\theta}}$ on MIT testing dataset. In which, the basic block contains a convolutional layer with the size of $3\times 3$ and a ReLU layer. ``Blocks'' and ``Channels'' represent the numbers of the basic block and the variation of channels in the basic block, respectively. }
	\label{table:adaptability}
	\vspace{-0.2cm}
\end{table}

\begin{figure}[t]
	\centering
	\begin{tabular}{c@{\extracolsep{0.3em}}c@{\extracolsep{0.3em}}c} 
		\includegraphics[width=0.147\textwidth]{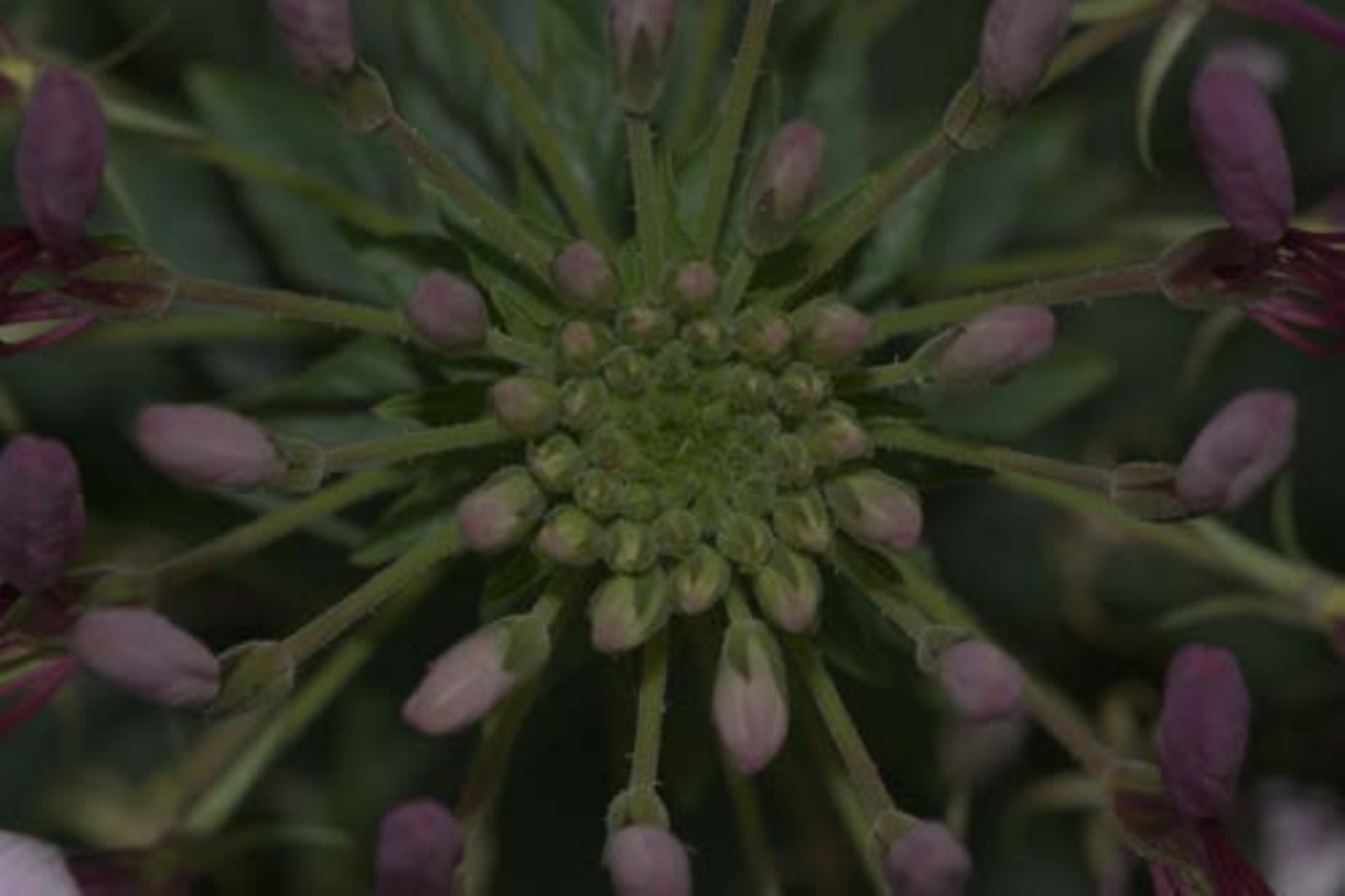}&
		\includegraphics[width=0.147\textwidth]{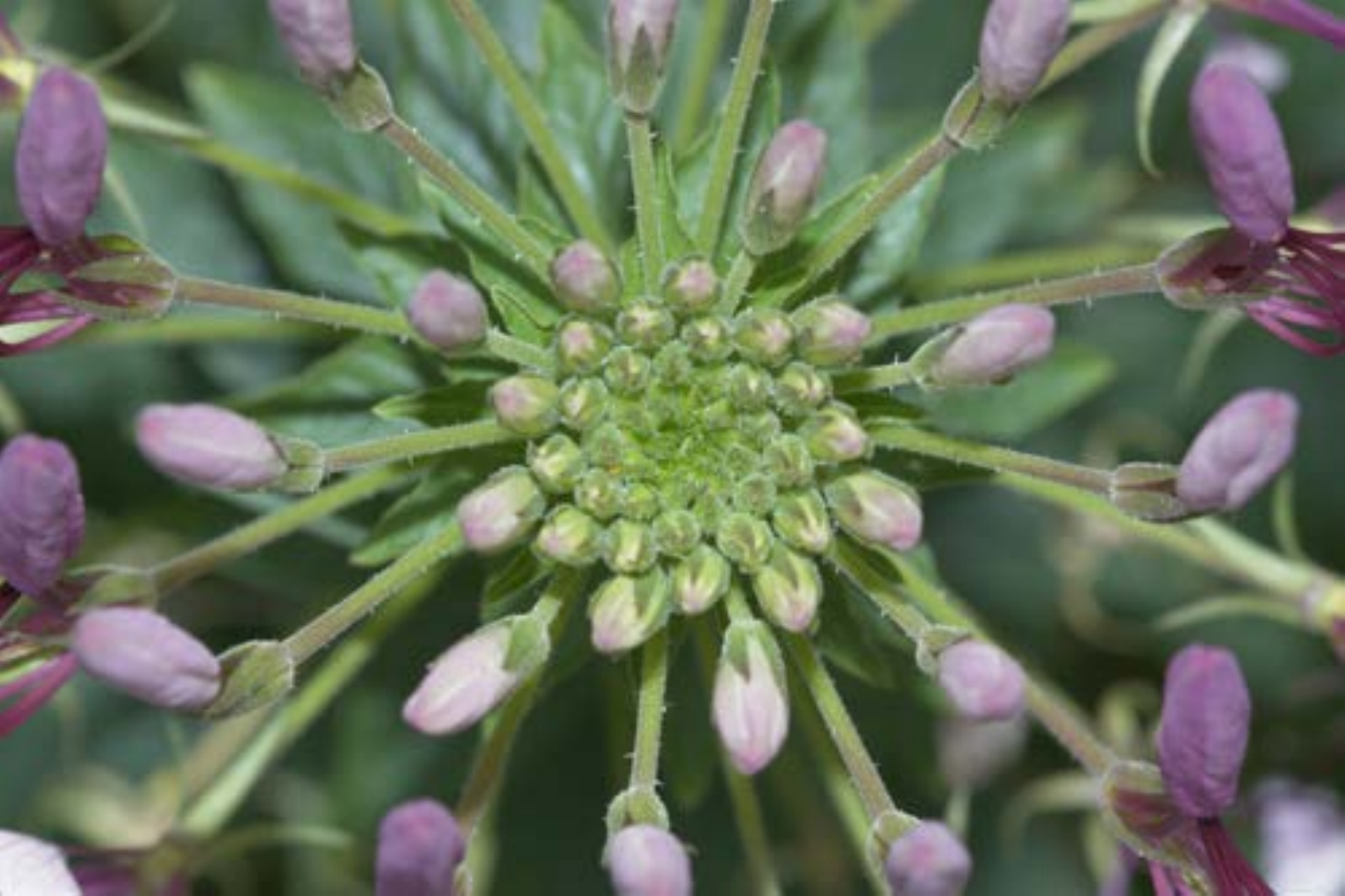}&
		\includegraphics[width=0.147\textwidth]{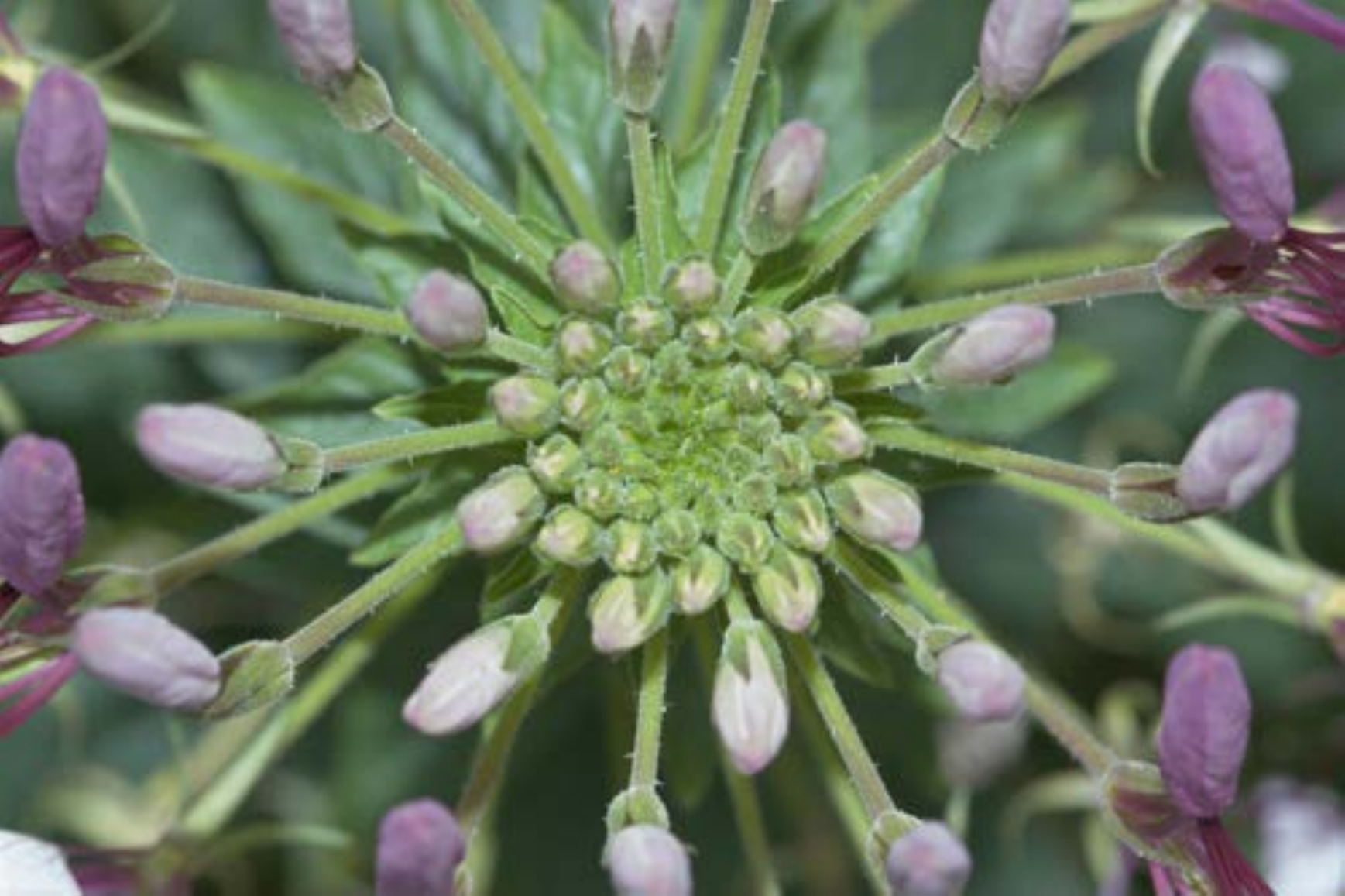}\\
		\footnotesize Input&\footnotesize 1 block(3-3)&\footnotesize 2 blocks (3-3-3)\\	
		\includegraphics[width=0.147\textwidth]{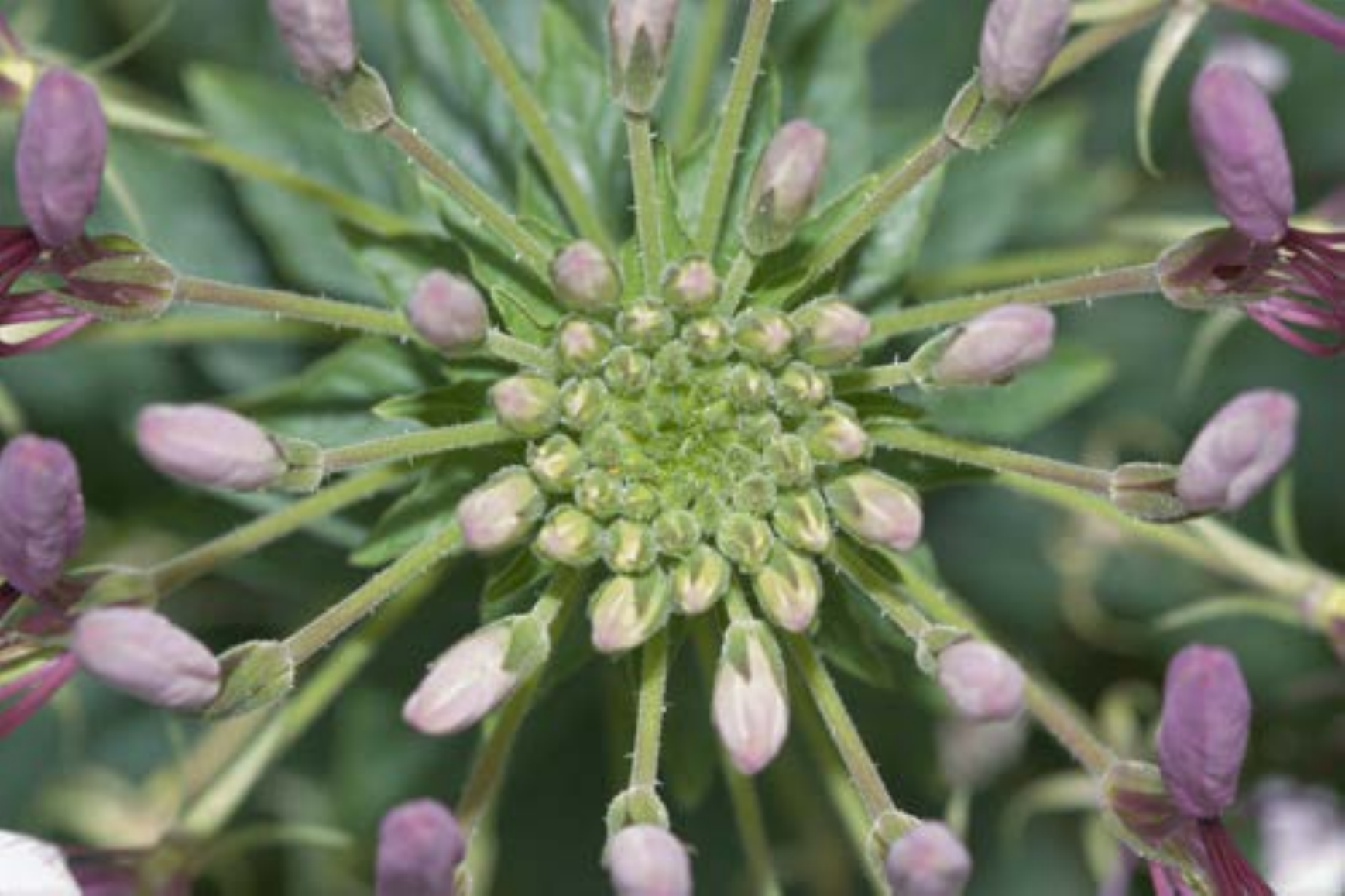}&
		\includegraphics[width=0.147\textwidth]{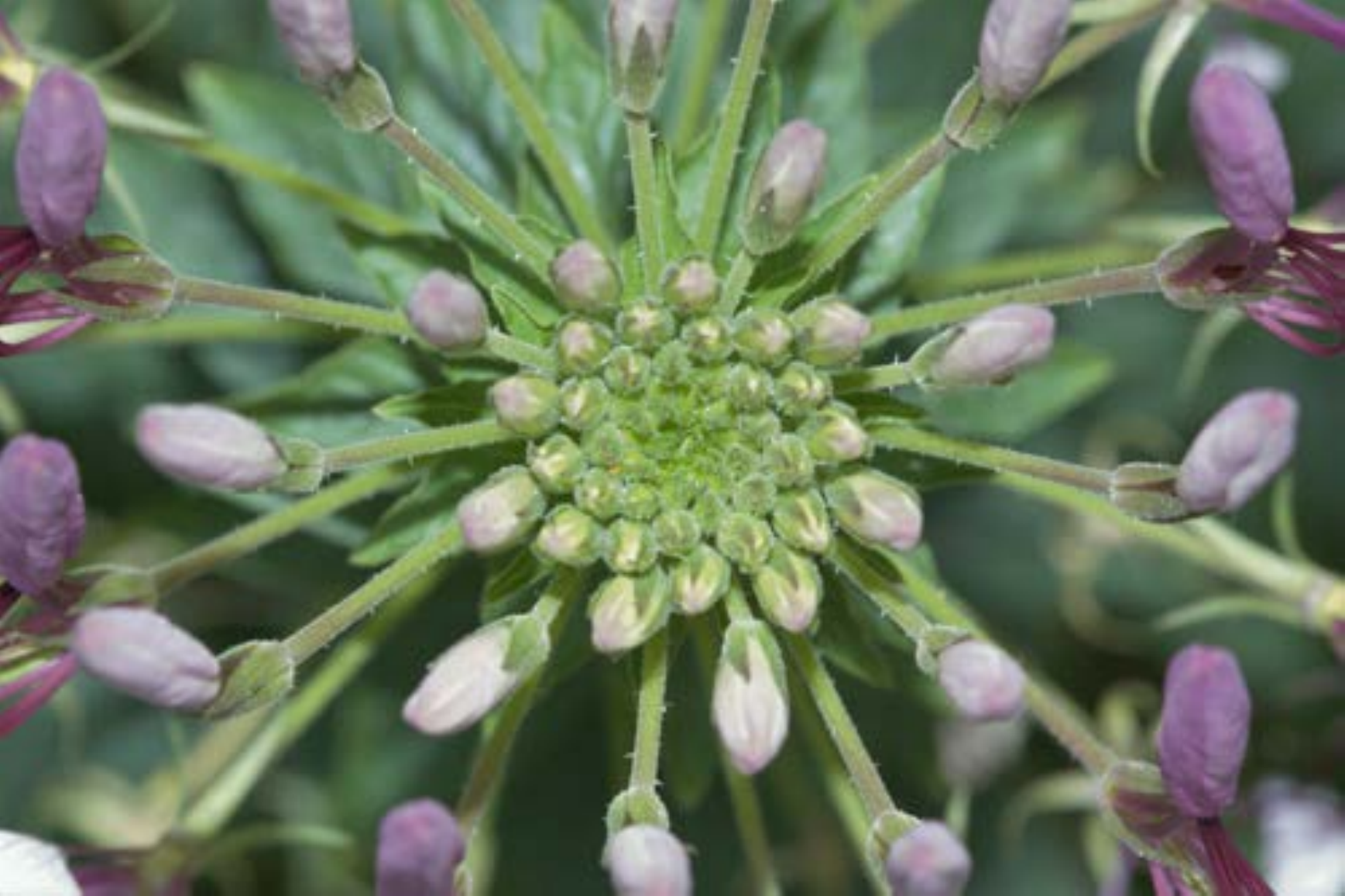}&
		\includegraphics[width=0.147\textwidth]{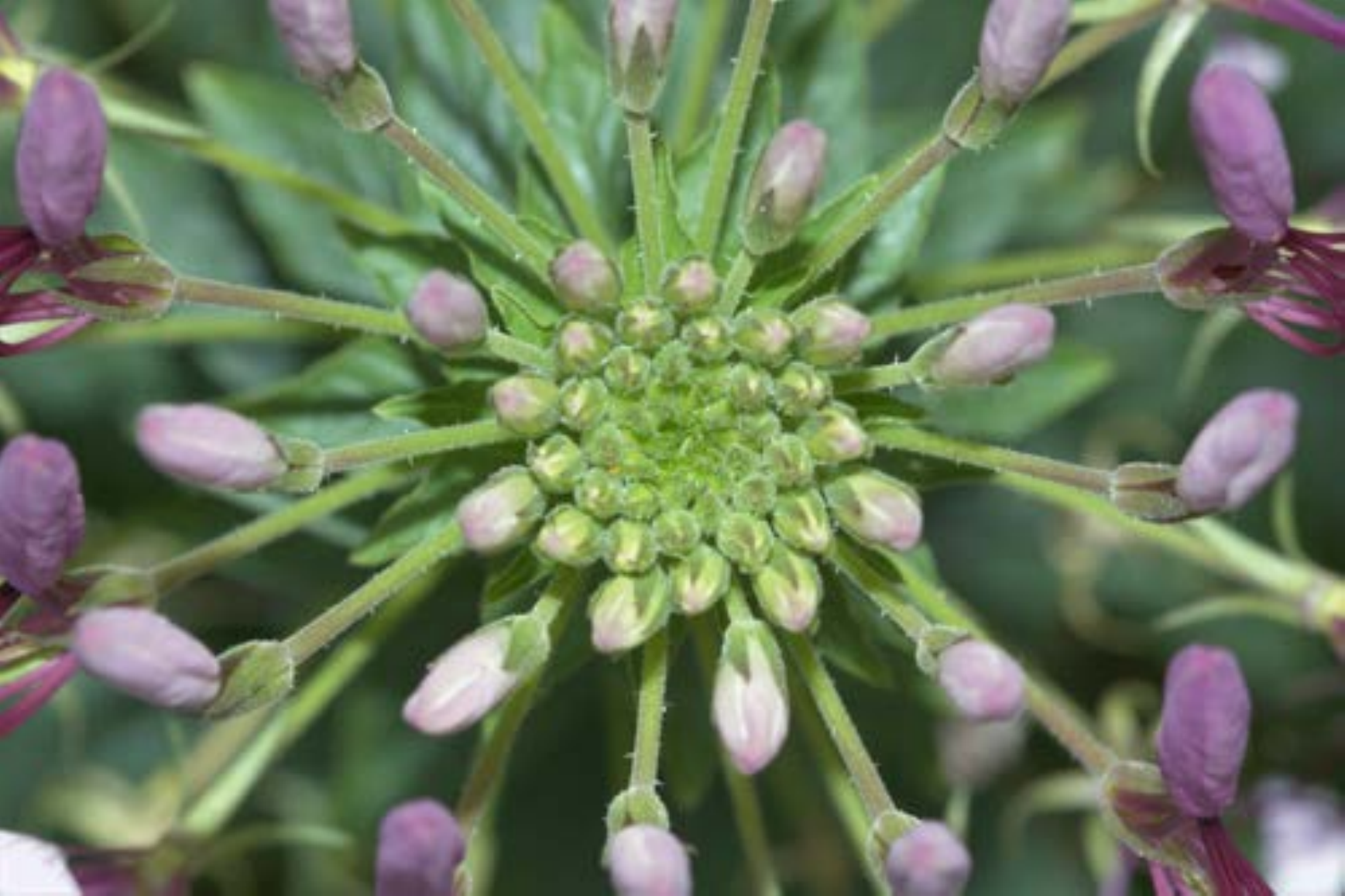}\\
		\footnotesize 3 blocks (3-3-3-3)&\footnotesize 3 blocks (3-8-8-3)&\footnotesize 3 blocks (3-16-16-3)\\
	\end{tabular}
	\vspace{-0.2cm}
	\caption{Visual comparison among different cases in Table~\ref{table:adaptability}. }
	\label{fig:adaptability}
	\vspace{-0.2cm}
\end{figure}

\subsection{Discussion}\label{sec:discussion}
In essence, the self-calibrated module plays an auxiliary role in learning a better basic block (the illumination estimation block in this work) that is cascaded to generate the overall illumination learning process with the weight sharing mechanism. More importantly, the self-calibrated module confers the convergence between results of each stage, it yet has not been explored in existing works. Moreover, the core idea of SCI is actually introducing the additional network module to assist in training, but not in the testing. It improves model characterization to realize that only using the single block for testing. That is to say, the mechanism ``\textit{weight sharing + task-related self-calibrated module}''  may be transferred to handle other tasks for acceleration. 
%The network architecture and optimization objective for training have been detailed expounded in the above. Now we introduce the testing procedure. Different from existing works which utilize the same architecture in the training and testing phases, our method aims to improve the capability of the given simple network by the newly-constructed learning mechanism, to avoid introducing the additional computational overhead. Moreover, in our training procedure, $\mathcal{F}$ remains the weight sharing at each stage, and the $\mathcal{G}$ plays an auxiliary role for $\mathcal{F}$ and it also keeps the weight sharing mechanism at each stage. Actually, the weight sharing manner can ensure these two modules cost less time to a stable state. Thus, we can directly adopt a single illumination estimation module as our final testing module. Extensive analytical experiments have been conducted in Sec.~\ref{sec:ana} to prove the effectiveness and necessity of the method above. 

\section{Exploring Algorithmic Properties}
In this section, we perform explorations about our proposed SCI to deeply analyze its properties.

\begin{table}[t]
	\renewcommand\arraystretch{1.1} 
	\setlength{\tabcolsep}{1.2mm}
	\centering
	\begin{tabular}{|c|ccccc|}
		\hline
		\footnotesize Model&\footnotesize PSNR&\footnotesize EME&\footnotesize NIQE&\footnotesize FLOPs (G)&\footnotesize TIME (S)\\
		\hline
		\footnotesize RUAS (3)&\footnotesize  14.4372&\footnotesize 23.5139&\footnotesize 4.1684&\footnotesize  0.2813&\footnotesize 0.0063\\ 
		\hline
		\footnotesize RUAS (1) + SCI&\footnotesize 14.7352&\footnotesize 24.4884&\footnotesize 3.8588&\footnotesize  0.0936&\footnotesize 0.0022\\ 
		\hline
	\end{tabular}
	\vspace{-0.2cm}
	\caption{SCI can be applied to improve the performance for existing works, e.g., RUAS~\cite{liu2021retinex}. In which RUAS (d) represents adopting d iterative blocks for the unrolling process appeared in RUAS. Here we adopt the LSRW~\cite{hai2021r2rnet} dataset for testing. }
	\label{table:generality}
	\vspace{-0.2cm}
\end{table}

\begin{figure}[t]
	\centering
	\begin{tabular}{c@{\extracolsep{0.3em}}c@{\extracolsep{0.3em}}c} 
		\includegraphics[width=0.147\textwidth]{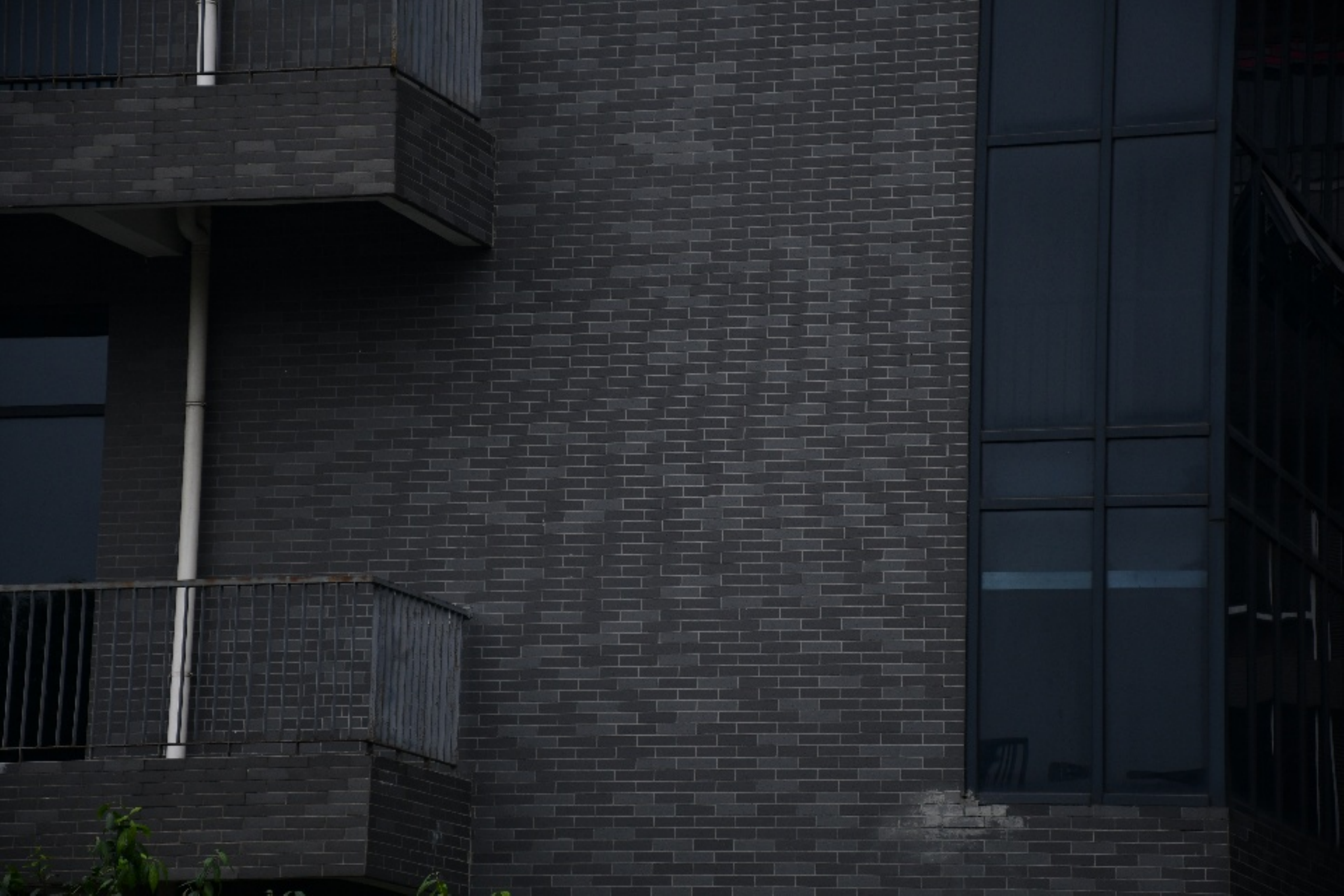}&
		\includegraphics[width=0.147\textwidth]{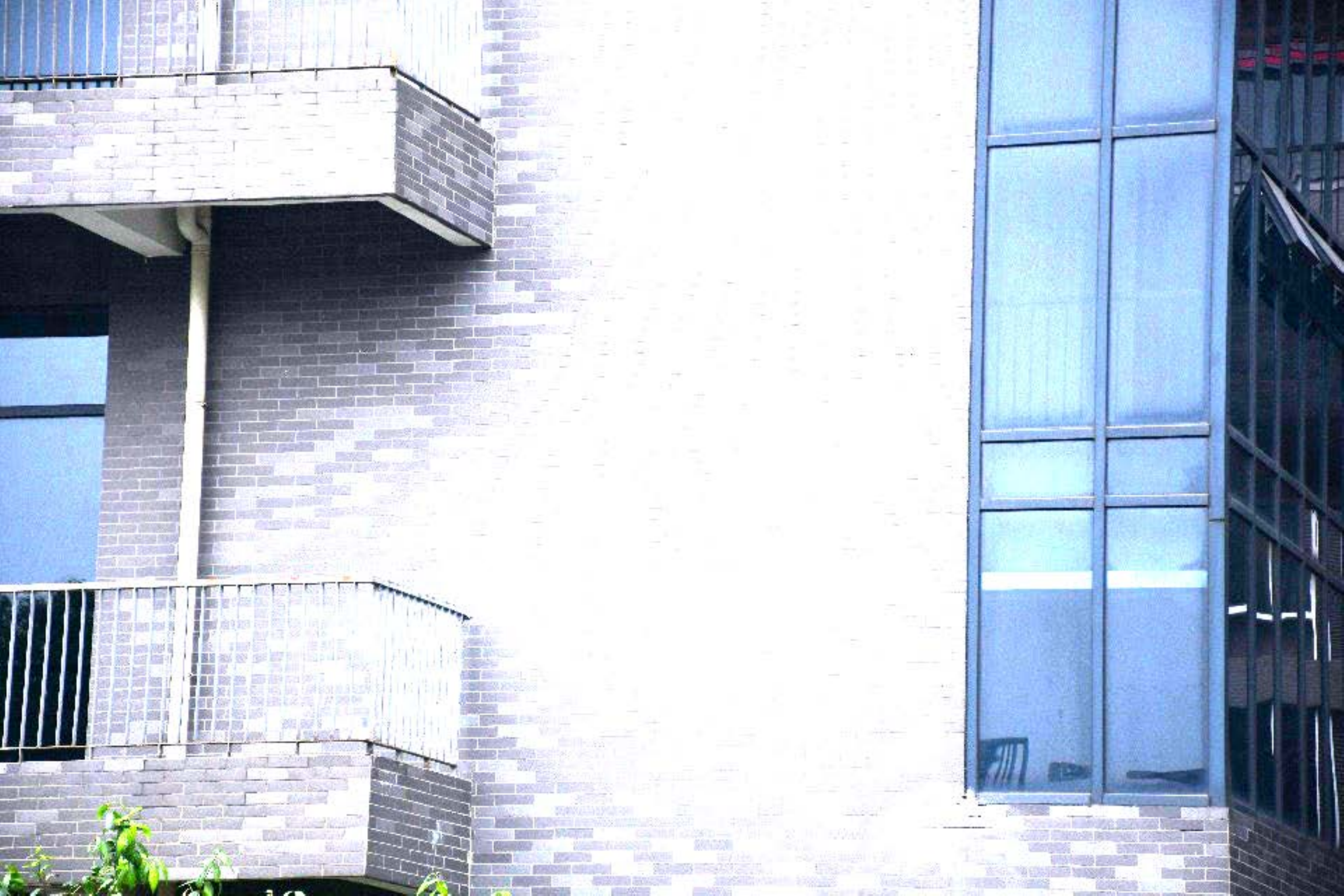}&
		\includegraphics[width=0.147\textwidth]{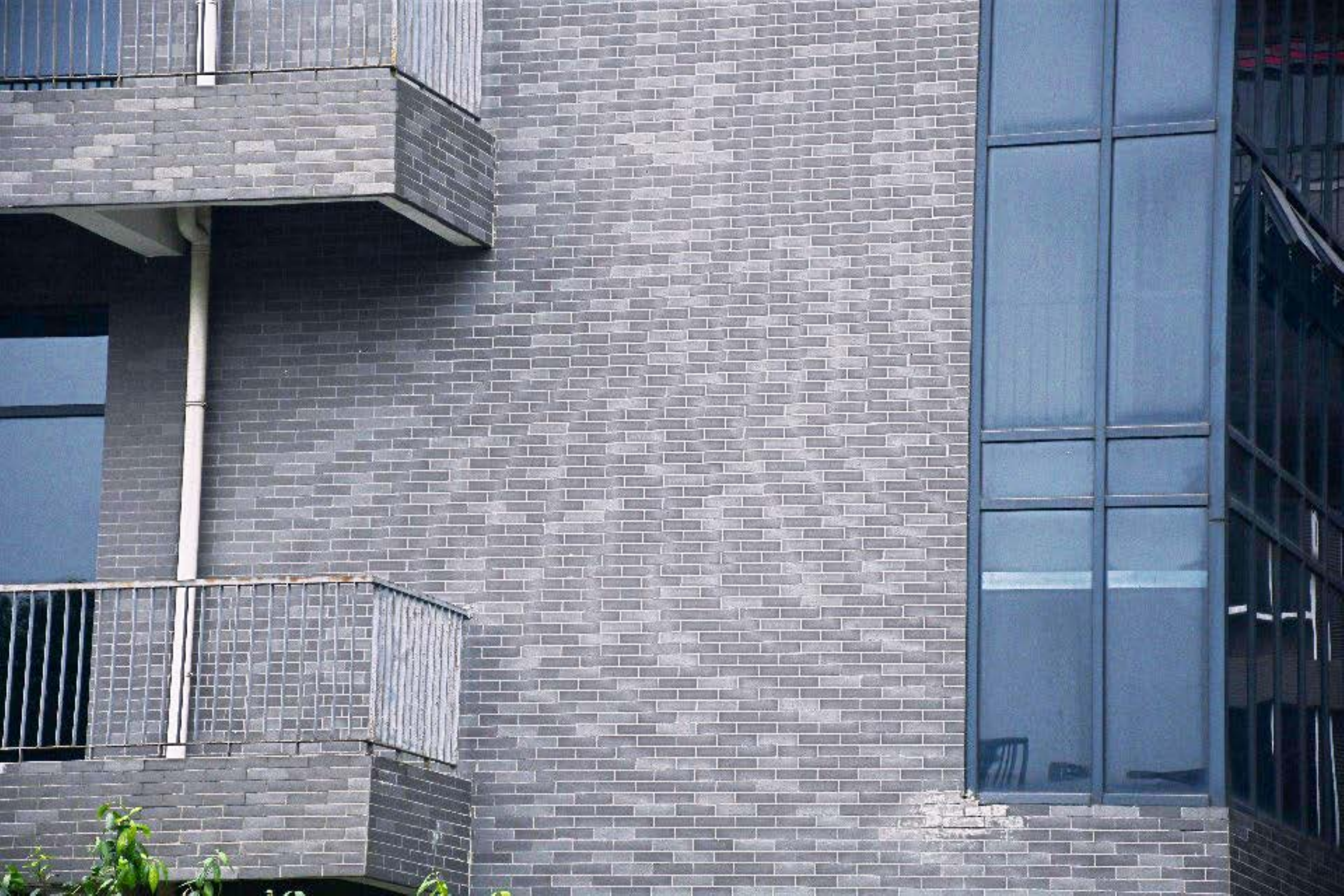}\\
		\includegraphics[width=0.147\textwidth]{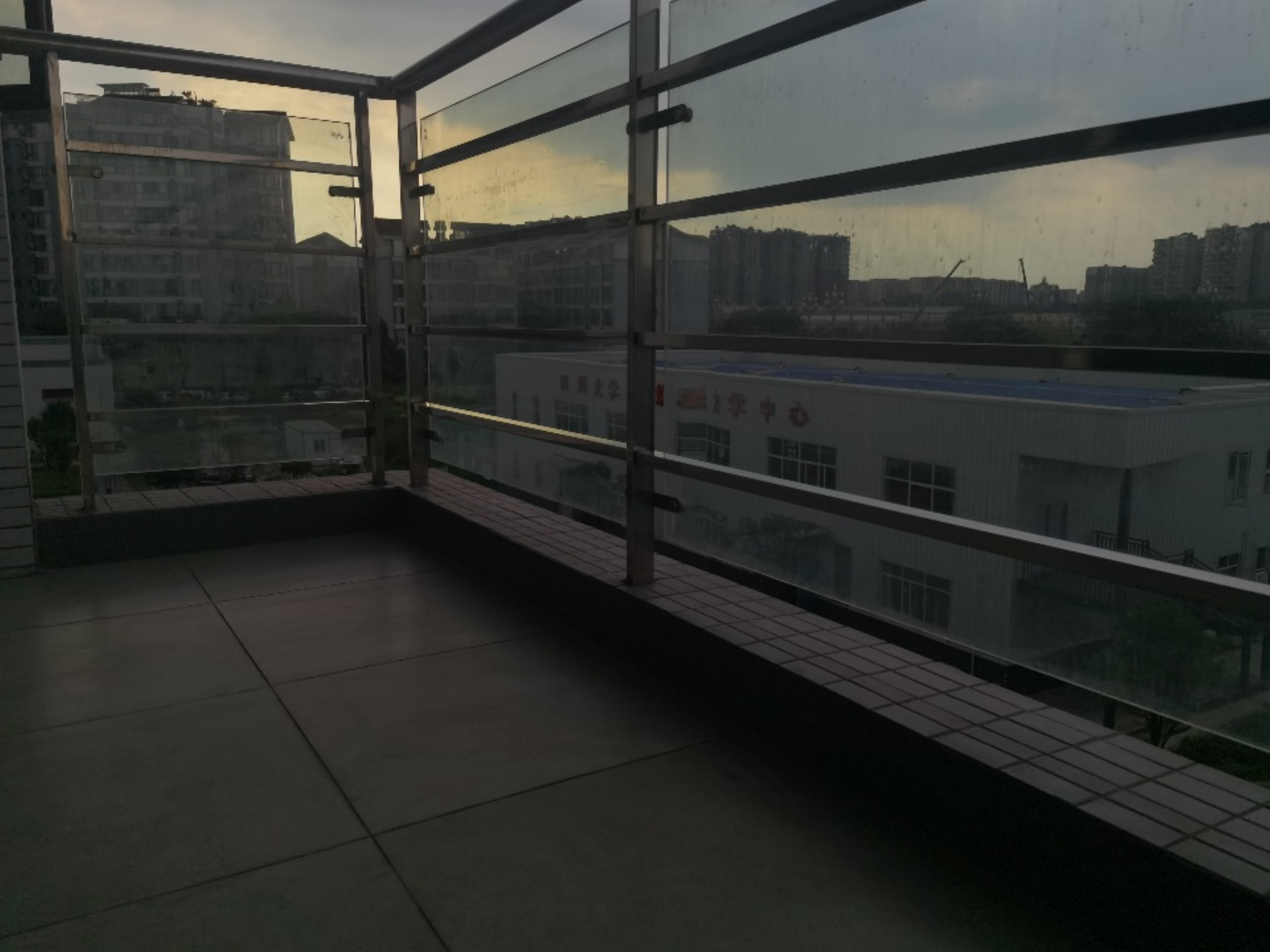}&
		\includegraphics[width=0.147\textwidth]{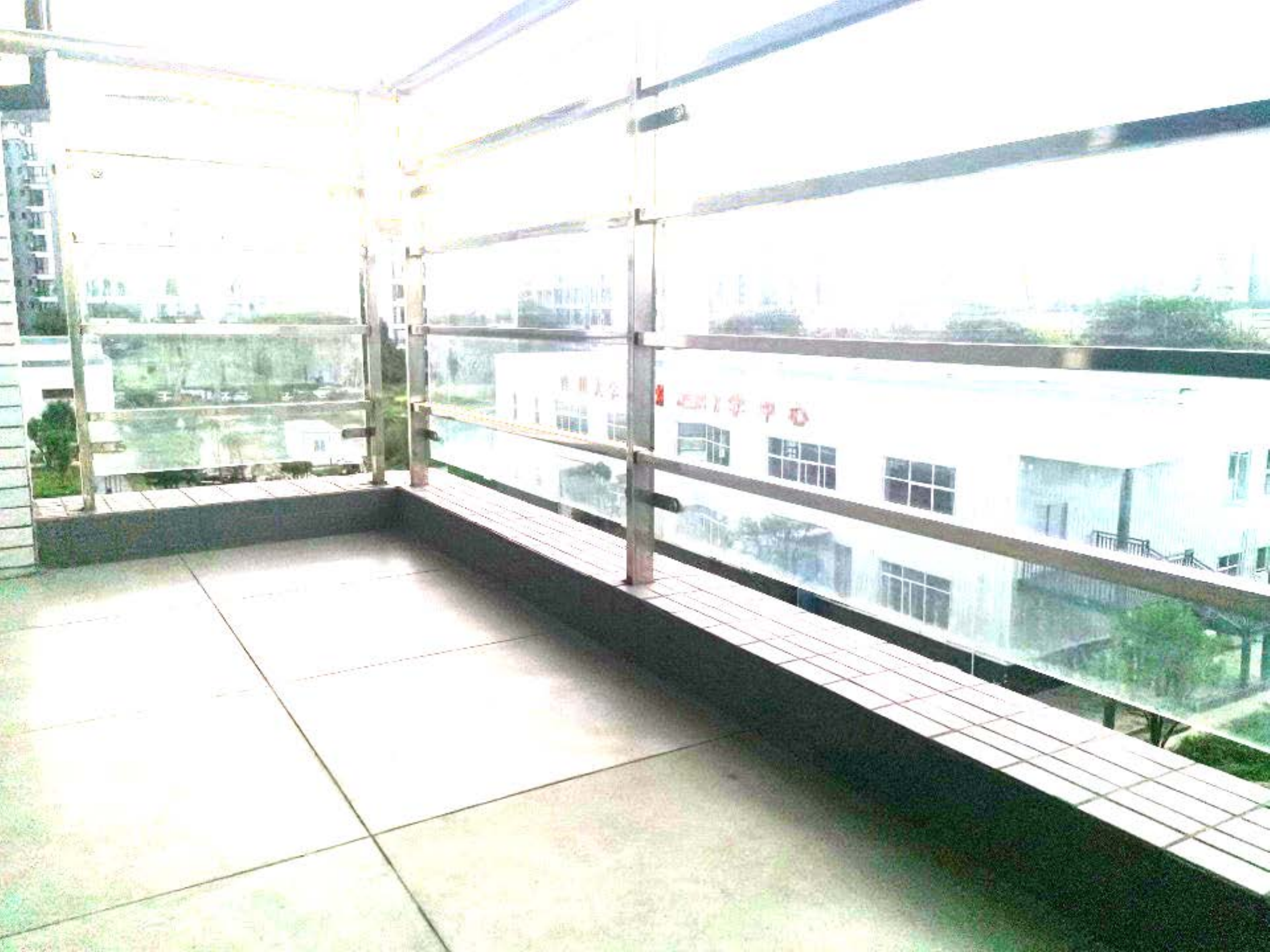}&
		\includegraphics[width=0.147\textwidth]{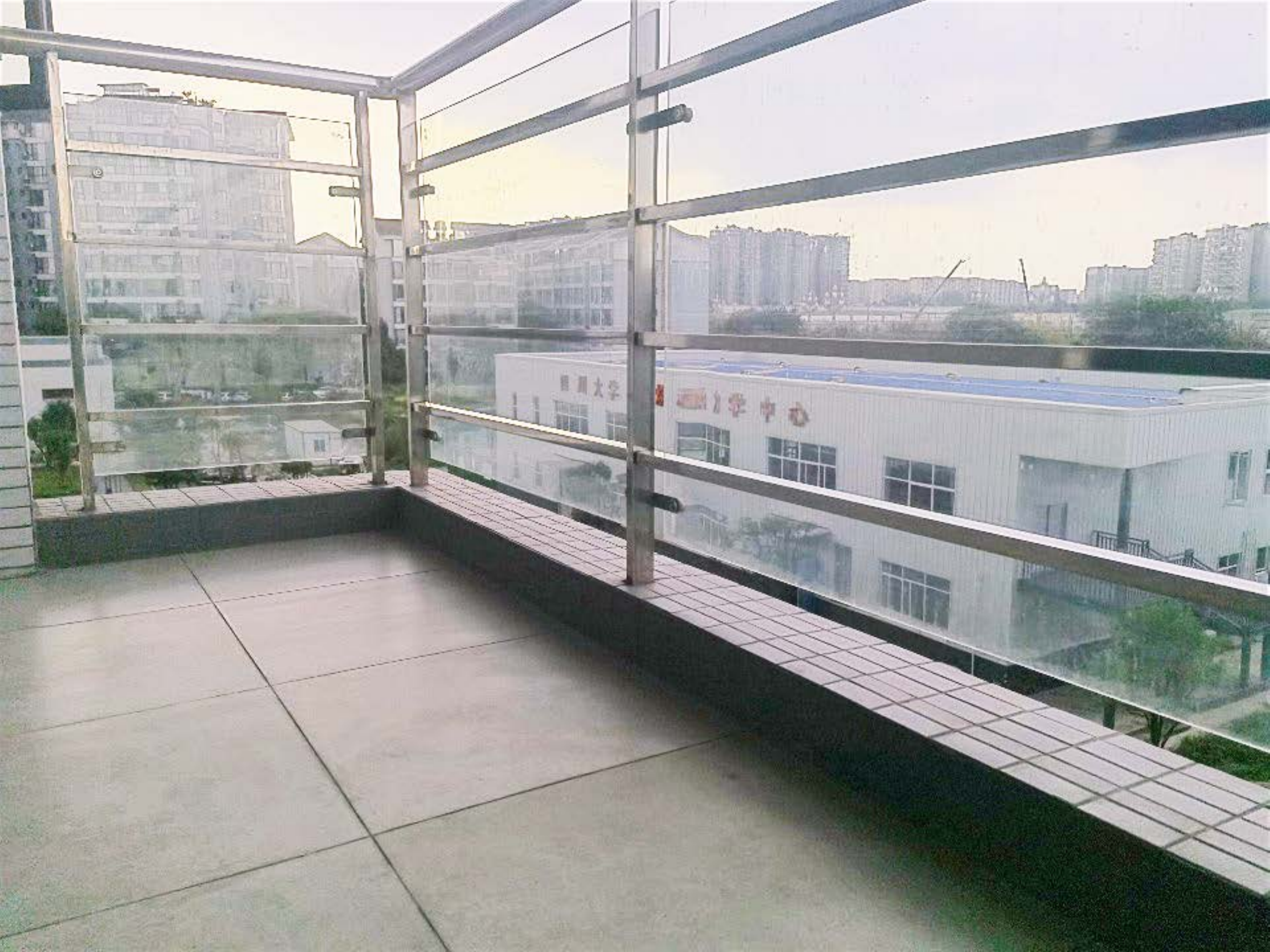}\\
		\footnotesize Input&\footnotesize RUAS (3)&\footnotesize RUAS (1) + SCI\\
	\end{tabular}
	\vspace{-0.2cm}
	\caption{Visual comparison among different cases in Table~\ref{table:generality}. }
	\label{fig:generality}
	\vspace{-0.2cm}
\end{figure}

\begin{table*}[t]
	\renewcommand\arraystretch{1.1} 
	\setlength{\tabcolsep}{0.9mm}
	\centering
	\begin{tabular}{|c|c|ccc|cccc|cccc|c|}
		\hline
		\multirow{2}{*}{\footnotesize Dataset} &{\multirow{2}{*}{\footnotesize Metrics}}&\multicolumn{3}{|c|}{\footnotesize \textit{Recent Traditional Methods}}&\multicolumn{4}{c|}{\footnotesize \textit{Supervised Learning Methods}}&\multicolumn{5}{c|}{\footnotesize \textit{Unsupervised Learning Methods}}\\
		\cline{3-14}
		~&~&\footnotesize LECARM&\footnotesize SDD&\footnotesize STAR&\footnotesize RetinexNet&\footnotesize FIDE&\footnotesize DRBN&\footnotesize KinD&\footnotesize EnGAN&\footnotesize SSIENet&\footnotesize ZeroDCE&\footnotesize RUAS&{\footnotesize Ours}\\
		%			~&~&\scriptsize  (TCSVT'20)&\scriptsize  (TMM'20)&\scriptsize  (TIP'20)&\scriptsize  (BMVC'18)&\scriptsize (CVPR'20)&\scriptsize (CVPR'20)&\scriptsize  (IJCV'21)&\scriptsize (TIP'21)&\scriptsize (Arxiv'20)&\scriptsize (TPAMI'21)&\scriptsize (CVPR'21)&~\\
		\hline
		\multirow{6}{*}{\footnotesize MIT}&\footnotesize PSNR$\uparrow$&\footnotesize 17.5993&\footnotesize \textcolor{blue}{\textbf{19.5241}}&\footnotesize 17.6464&\footnotesize  13.7444&\footnotesize 17.1902&\footnotesize 17.5910&\footnotesize 17.0935&\footnotesize 16.7682&\footnotesize 10.1396 &\footnotesize 16.6114&\footnotesize 18.5372&\footnotesize \textcolor{red}{\textbf{20.4459}}\\
		&\footnotesize SSIM$\uparrow$&\footnotesize 0.8556&\footnotesize \textcolor{blue}{\textbf{0.8690}}&\footnotesize  0.7793&\footnotesize 0.7394&\footnotesize 0.7853&\footnotesize 0.7840&\footnotesize 0.8307&\footnotesize 0.8346&\footnotesize 0.6456&\footnotesize 0.8144&\footnotesize 0.8642&\footnotesize \textcolor{red}{\textbf{0.8934}}\\
		\cline{2-14}
		~&\footnotesize DE$\uparrow$&\footnotesize 6.8069&\footnotesize 6.8253&\footnotesize  6.3677&\footnotesize 6.2850&\footnotesize 6.6543&\footnotesize 6.5914&\footnotesize 6.7233&\footnotesize  \textcolor{blue}{\textbf{7.0382}}&\footnotesize 6.3879&\footnotesize 6.2116&\footnotesize 6.9068&\footnotesize  \textcolor{red}{\textbf{7.0429}}\\
		&\footnotesize EME$\uparrow$&\footnotesize  8.8779&\footnotesize 8.6987&\footnotesize   5.9128&\footnotesize 9.1800&\footnotesize 8.4146&\footnotesize 7.4620&\footnotesize 8.5482&\footnotesize 7.9499&\footnotesize 5.3423&\footnotesize 7.8658&\footnotesize  \textcolor{blue}{\textbf{10.6396}}&\footnotesize  \textcolor{red}{\textbf{10.9627}}\\
		%			&\footnotesize VIF$\downarrow$&\footnotesize  3.6963&\footnotesize 3.5990&\footnotesize  1.3509&\footnotesize  2.2066&\footnotesize 2.3121&\footnotesize 2.1923&\footnotesize 3.7770&\footnotesize 3.9640&\footnotesize 2.8350&\footnotesize 2.2602&\footnotesize 4.9012&\footnotesize 3.8854\\
		&\footnotesize LOE$\downarrow$&\footnotesize 613.2689&\footnotesize 505.2951&\footnotesize   \textcolor{red}{\textbf{70.5651}}&\footnotesize 1812.853&\footnotesize 264.4661&\footnotesize 705.2620&\footnotesize 500.6578&\footnotesize 812.9041&\footnotesize 646.9047&\footnotesize 508.2960&\footnotesize 579.0181&\footnotesize  \textcolor{blue}{\textbf{273.3409}}\\
		&\footnotesize NIQE$\downarrow$&\footnotesize 4.3627&\footnotesize 4.6477&\footnotesize  4.2611&\footnotesize 4.5289&\footnotesize 5.2720&\footnotesize 4.8166&\footnotesize 4.2658&\footnotesize \textcolor{red}{\textbf{3.9997}}&\footnotesize 5.2792&\footnotesize 4.0933&\footnotesize 4.1754&\footnotesize \textcolor{red}{\textbf{3.9630}}\\
		\hline
		\multirow{6}{*}{\footnotesize LSRW}&\footnotesize PSNR$\uparrow$&\footnotesize 15.4747&\footnotesize 14.6694&\footnotesize  14.6080&\footnotesize 15.9062&\footnotesize \textcolor{red}{\textbf{17.6694}}&\footnotesize 16.1497&\footnotesize 16.4717&\footnotesize 16.3106&\footnotesize \textcolor{blue}{\textbf{16.7380}}&\footnotesize 15.8337&\footnotesize 14.4372&\footnotesize 15.0168\\
		&\footnotesize SSIM$\uparrow$&\footnotesize 0.4635&\footnotesize 0.5061&\footnotesize   0.5039&\footnotesize 0.3725&\footnotesize \textcolor{red}{\textbf{0.5485}}&\footnotesize \textcolor{blue}{\textbf{0.5422}}&\footnotesize 0.4929&\footnotesize 0.4697&\footnotesize  0.4873&\footnotesize  0.4664&\footnotesize  0.4276&\footnotesize 0.4846\\
		\cline{2-14}
		~&\footnotesize DE$\uparrow$&\footnotesize 5.9980&\footnotesize 6.7307&\footnotesize  6.4943&\footnotesize 6.9392&\footnotesize 6.8745&\footnotesize 7.2051&\footnotesize \textcolor{blue}{\textbf{7.0368}}&\footnotesize 6.6692&\footnotesize \textcolor{red}{\textbf{7.0988}}&\footnotesize 6.8729&\footnotesize 5.6056&\footnotesize 6.5524\\
		&\footnotesize EME$\uparrow$&\footnotesize \textcolor{blue}{\textbf{24.4089}}&\footnotesize 8.5431&\footnotesize  9.4636&\footnotesize 14.6119&\footnotesize 5.6885&\footnotesize 9.9968&\footnotesize 12.0881&\footnotesize 	22.2345&\footnotesize 9.3801&\footnotesize 20.8010&\footnotesize  23.5139&\footnotesize \textcolor{red}{\textbf{24.9625}}\\
		%			&\footnotesize VIF$\uparrow$&\footnotesize  10.2713&\footnotesize 6.1723&\footnotesize  3.7332&\footnotesize 10.6206&\footnotesize 10.1958&\footnotesize 11.6561&\footnotesize 14.3339&\footnotesize 6.8308&\footnotesize 12.8606 &\footnotesize 7.8055&\footnotesize 13.2787&\footnotesize 12.9103\\
		&\footnotesize LOE$\downarrow$&\footnotesize \textcolor{red}{\textbf{34.1438}}&\footnotesize 296.0794&\footnotesize \textcolor{blue}{\textbf{103.2322}}&\footnotesize 591.2793&\footnotesize 194.7405&\footnotesize 755.1283&\footnotesize 379.8994&\footnotesize 248.1947&\footnotesize 261.2802&\footnotesize 219.1284 &\footnotesize 357.4125&\footnotesize 280.8935\\
		&\footnotesize NIQE$\downarrow$&\footnotesize 3.8189&\footnotesize 5.6401&\footnotesize 3.7537 &\footnotesize 4.1479&\footnotesize 4.3277&\footnotesize 4.5500&\footnotesize \textcolor{blue}{\textbf{3.6636}}&\footnotesize 3.7754&\footnotesize 4.0631&\footnotesize 3.7183&\footnotesize 4.1687&\footnotesize \textcolor{red}{\textbf{3.6590}}\\
		\hline
	\end{tabular}
	\vspace{-0.2cm}
	\caption{Quantitative results in terms of two full-reference metrics including PSNR and SSIM, and four no-reference metrics including DE, EME, LOE, and NIQE on the MIT and LSRW datasets.  }
	\label{table:LOLQuan}
	\vspace{-0.2cm}
\end{table*}

\begin{figure*}[t]
	\centering
	\begin{tabular}{c@{\extracolsep{0.3em}}c@{\extracolsep{0.3em}}c@{\extracolsep{0.3em}}c@{\extracolsep{0.3em}}c} 	
		\includegraphics[width=0.192\textwidth]{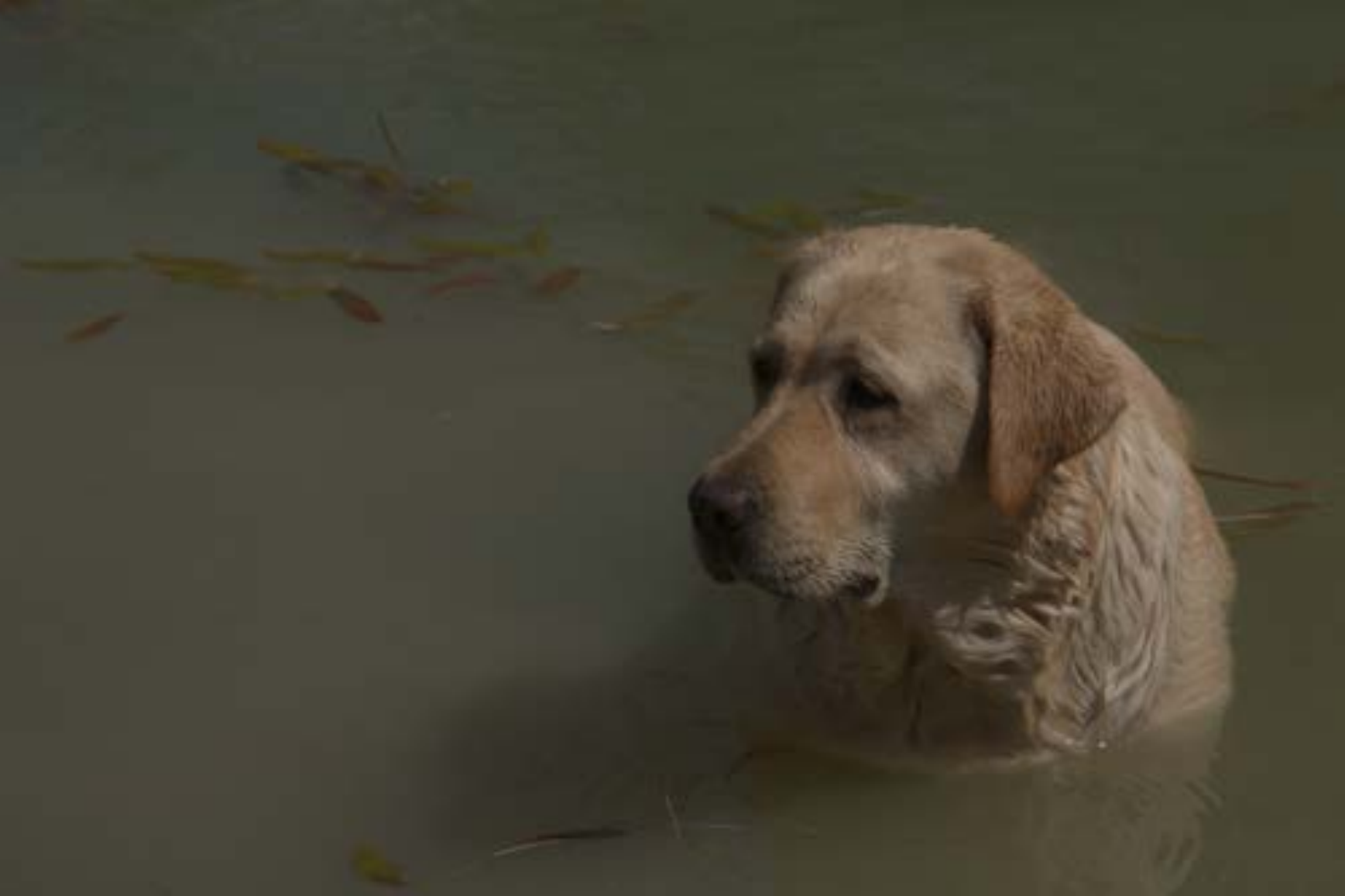}&
		\includegraphics[width=0.192\textwidth]{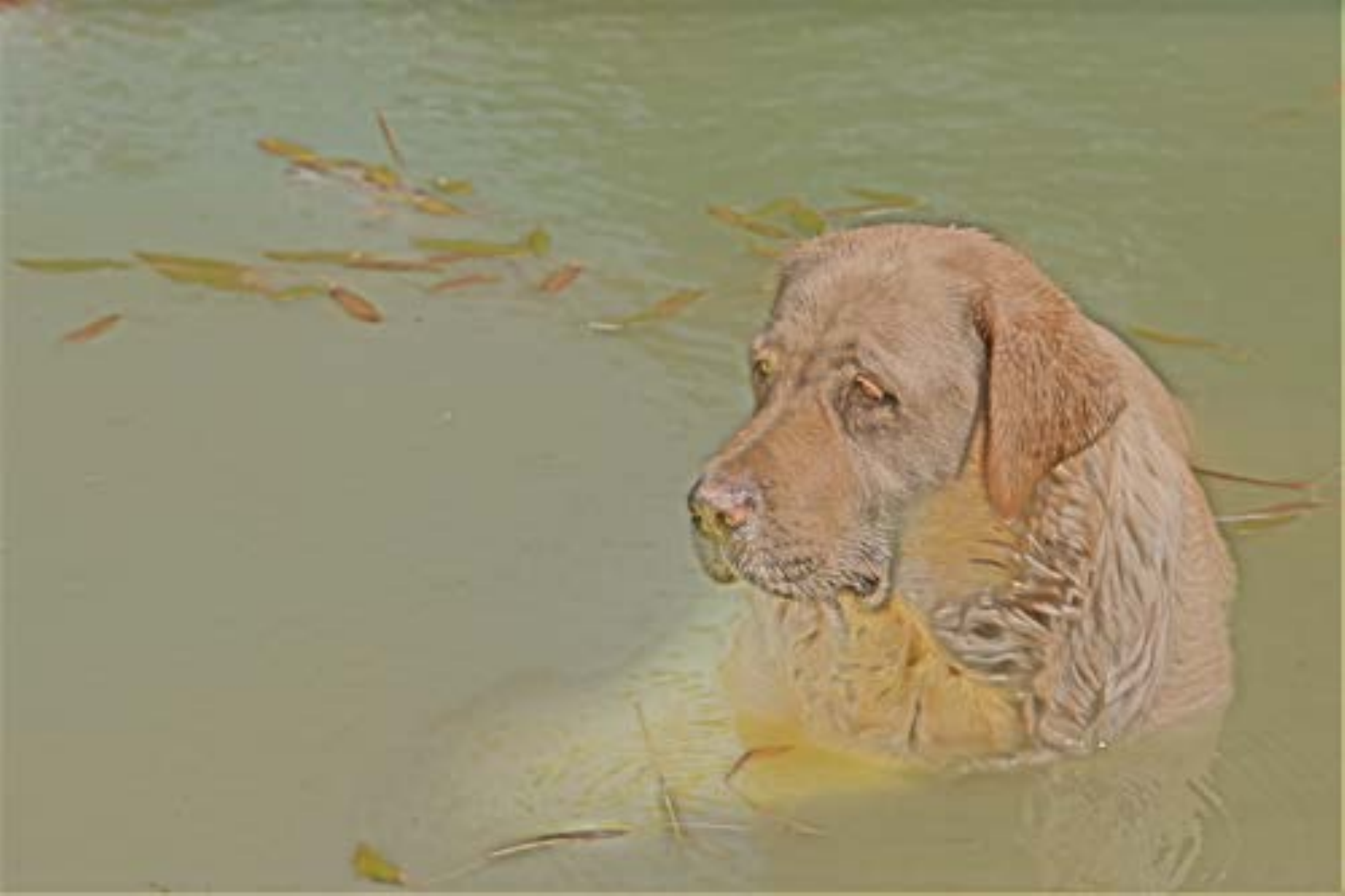}&
		\includegraphics[width=0.192\textwidth]{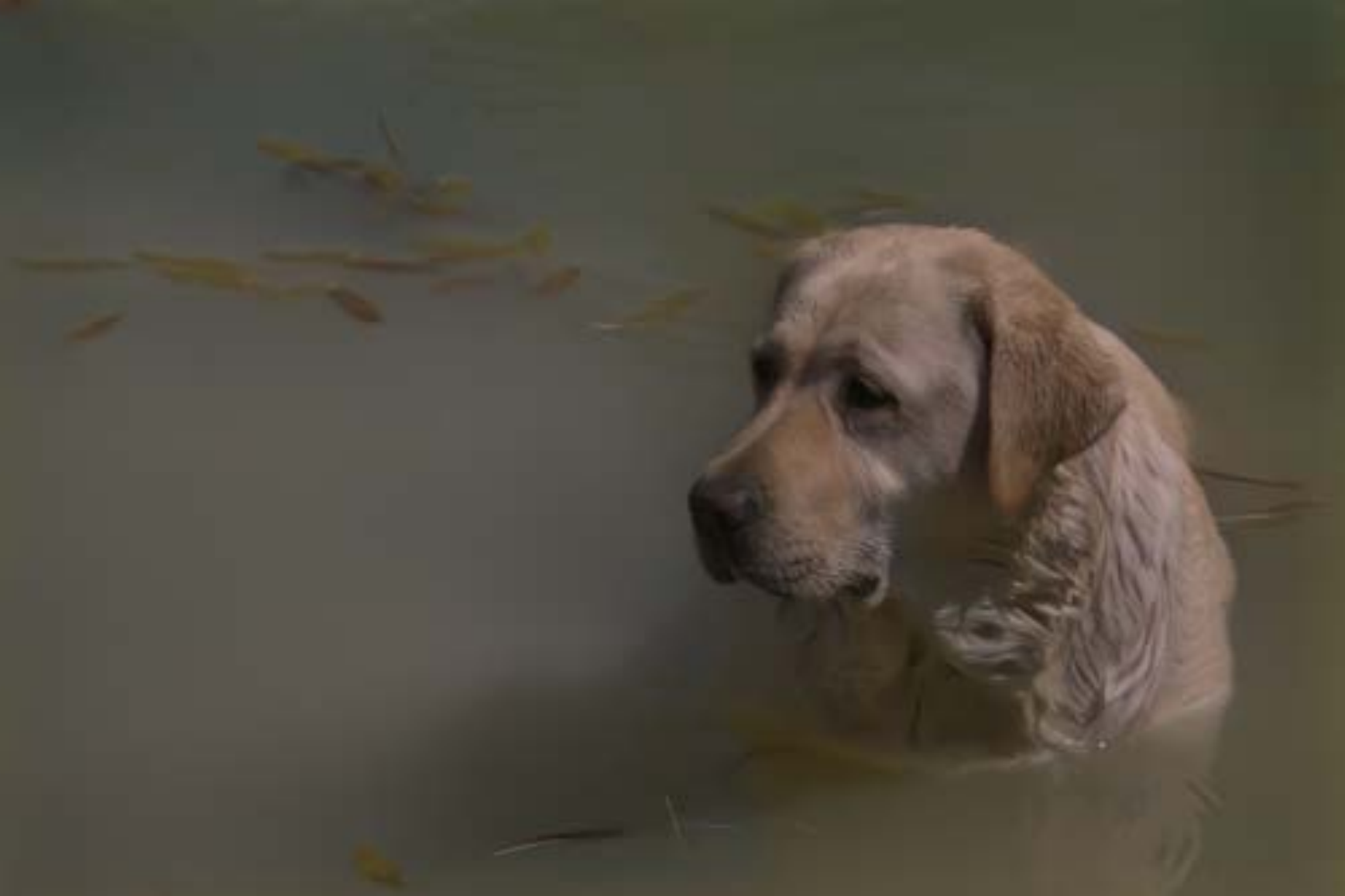}&
		\includegraphics[width=0.192\textwidth]{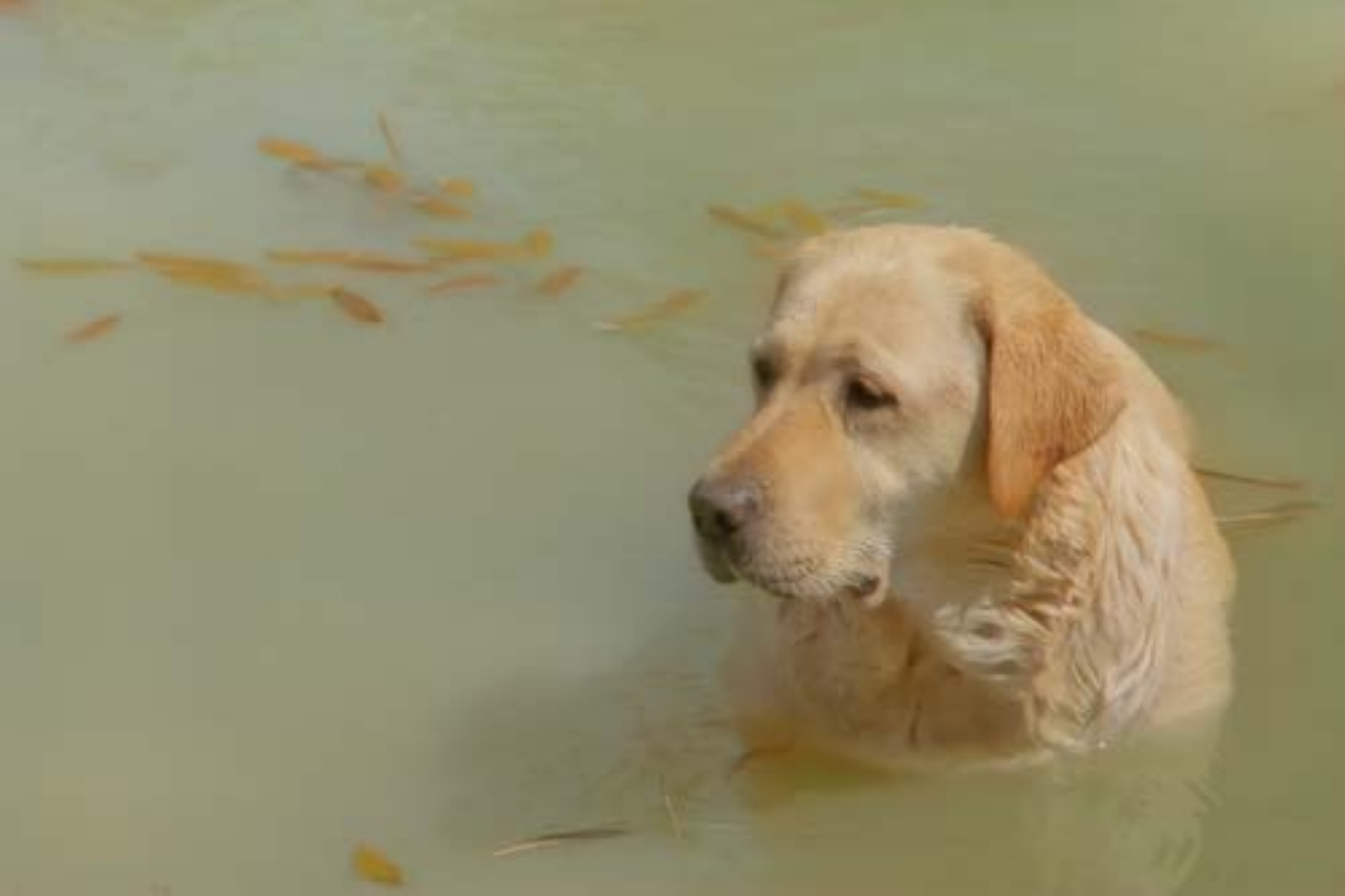}&
		\includegraphics[width=0.192\textwidth]{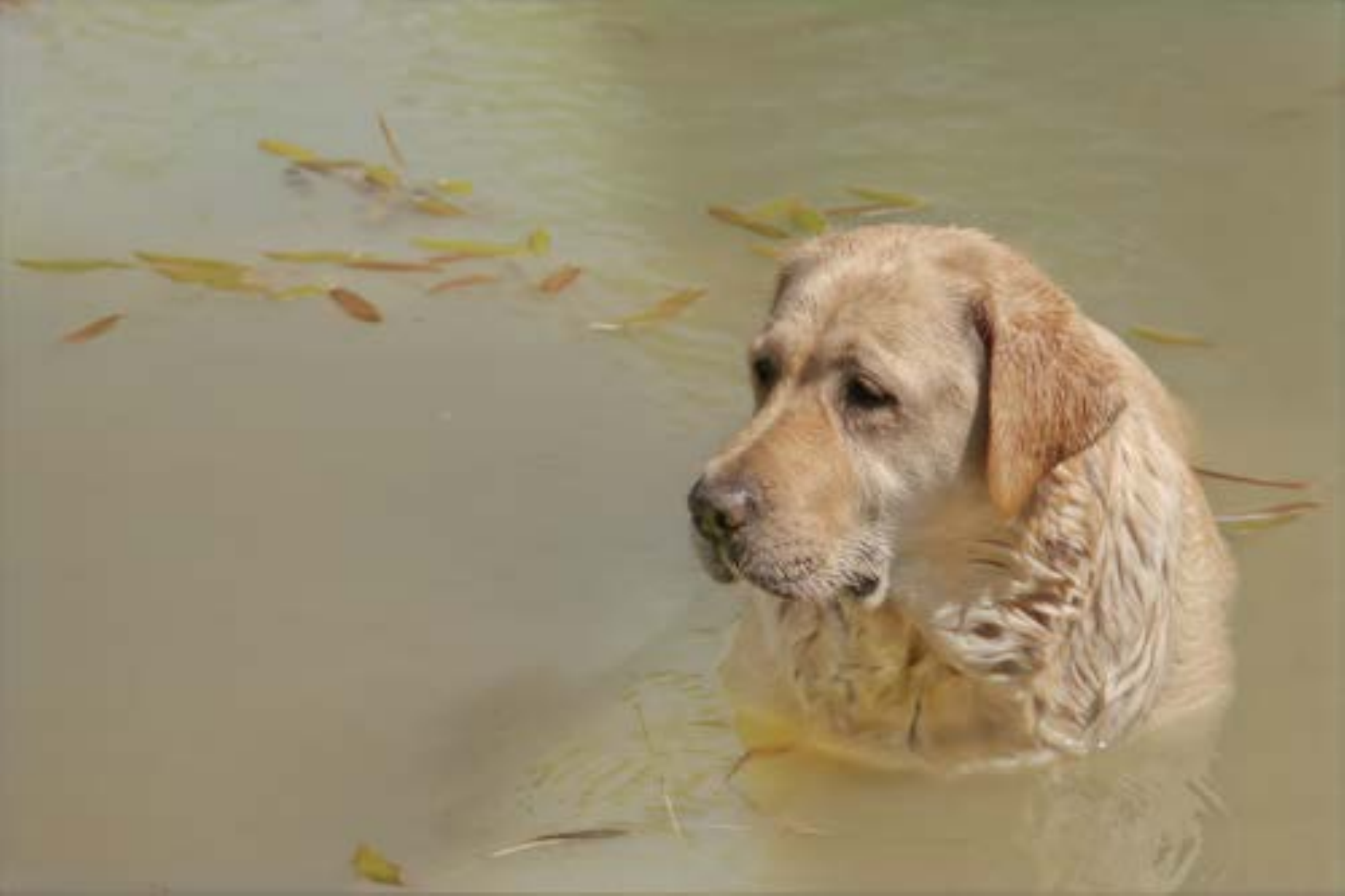}\\
		\footnotesize Input &\footnotesize RetinexNet&\footnotesize FIDE&\footnotesize DRBN&\footnotesize KinD\\
		\includegraphics[width=0.192\textwidth]{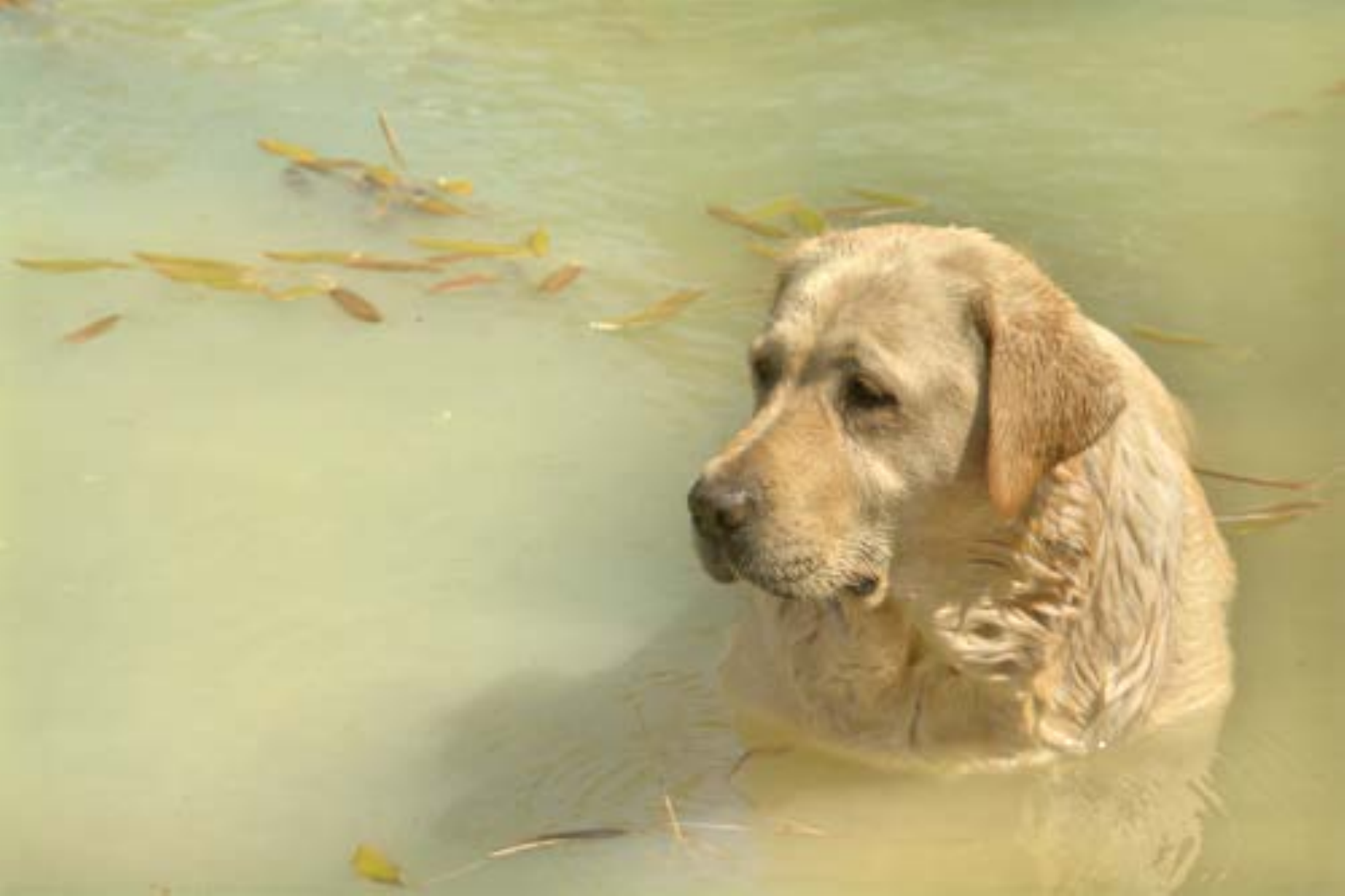}&
		\includegraphics[width=0.192\textwidth]{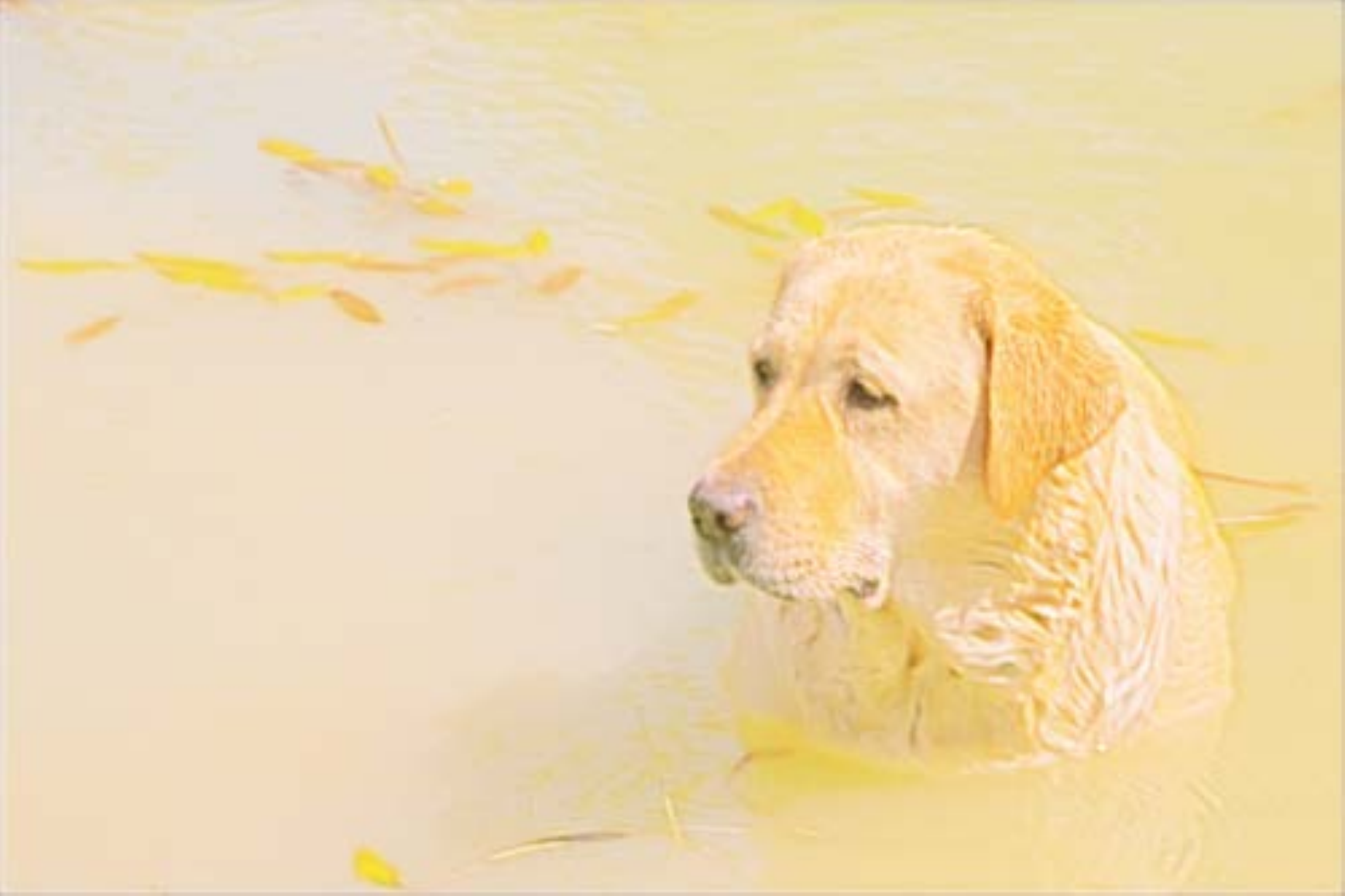}&
		\includegraphics[width=0.192\textwidth]{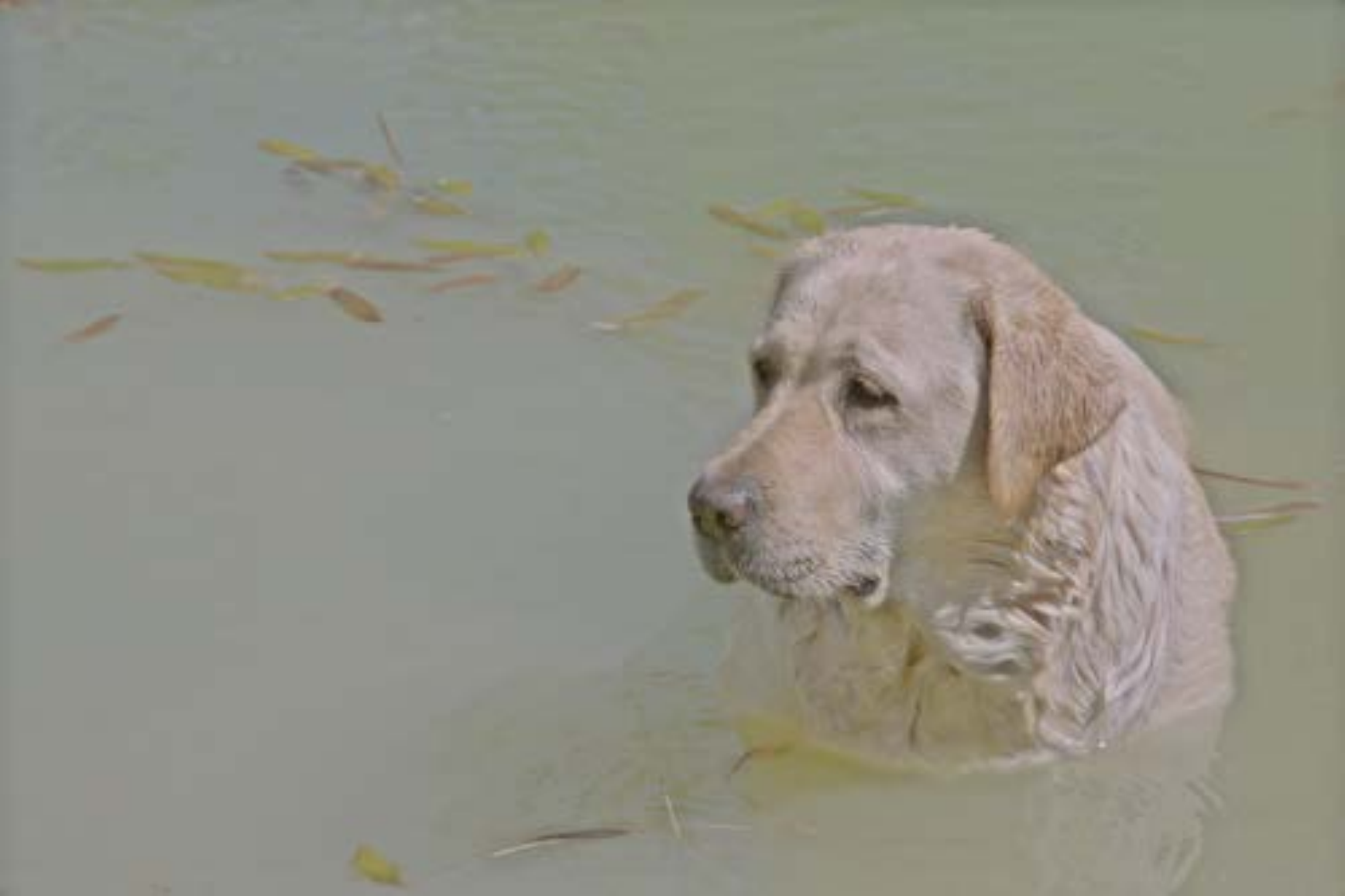}&
		\includegraphics[width=0.192\textwidth]{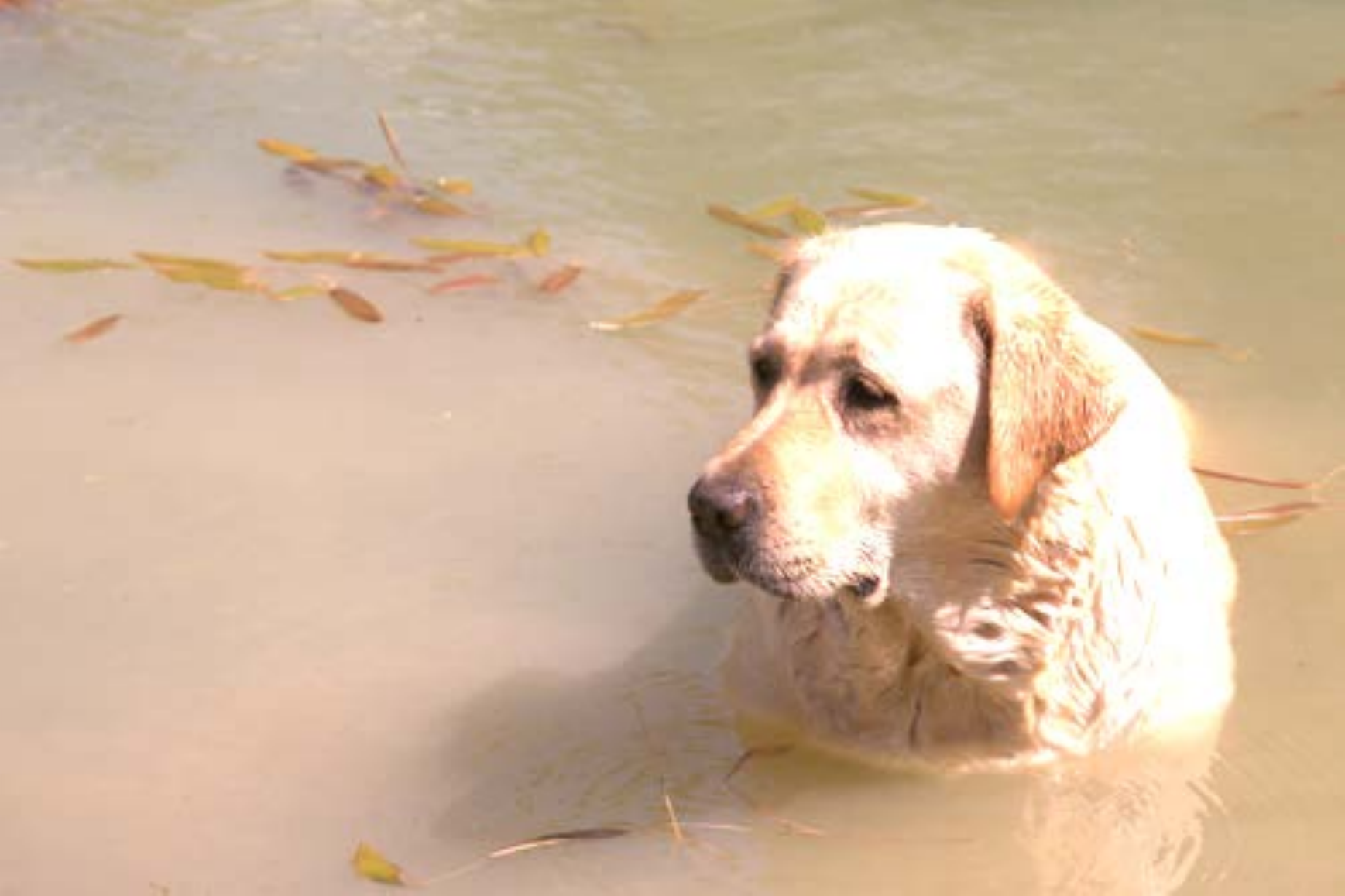}&
		\includegraphics[width=0.192\textwidth]{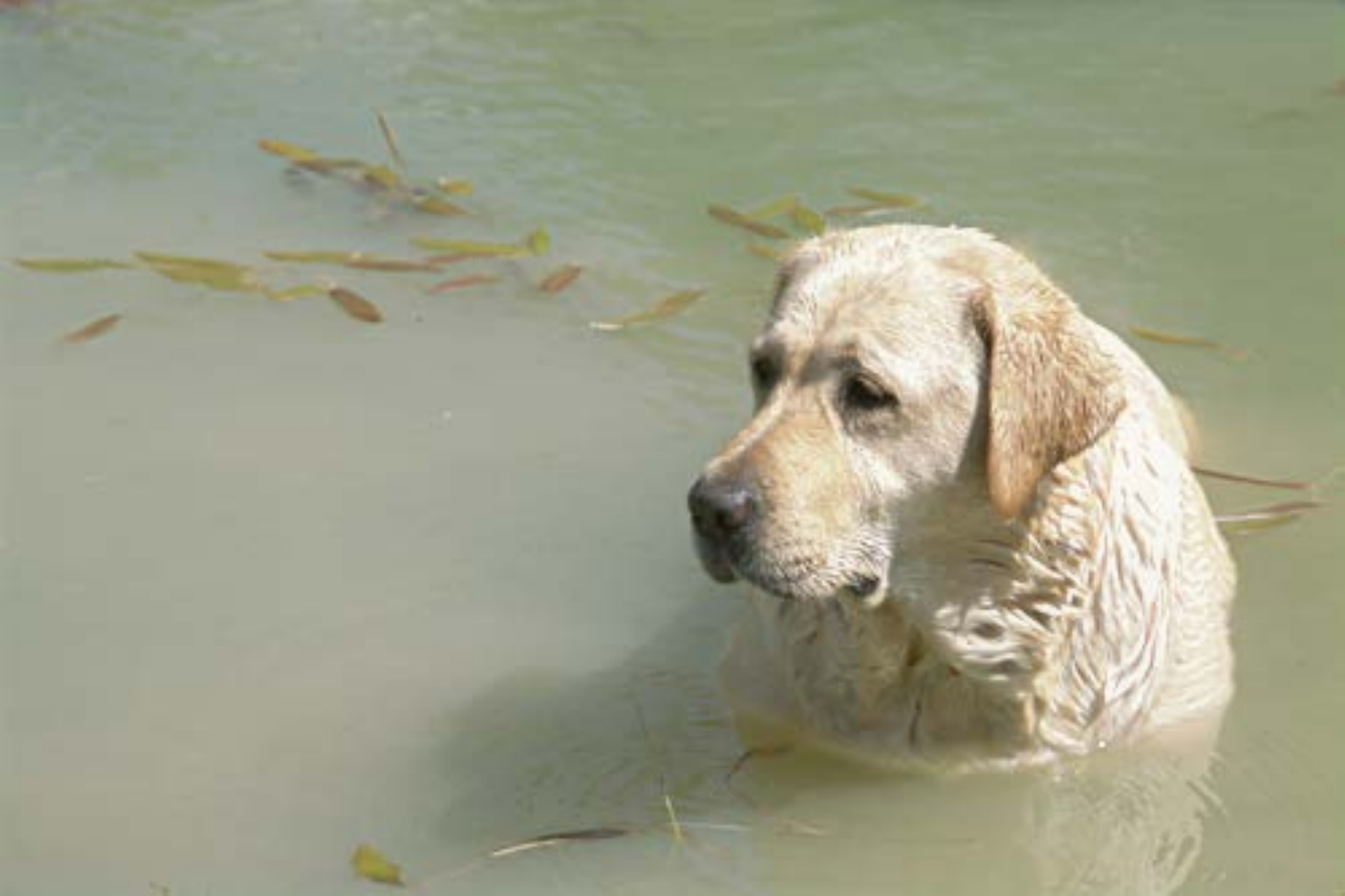}\\
		\footnotesize EnGAN &\footnotesize SSIENet&\footnotesize ZeroDCE &\footnotesize RUAS&\footnotesize Ours\\
	\end{tabular}
	\vspace{-0.3cm}
	\caption{Visual comparison on the MIT dataset among state-of-the-art low-light image enhancement approaches. }
	\label{fig:MIT}
		\vspace{-0.2cm}
\end{figure*}

\begin{figure*}[!htb]
	\centering
	\begin{tabular}{c@{\extracolsep{0.3em}}c@{\extracolsep{0.3em}}c@{\extracolsep{0.3em}}c@{\extracolsep{0.3em}}c} 	
		\includegraphics[width=0.192\textwidth]{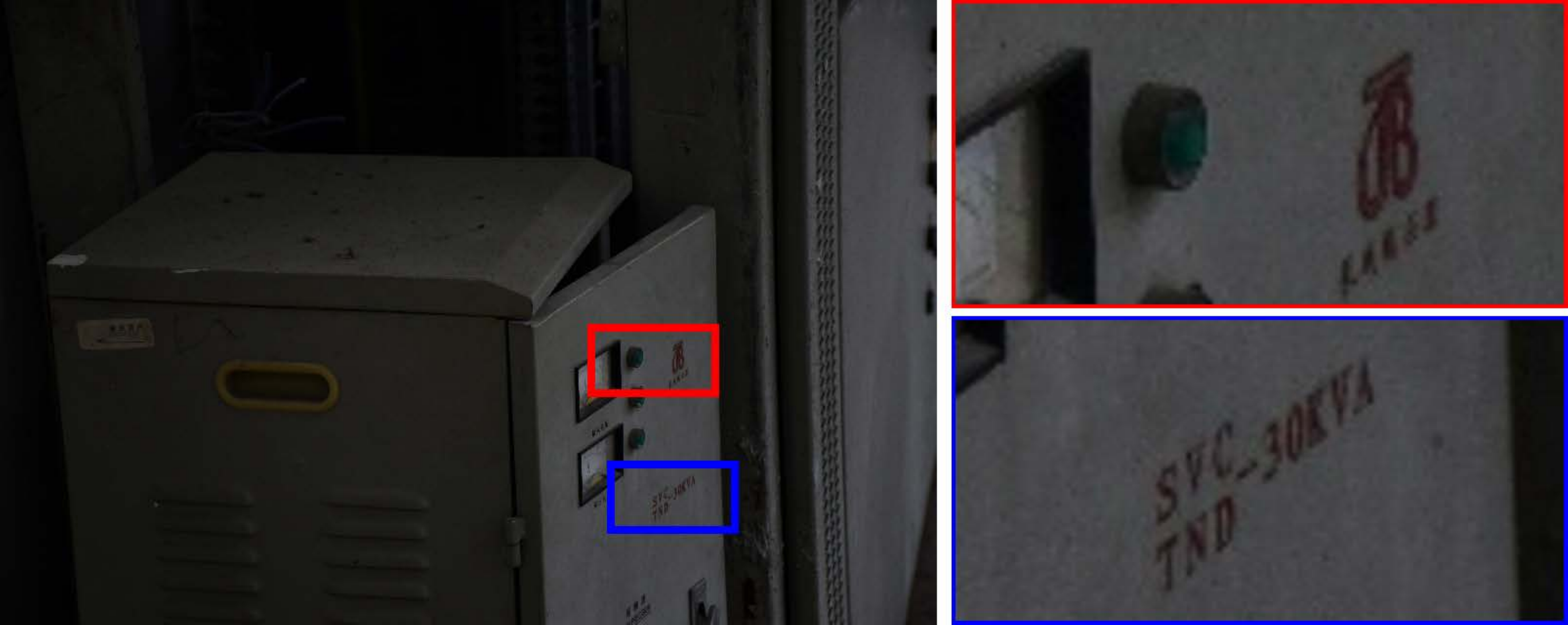}&
		\includegraphics[width=0.192\textwidth]{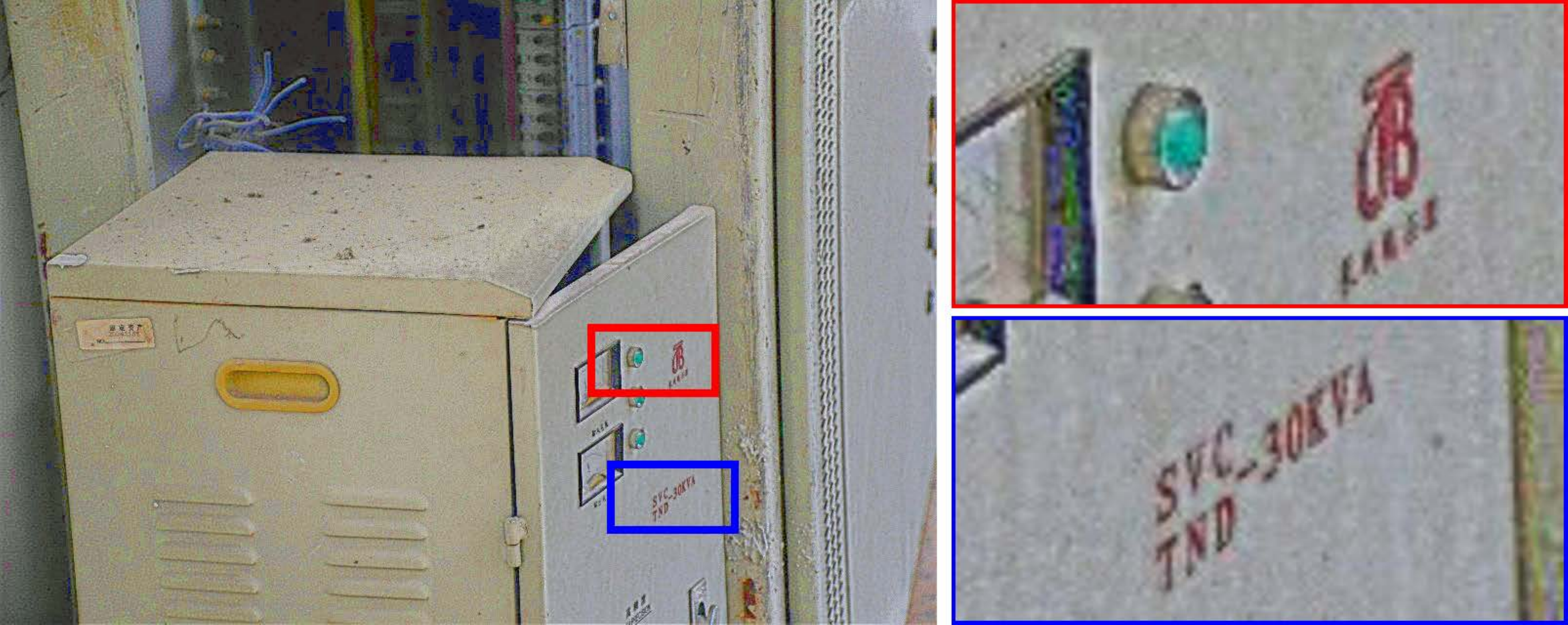}&
		\includegraphics[width=0.192\textwidth]{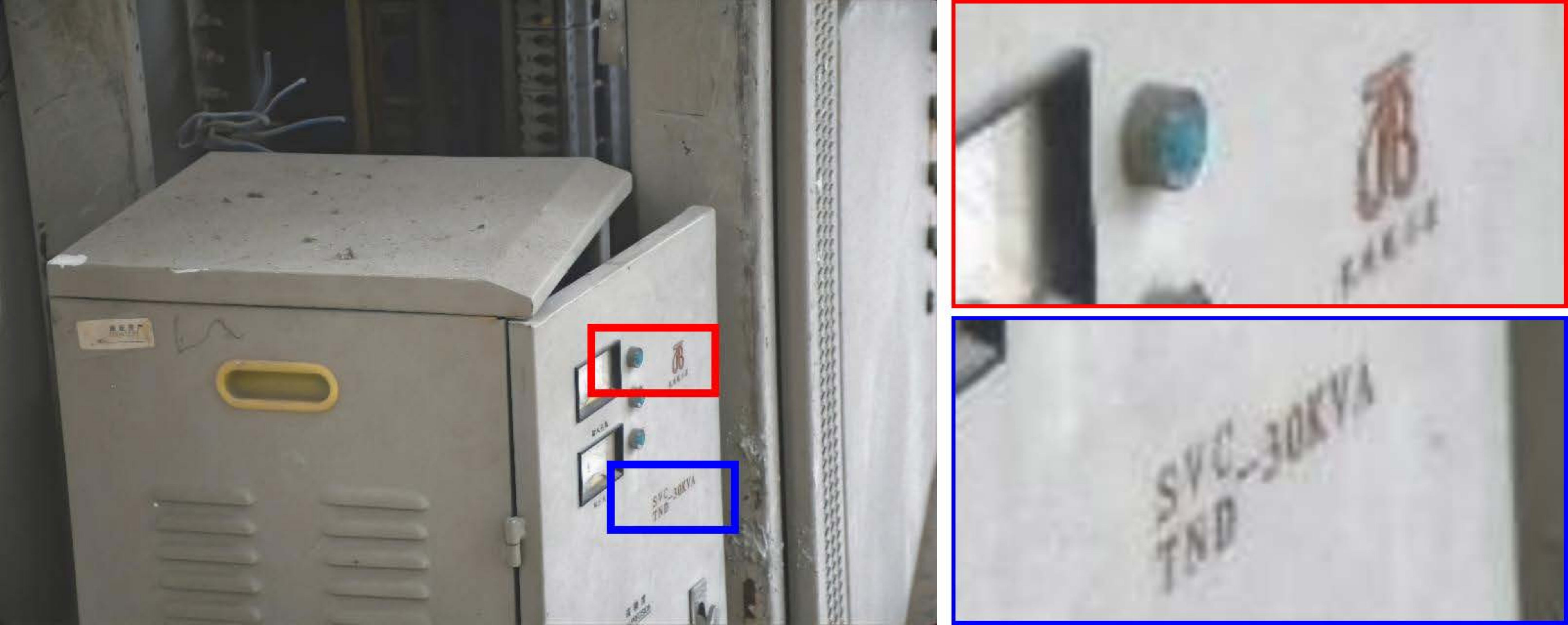}&
		\includegraphics[width=0.192\textwidth]{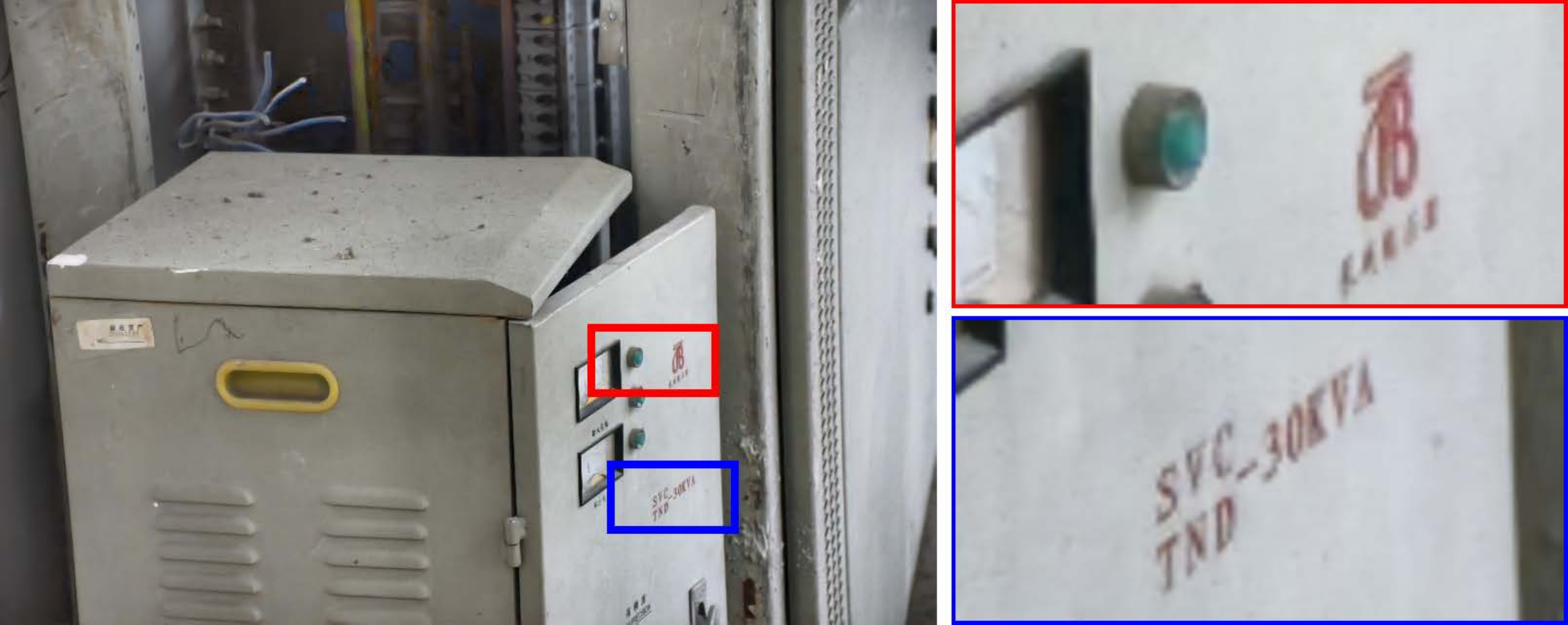}&
		\includegraphics[width=0.192\textwidth]{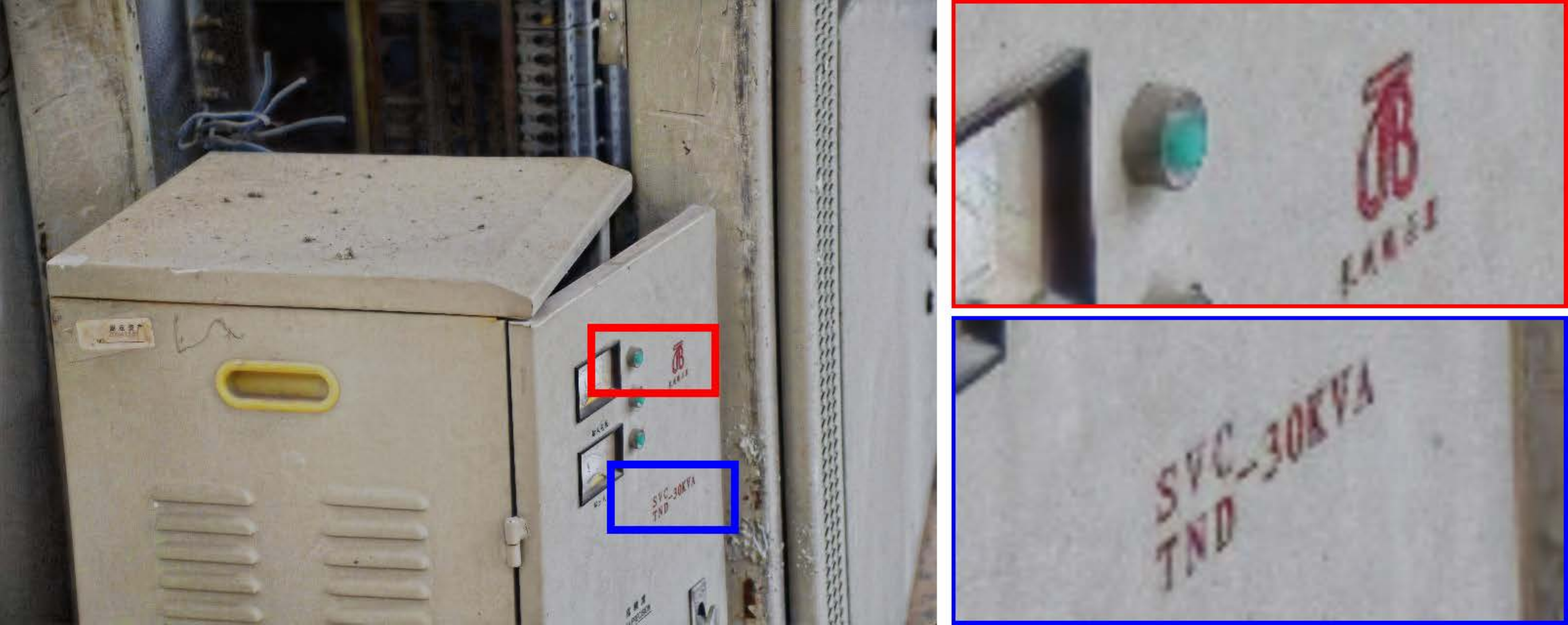}\\
		\footnotesize Input &\footnotesize RetinexNet&\footnotesize FIDE&\footnotesize DRBN&\footnotesize KinD\\
		\includegraphics[width=0.192\textwidth]{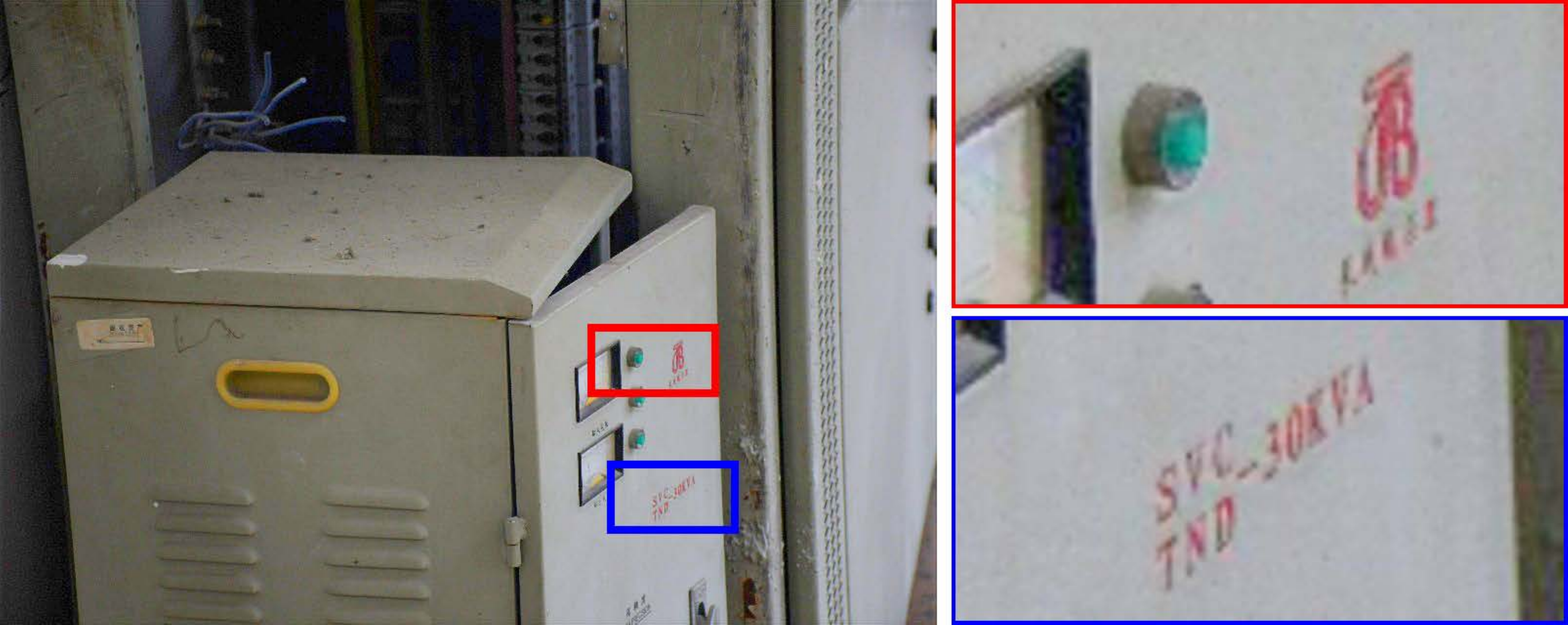}&
		\includegraphics[width=0.192\textwidth]{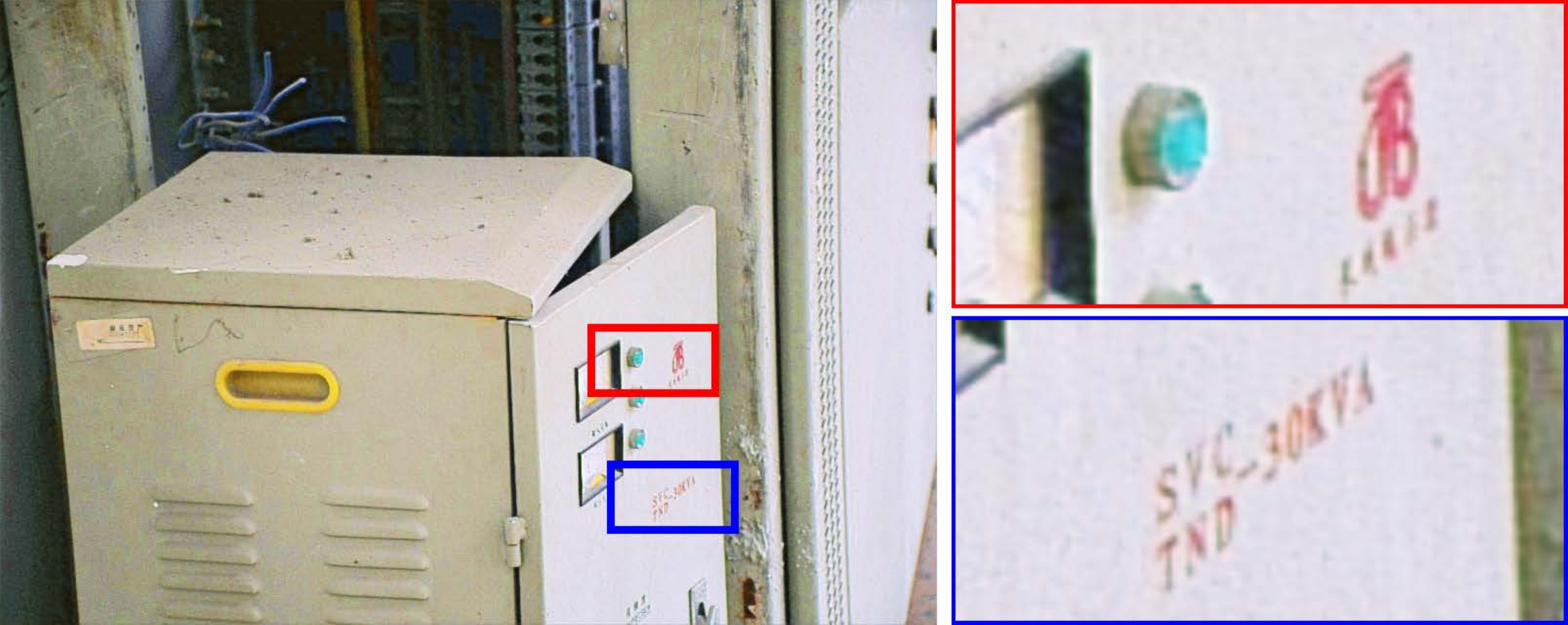}&
		\includegraphics[width=0.192\textwidth]{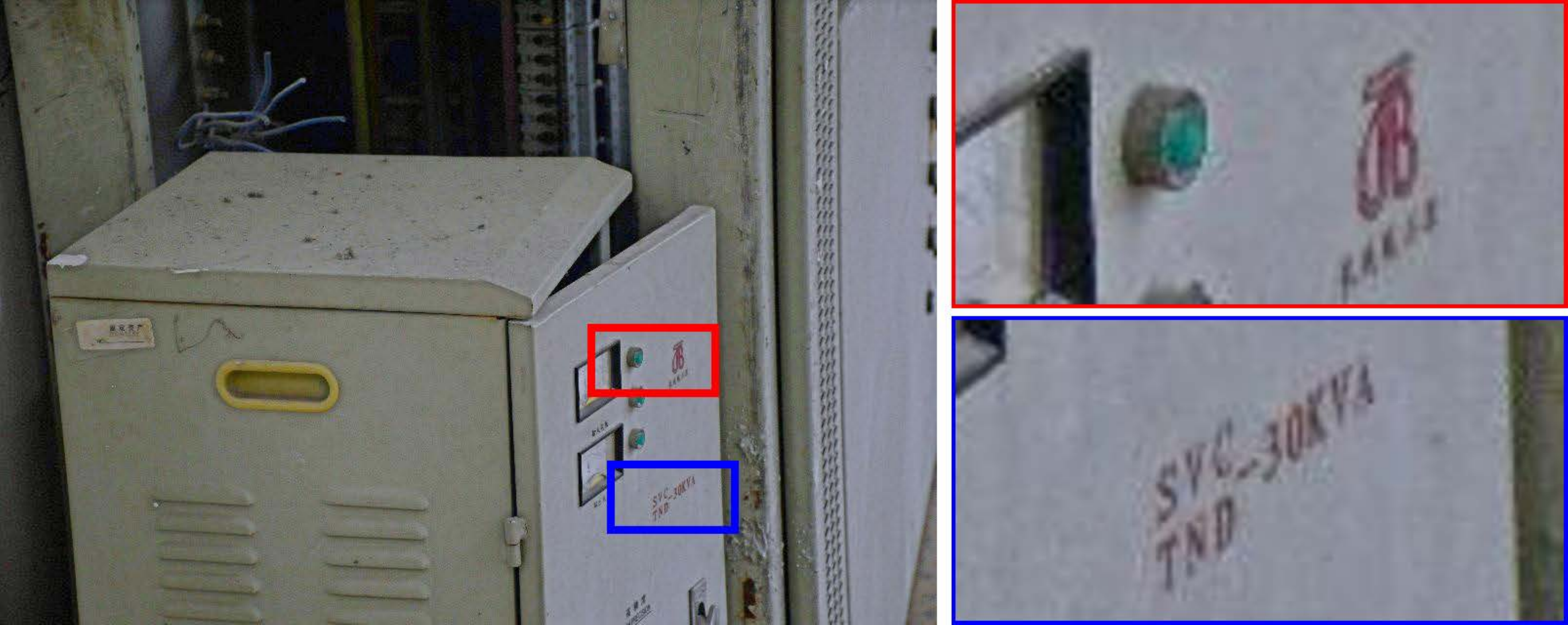}&
		\includegraphics[width=0.192\textwidth]{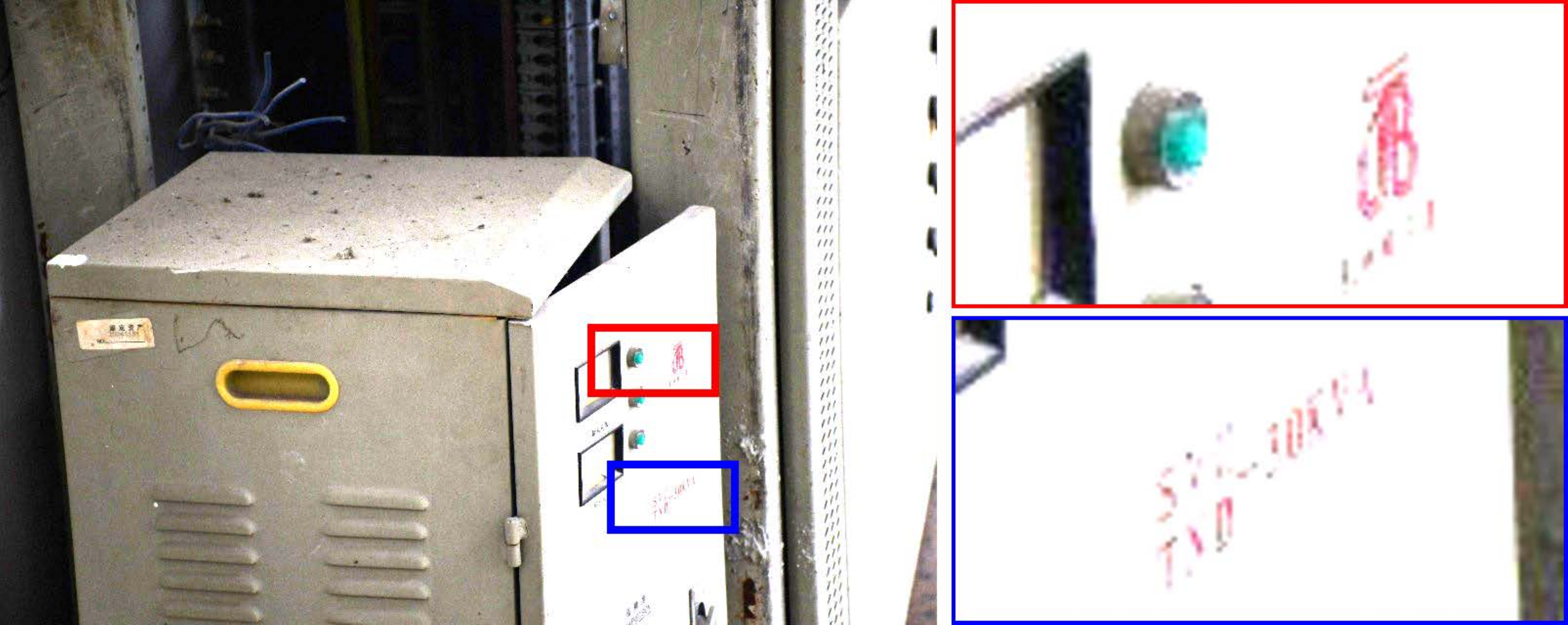}&
		\includegraphics[width=0.192\textwidth]{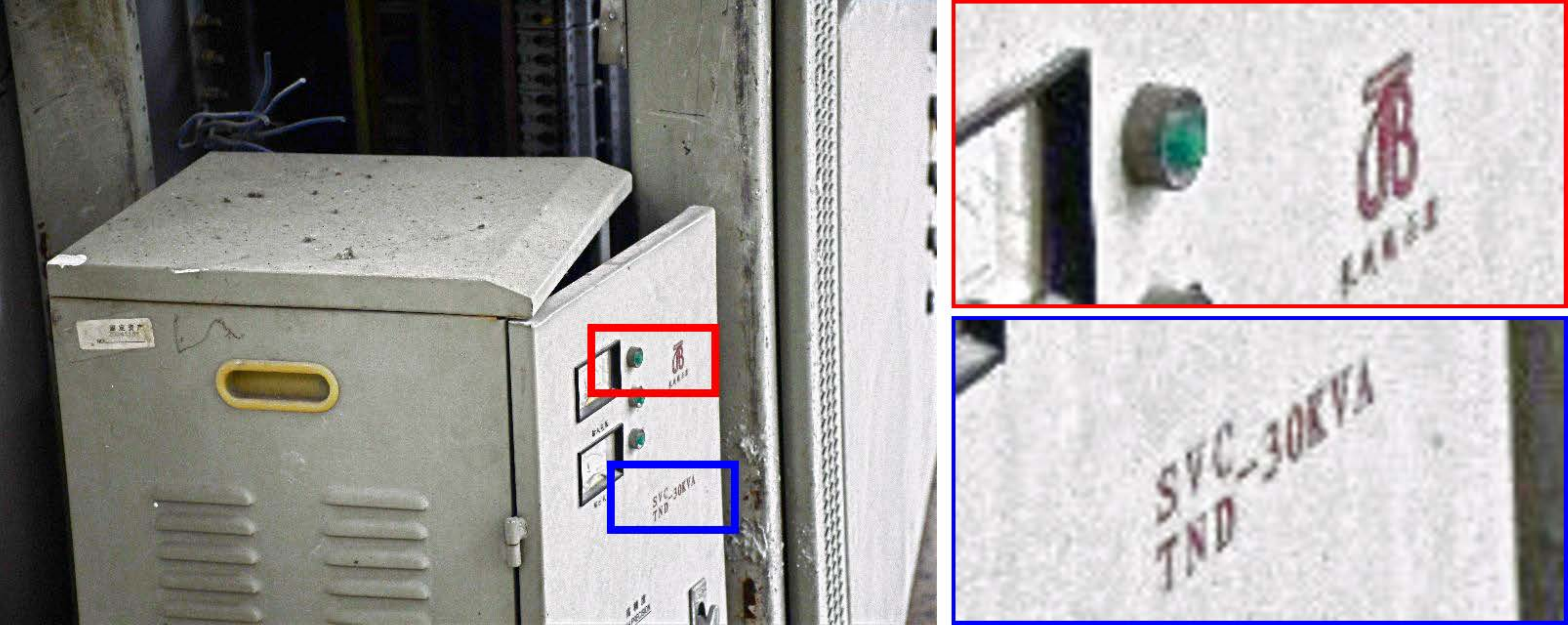}\\
		\footnotesize EnGAN &\footnotesize SSIENet&\footnotesize ZeroDCE &\footnotesize RUAS&\footnotesize Ours\\
	\end{tabular}
	\vspace{-0.3cm}
	\caption{Visual comparison on the LSRW dataset among state-of-the-art low-light image enhancement approaches. }
	\label{fig:LSRW}
	\vspace{-0.3cm}
\end{figure*}

\begin{figure*}[t]
	\centering
	\begin{tabular}{c@{\extracolsep{0.25em}}c@{\extracolsep{0.25em}}c@{\extracolsep{0.25em}}c@{\extracolsep{0.25em}}c@{\extracolsep{0.25em}}c@{\extracolsep{0.25em}}c@{\extracolsep{0.25em}}c} 
		\includegraphics[width=0.116\textwidth]{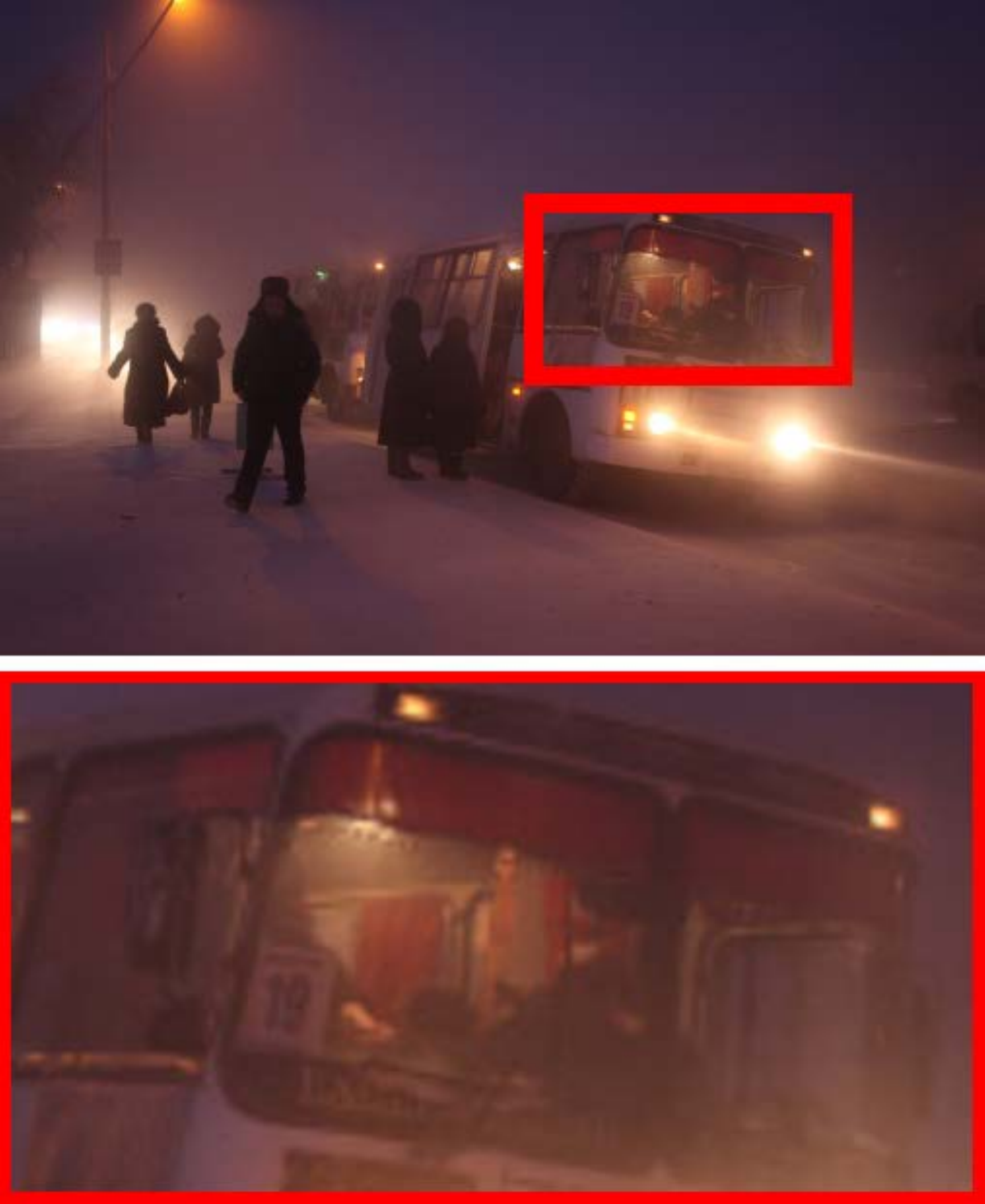}&
		\includegraphics[width=0.116\textwidth]{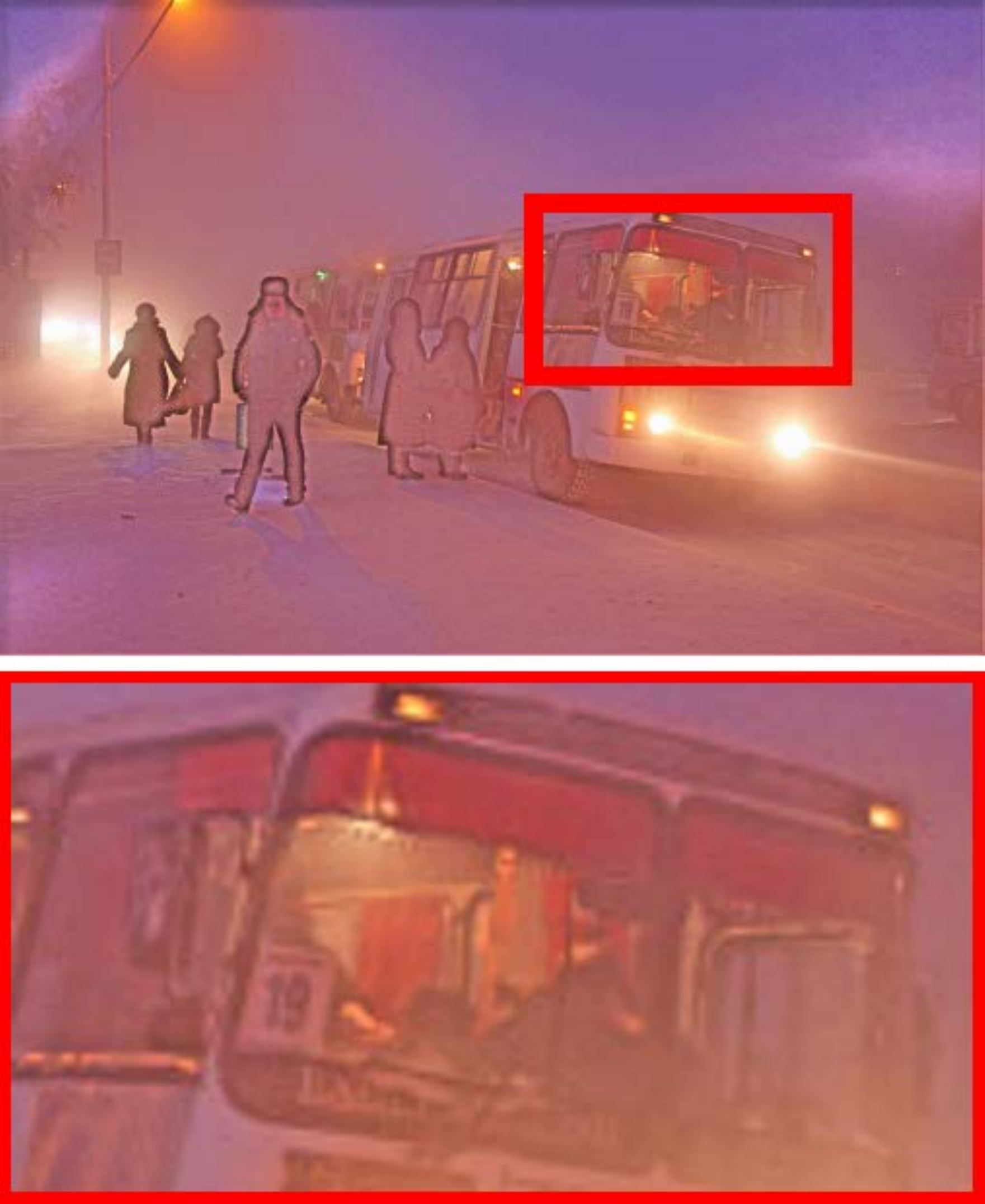}&
		\includegraphics[width=0.116\textwidth]{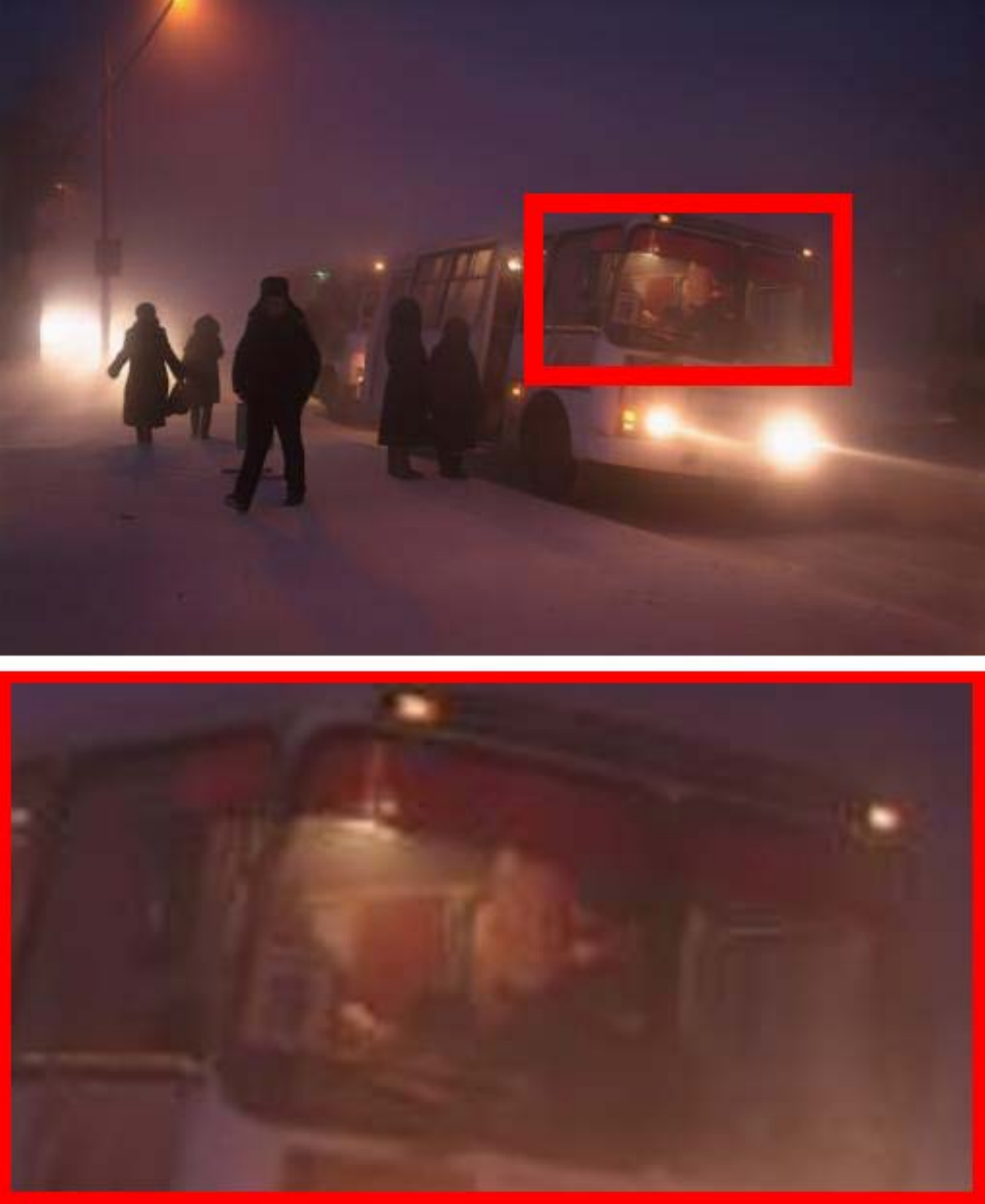}&
		\includegraphics[width=0.116\textwidth]{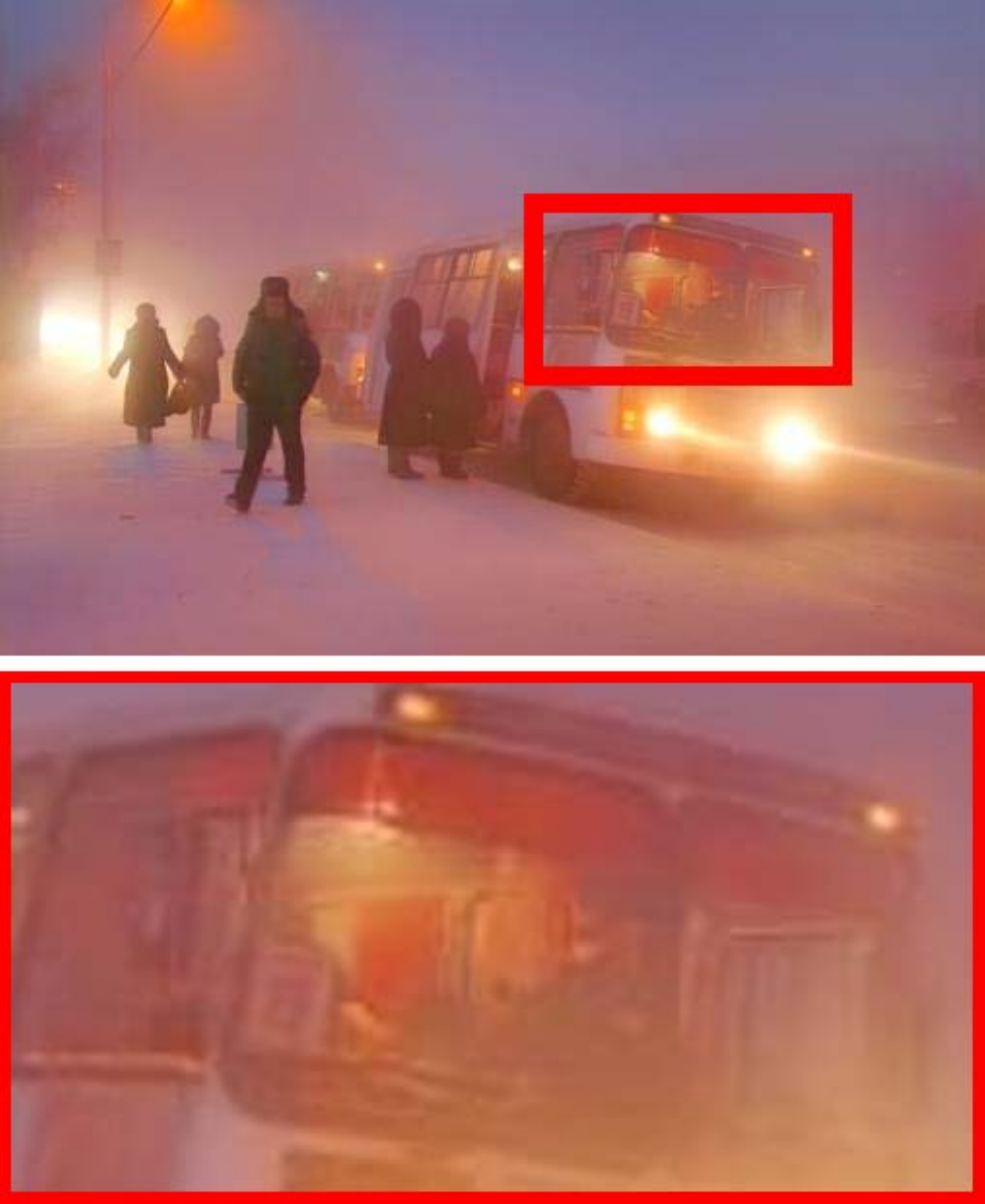}&
		\includegraphics[width=0.116\textwidth]{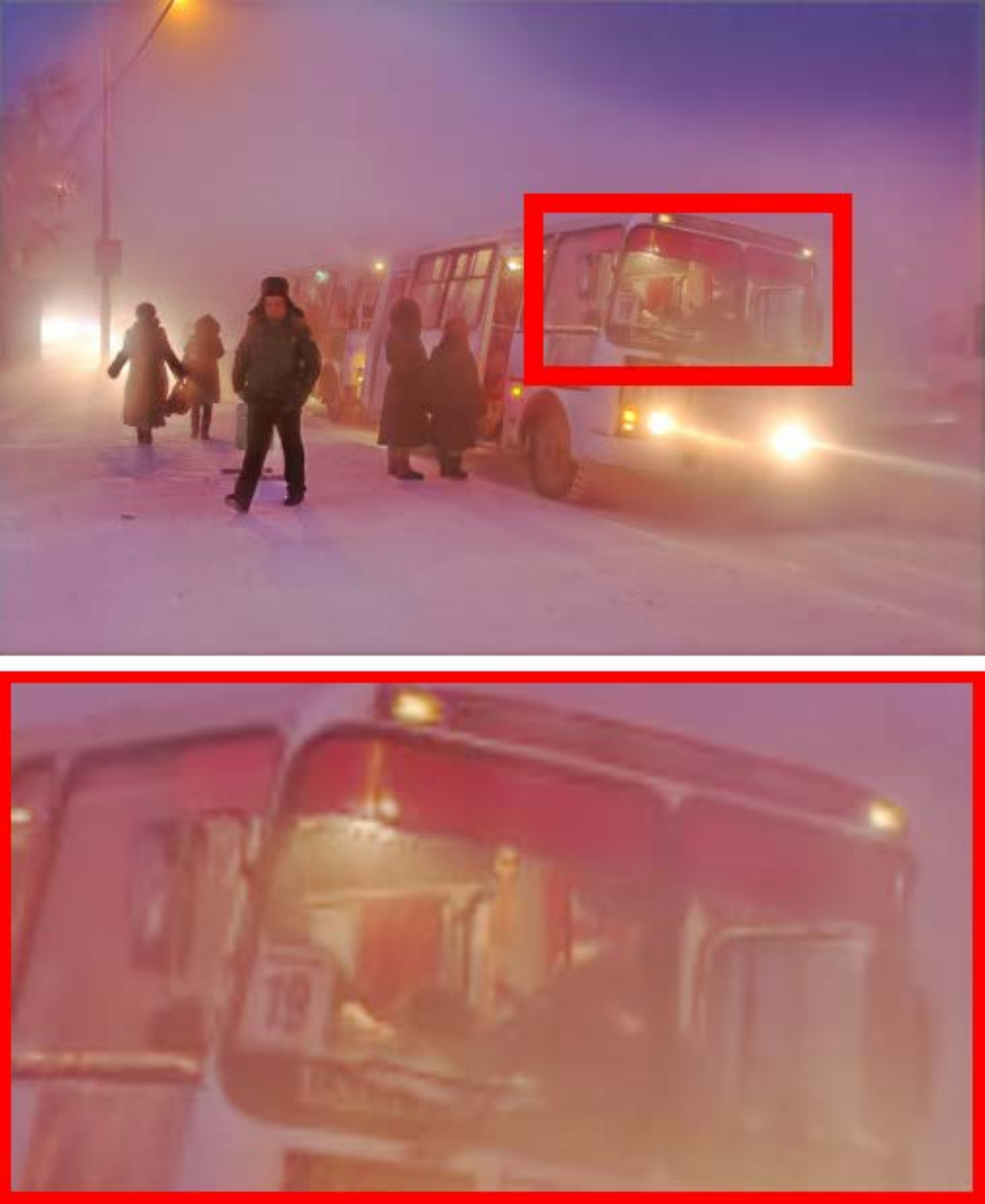}&
		\includegraphics[width=0.116\textwidth]{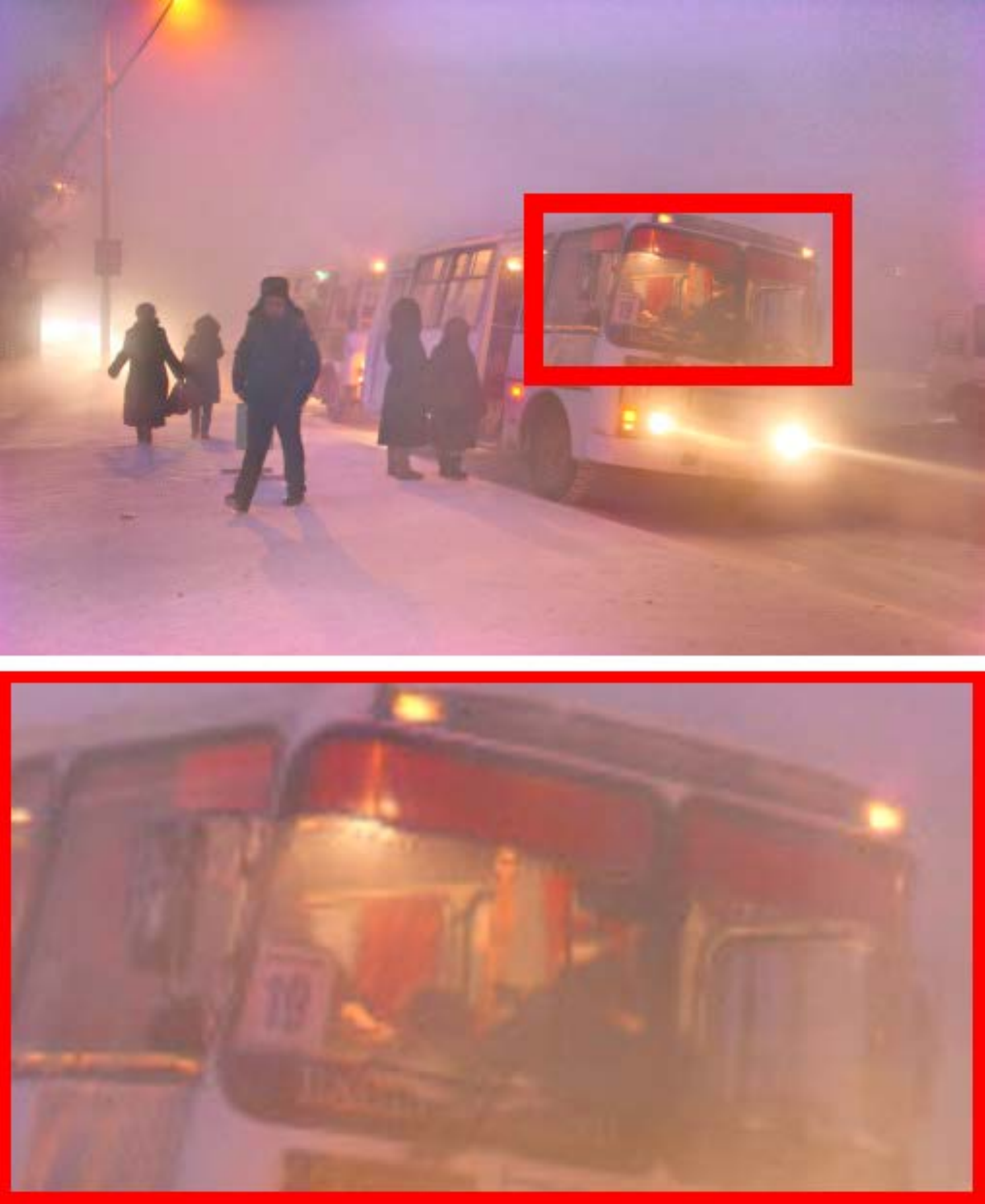}&
		\includegraphics[width=0.116\textwidth]{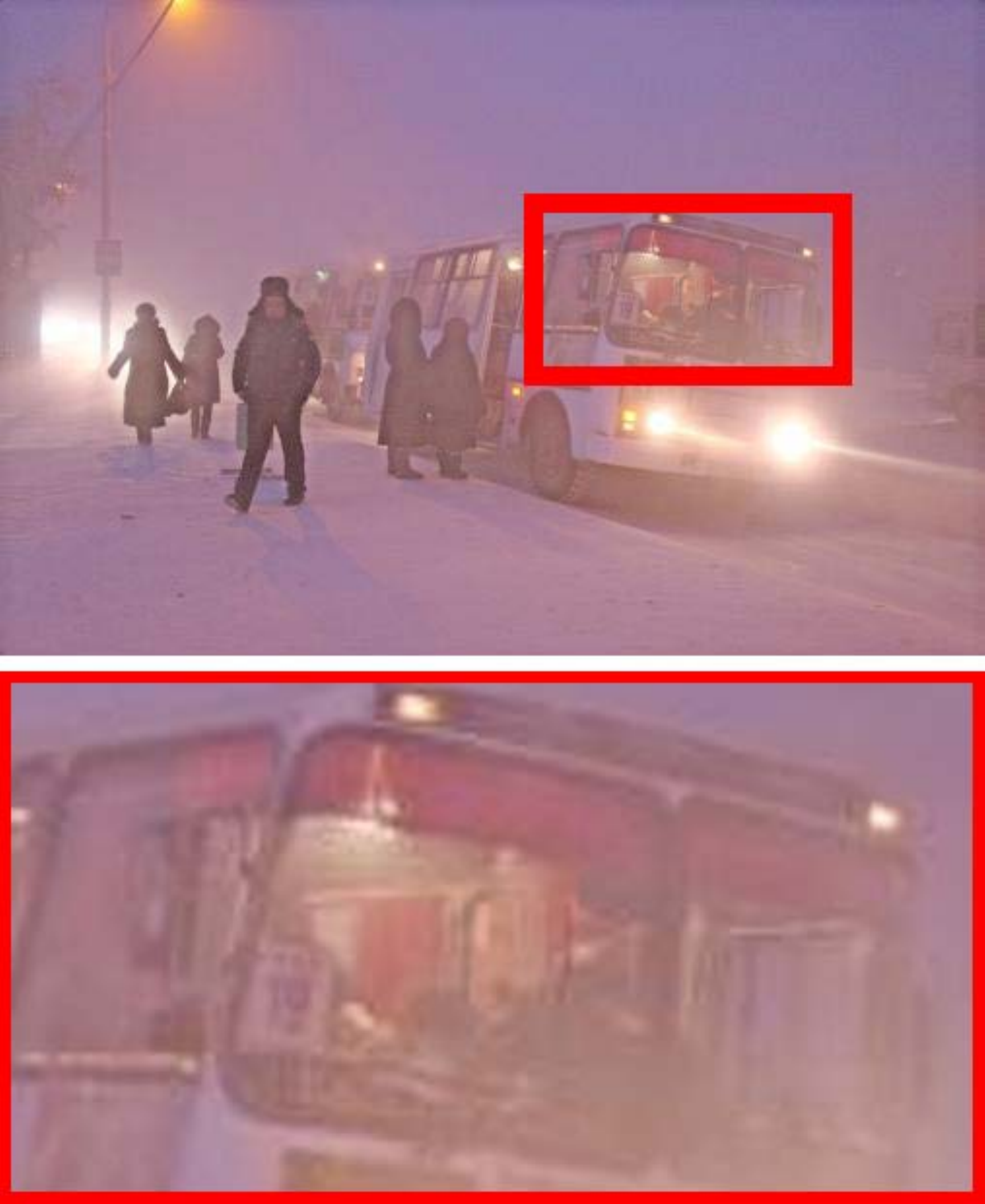}&
		\includegraphics[width=0.116\textwidth]{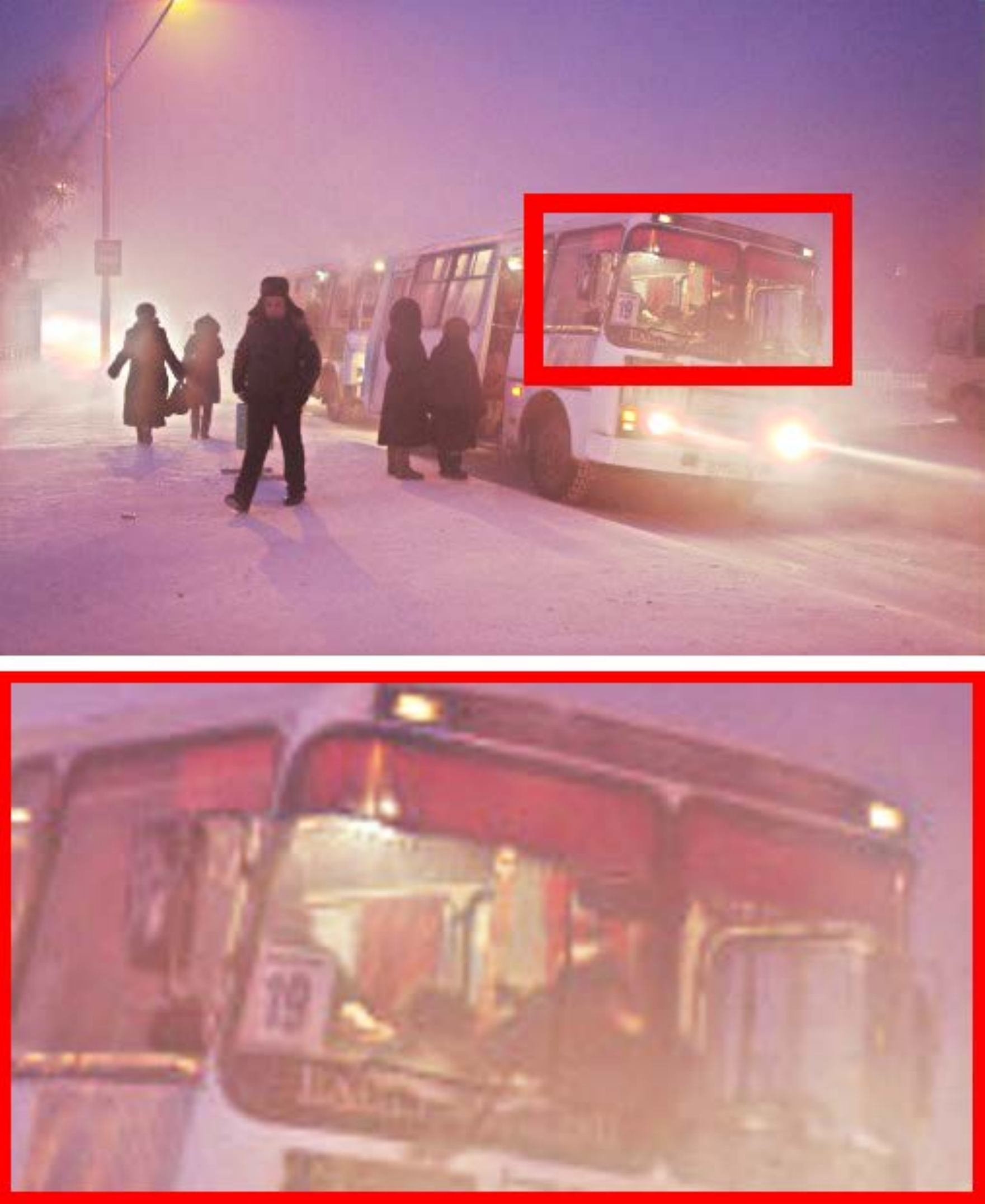}\\
		\includegraphics[width=0.116\textwidth]{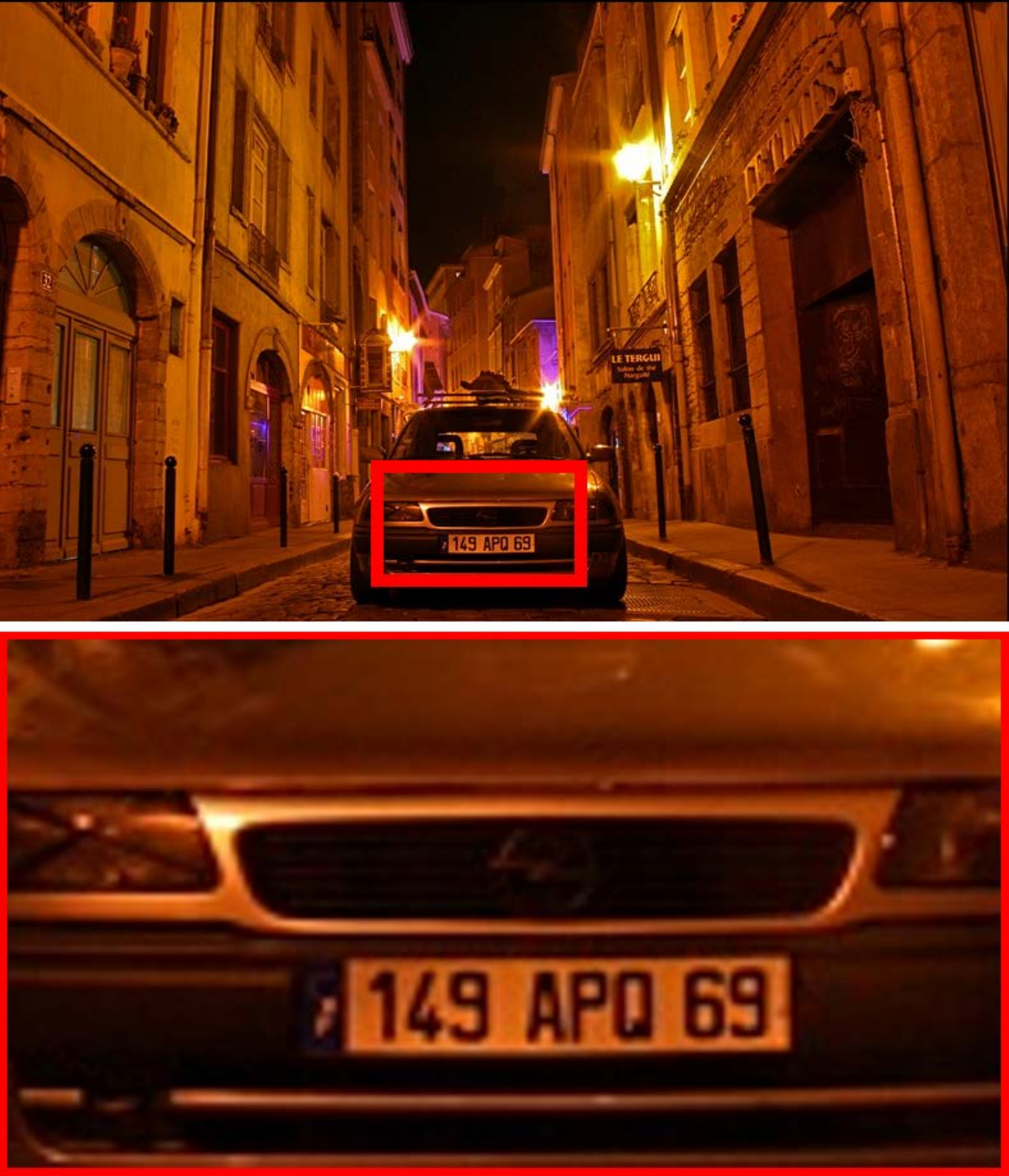}&
		\includegraphics[width=0.116\textwidth]{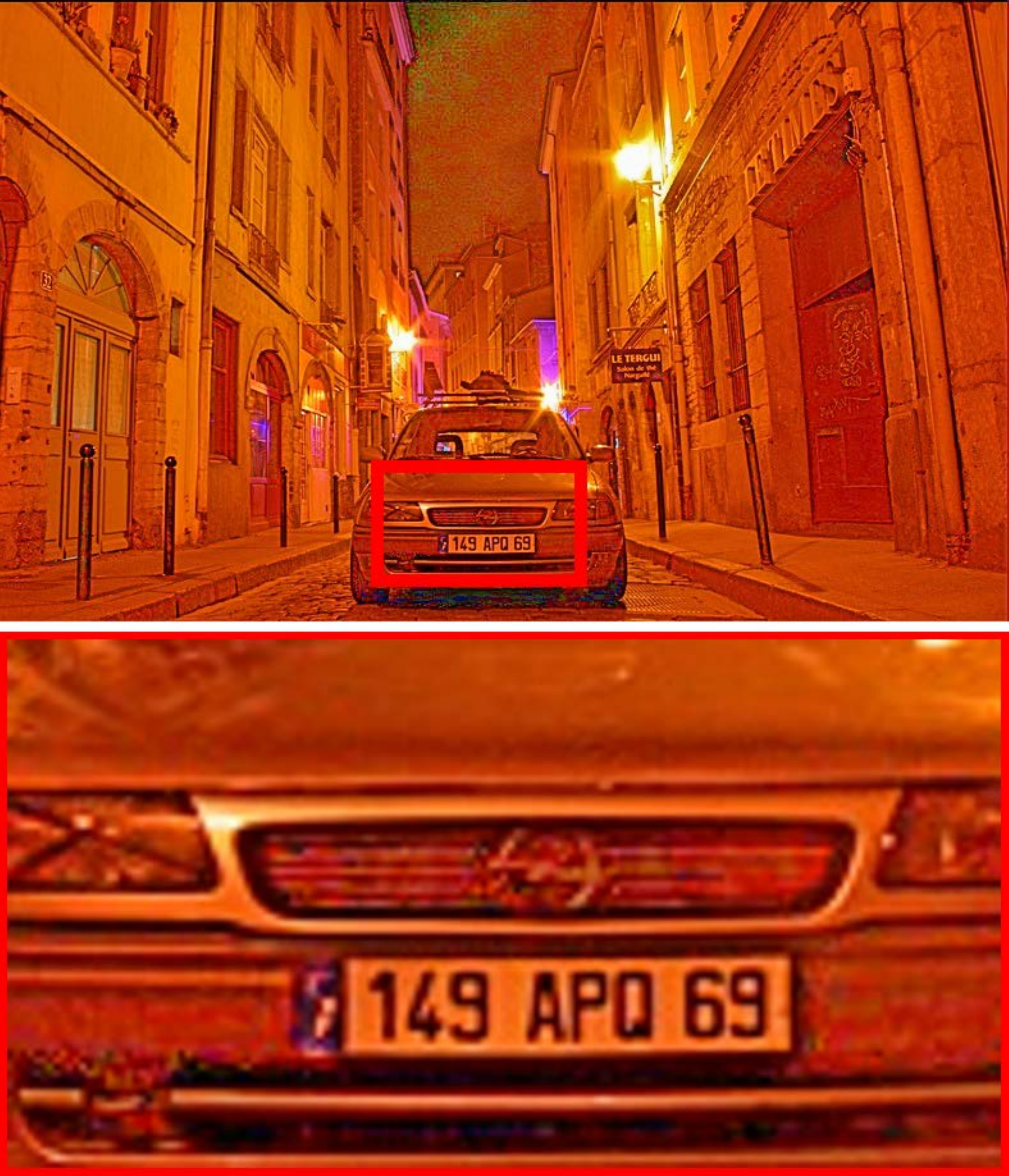}&
		\includegraphics[width=0.116\textwidth]{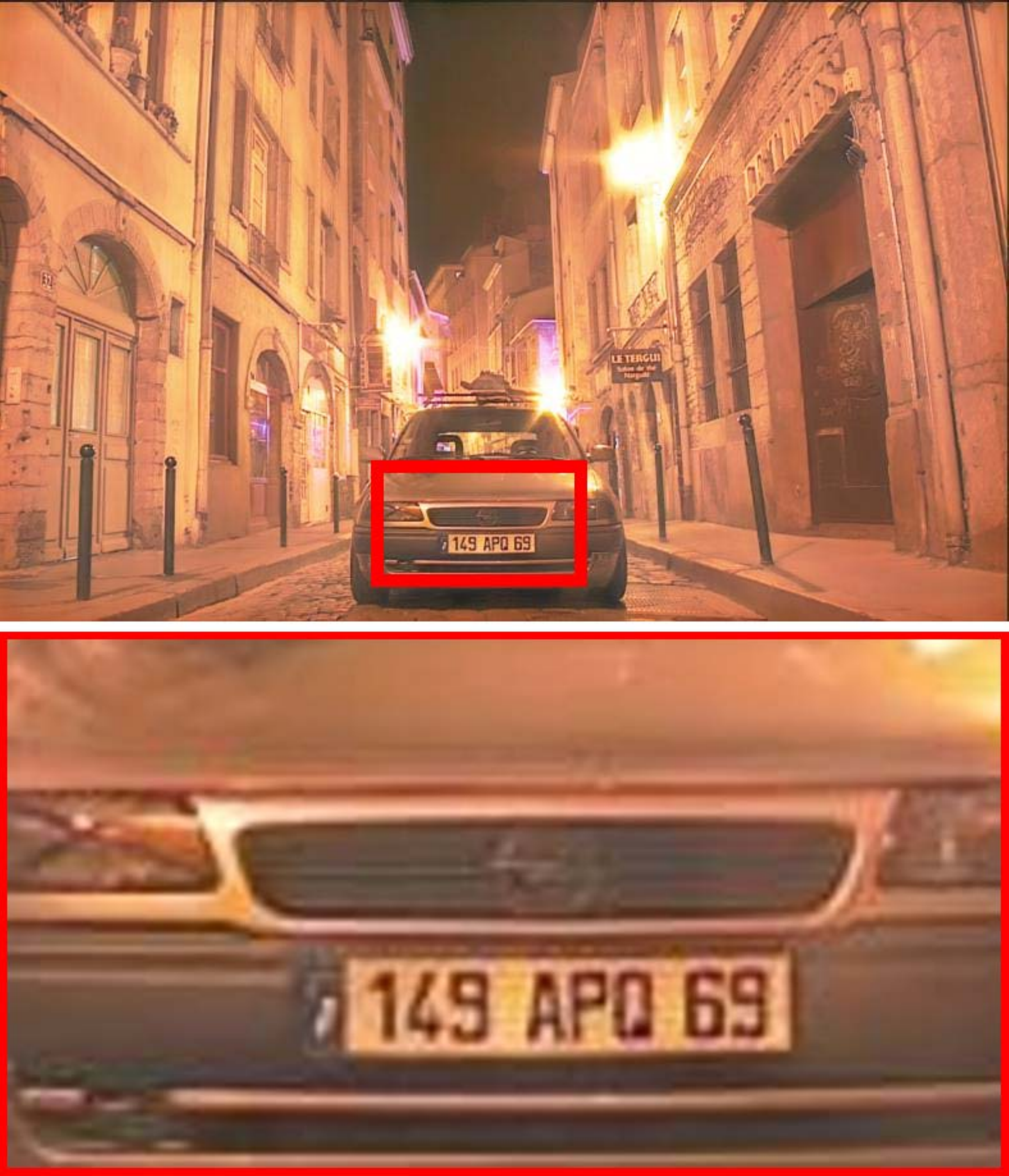}&
		\includegraphics[width=0.116\textwidth]{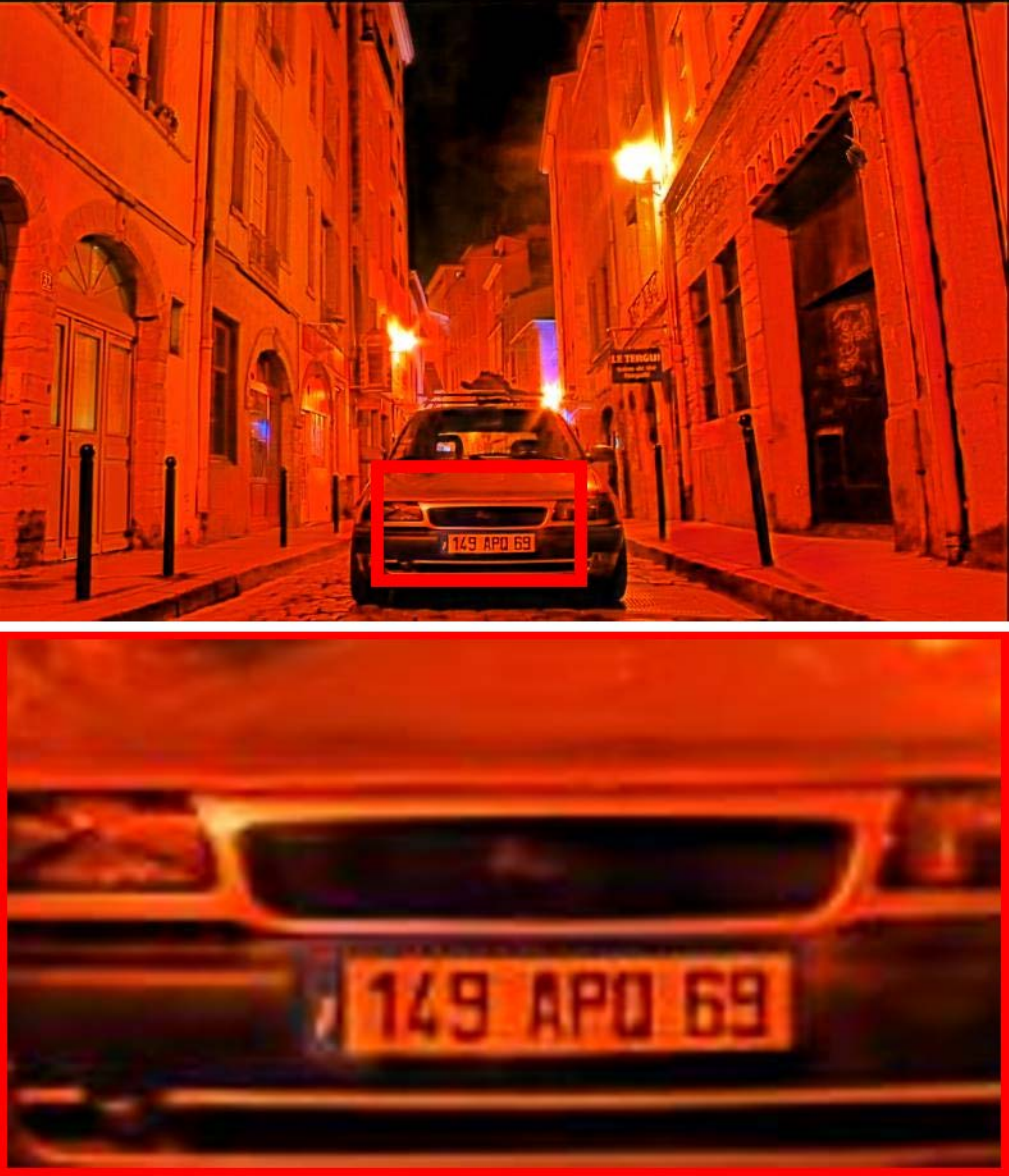}&
		\includegraphics[width=0.116\textwidth]{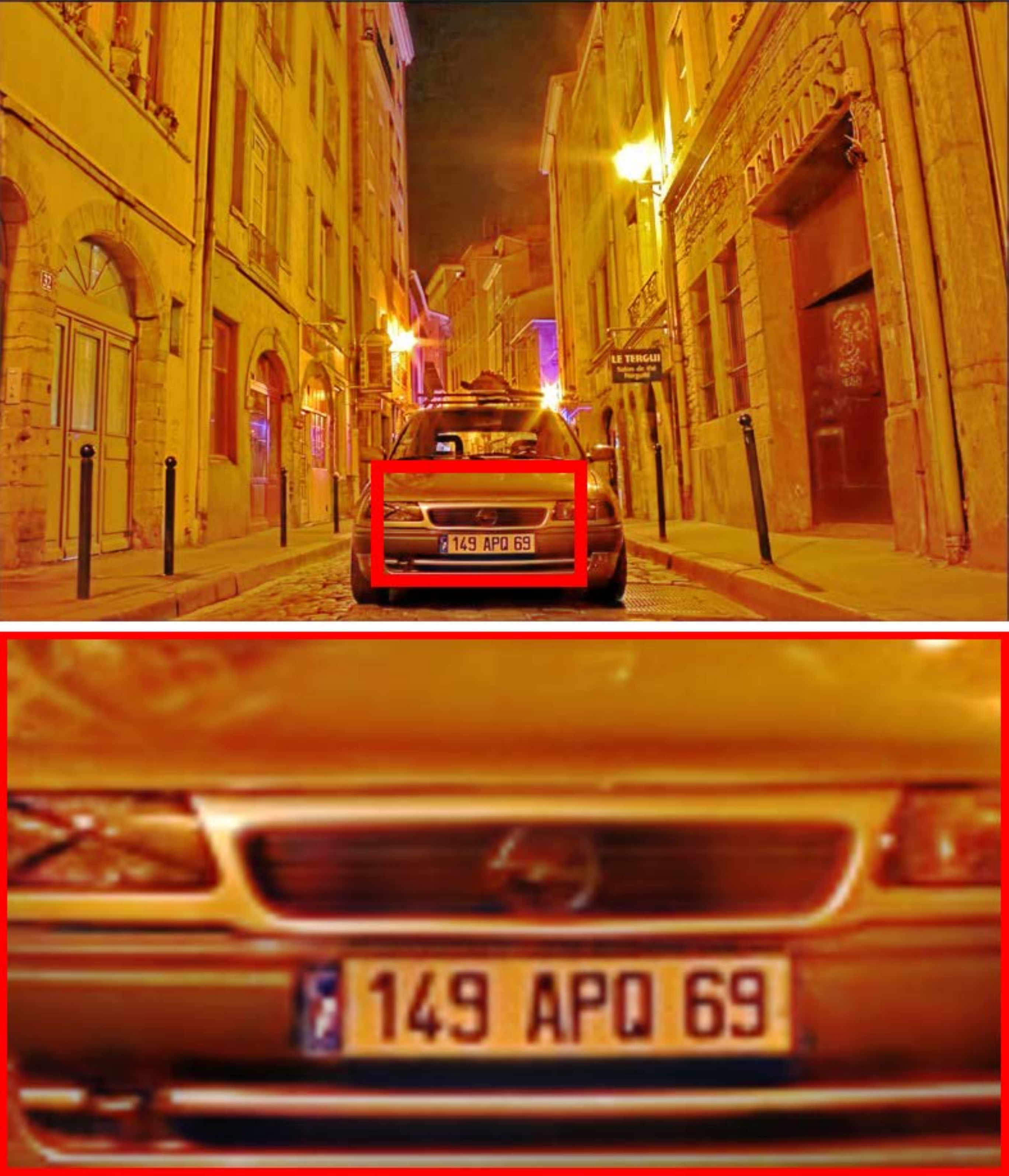}&
		\includegraphics[width=0.116\textwidth]{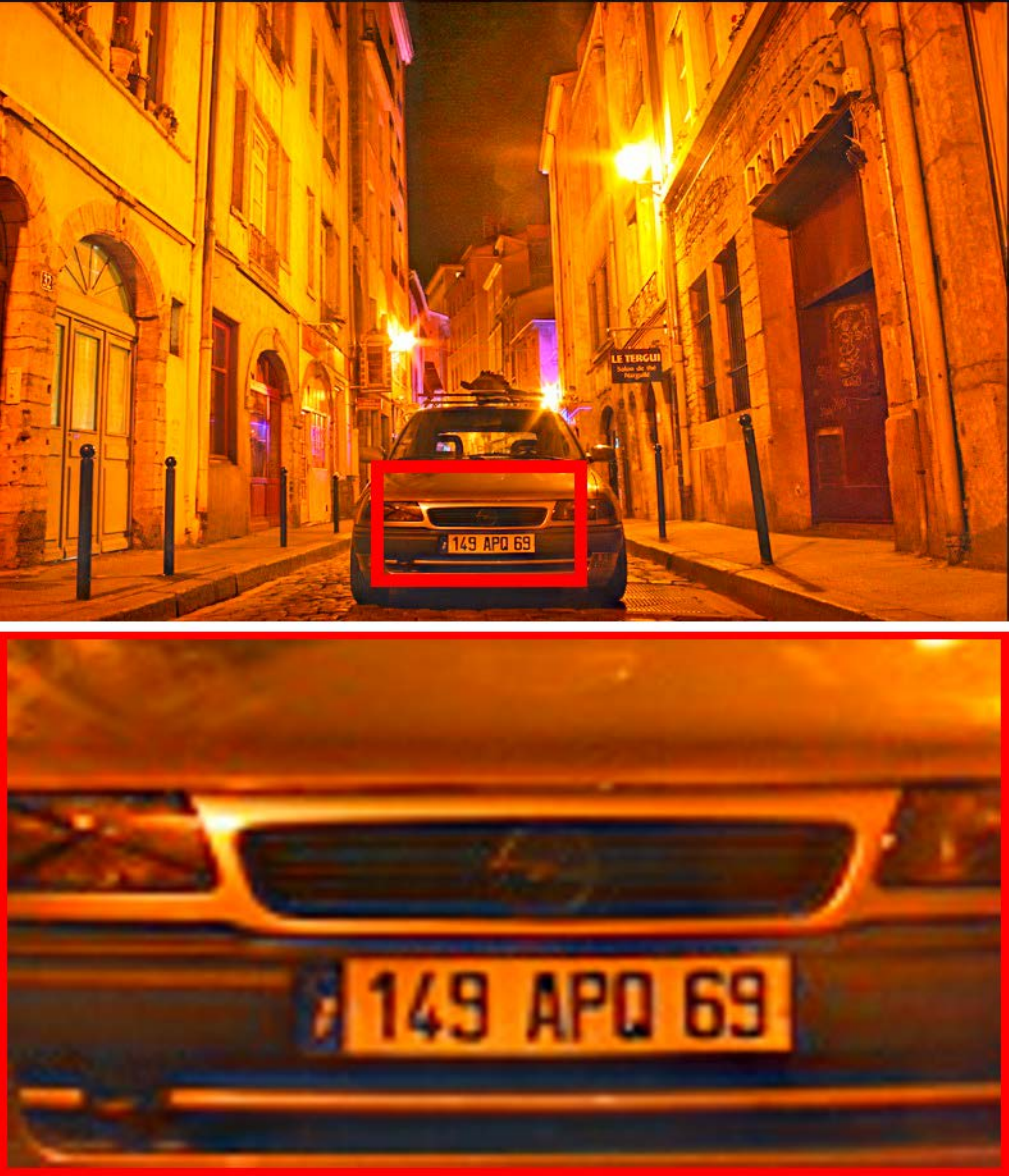}&
		\includegraphics[width=0.116\textwidth]{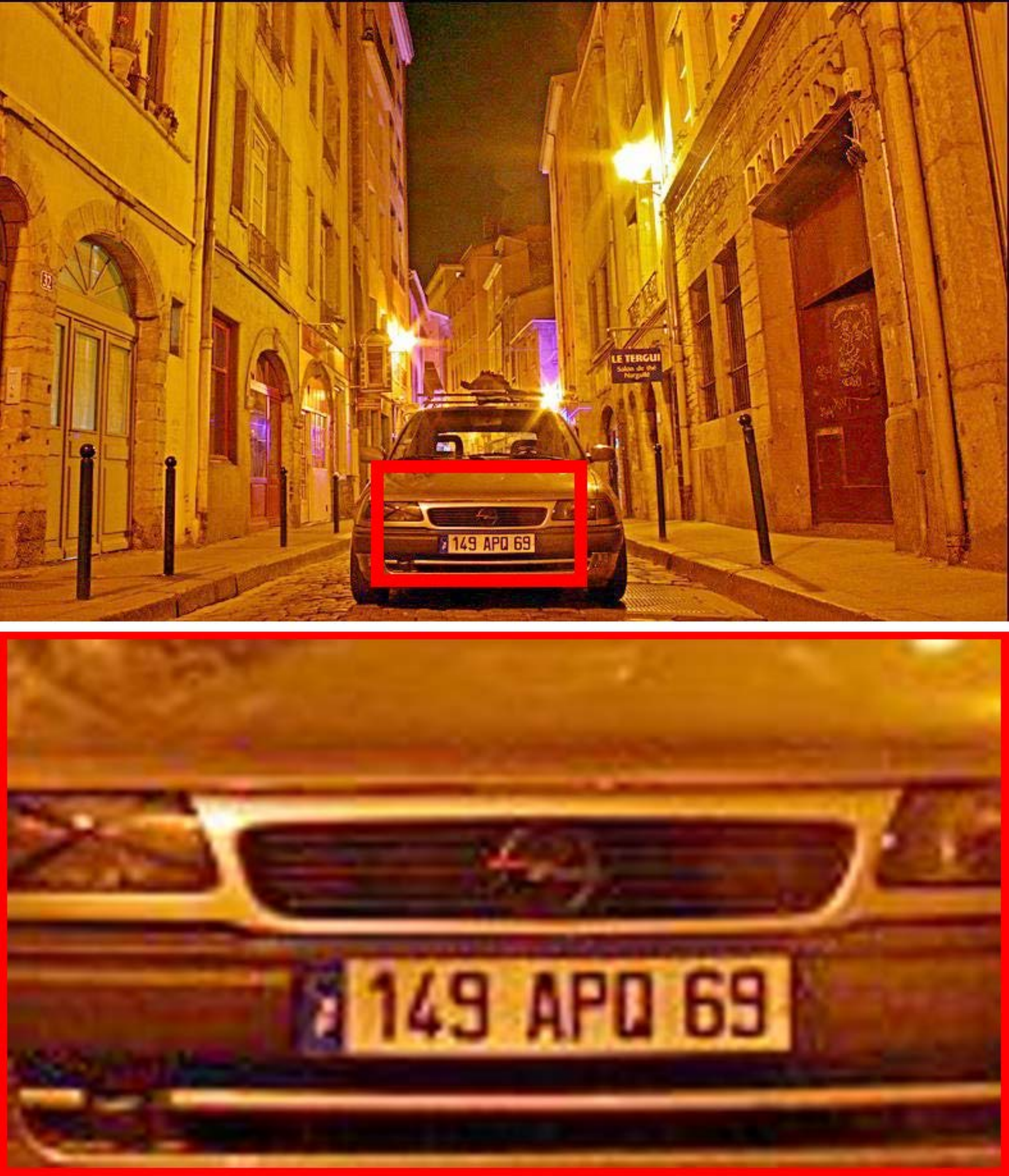}&
		\includegraphics[width=0.116\textwidth]{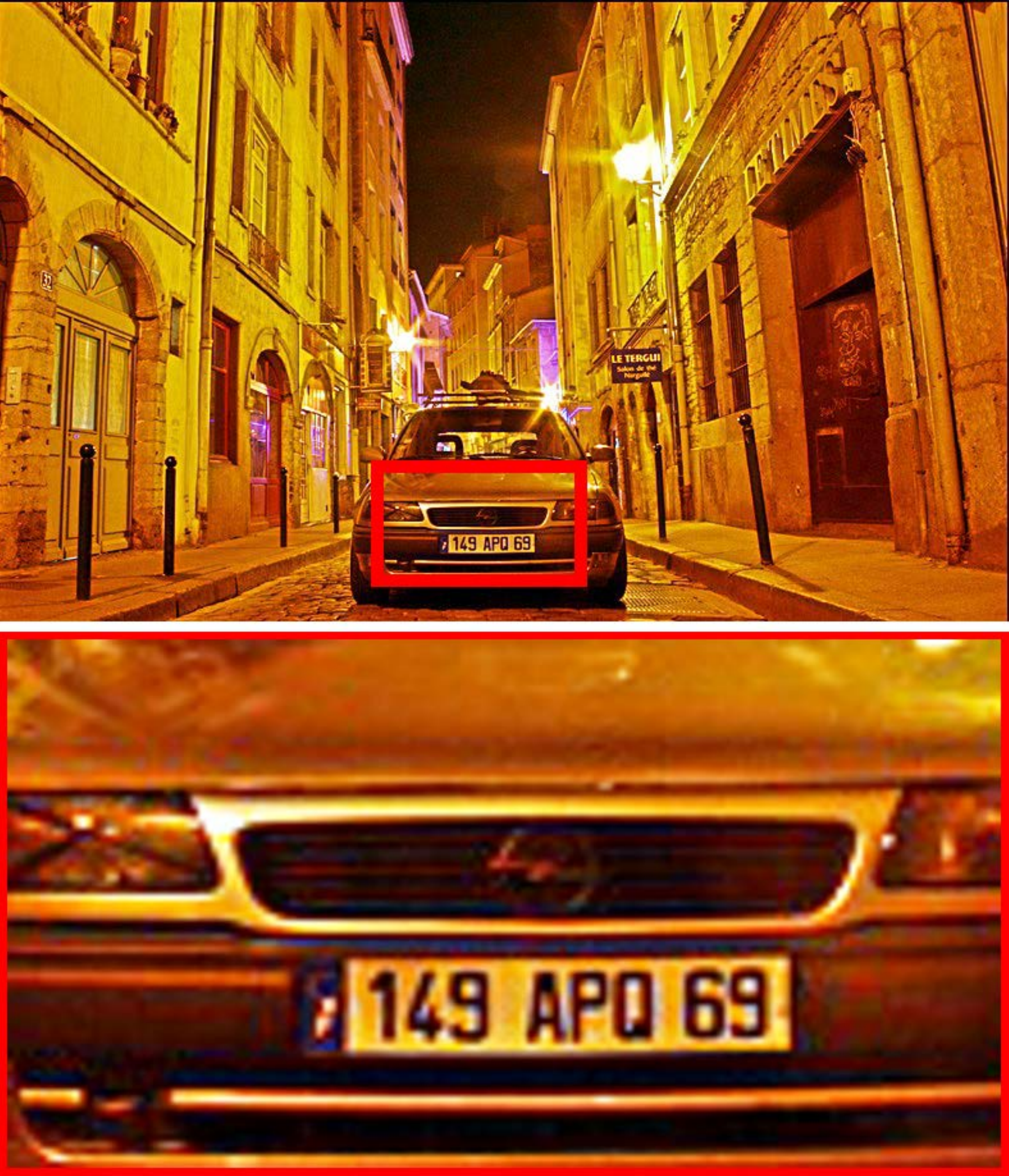}\\
		\includegraphics[width=0.116\textwidth]{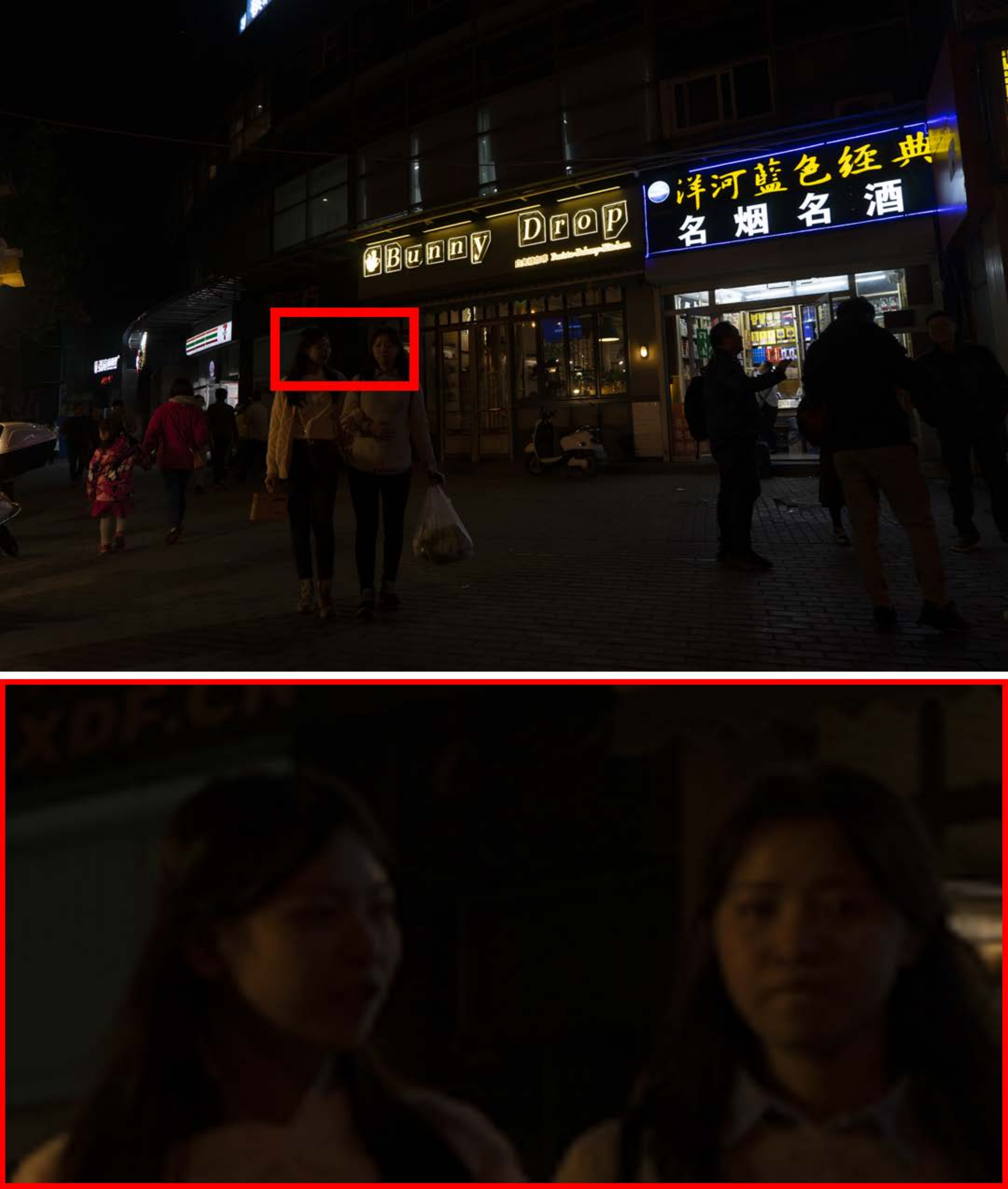}&
		\includegraphics[width=0.116\textwidth]{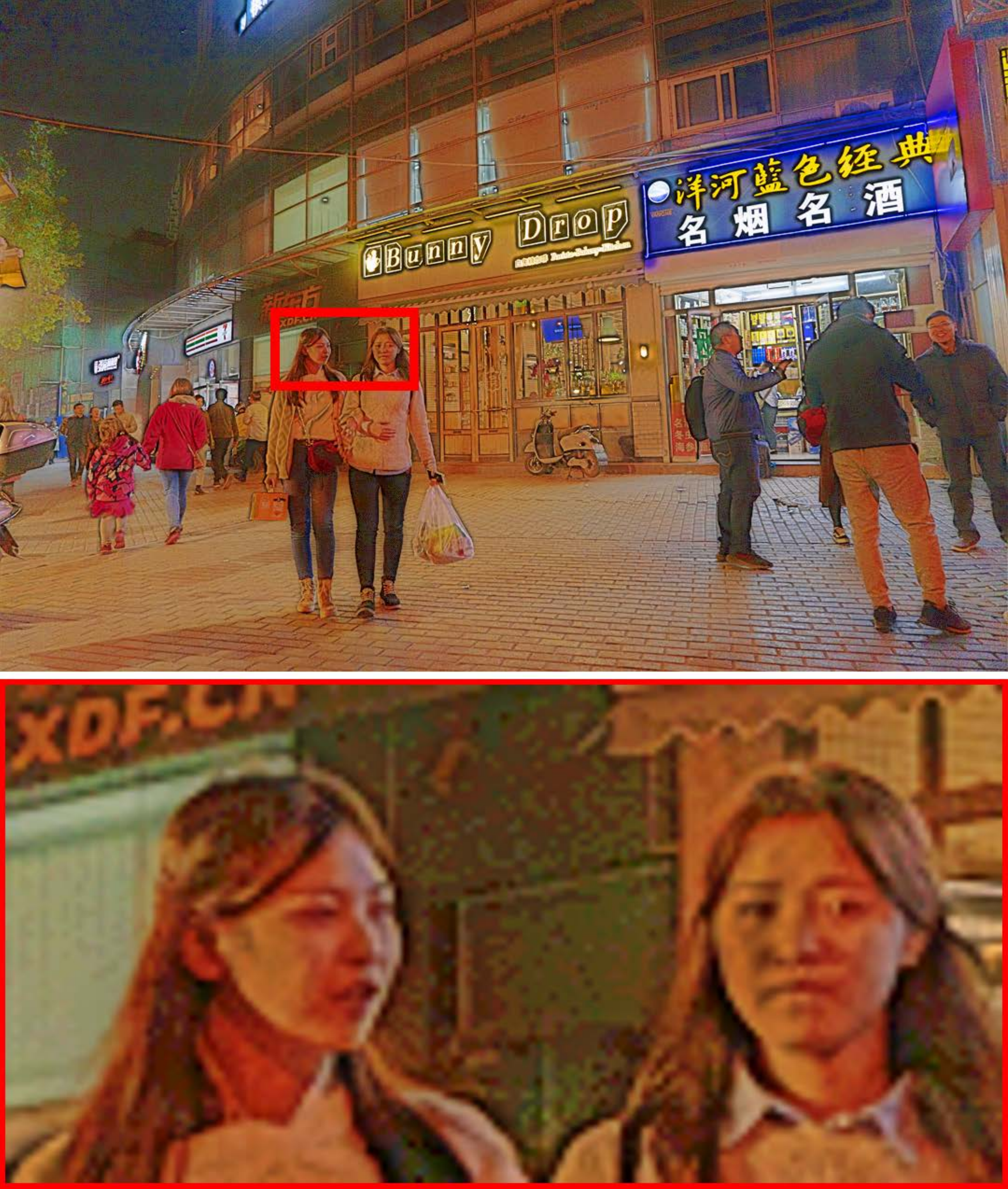}&
		\includegraphics[width=0.116\textwidth]{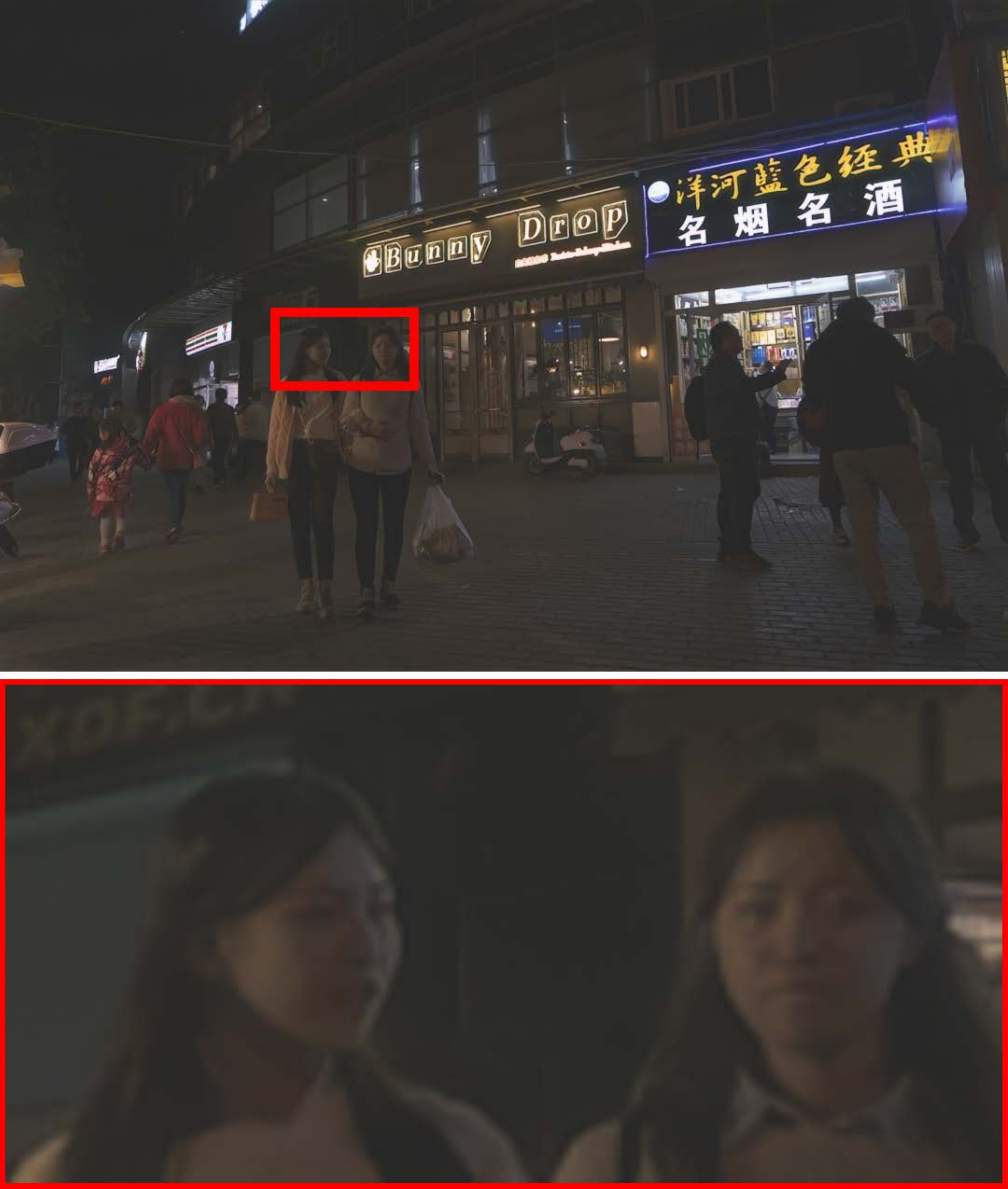}&
		\includegraphics[width=0.116\textwidth]{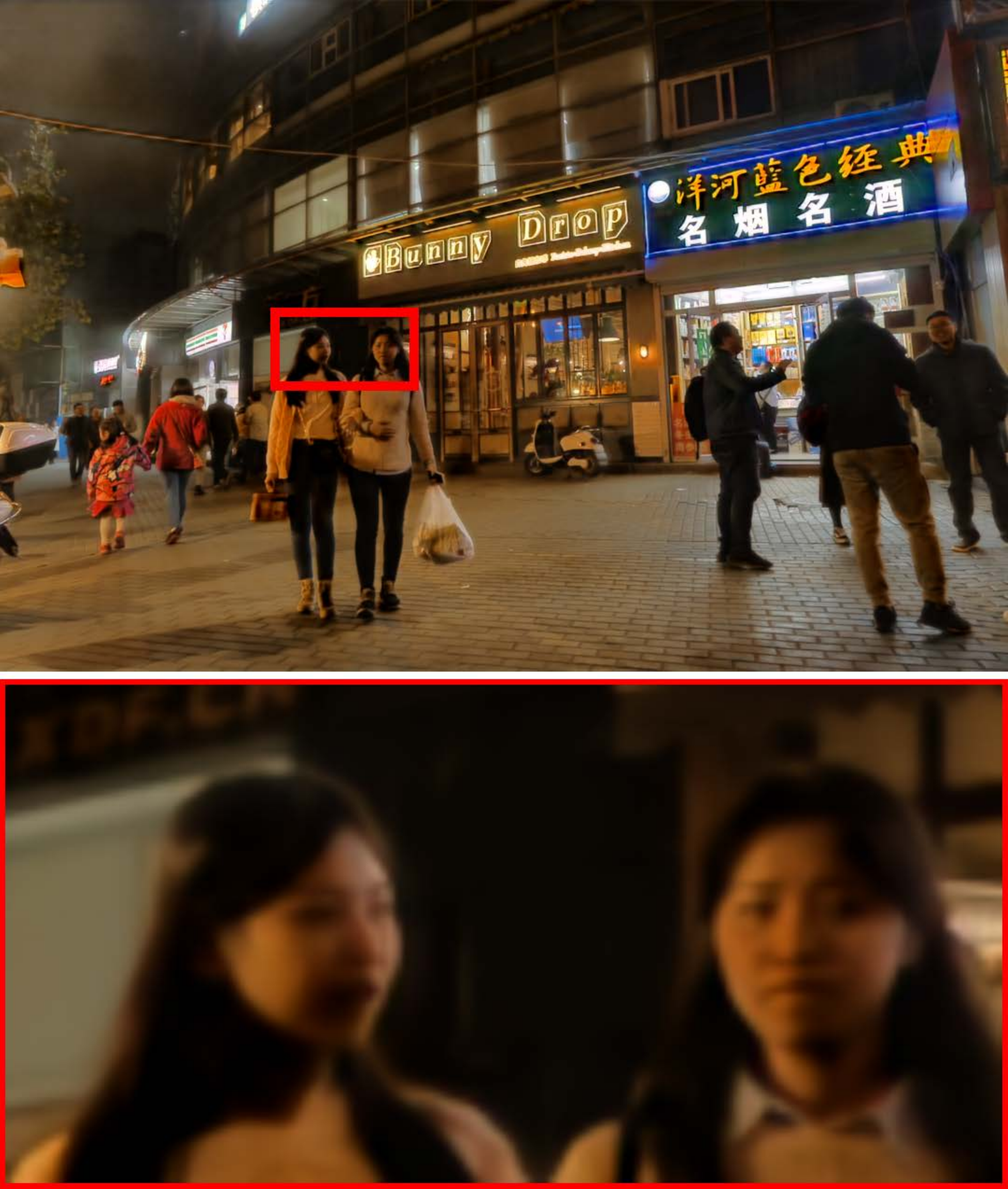}&
		\includegraphics[width=0.116\textwidth]{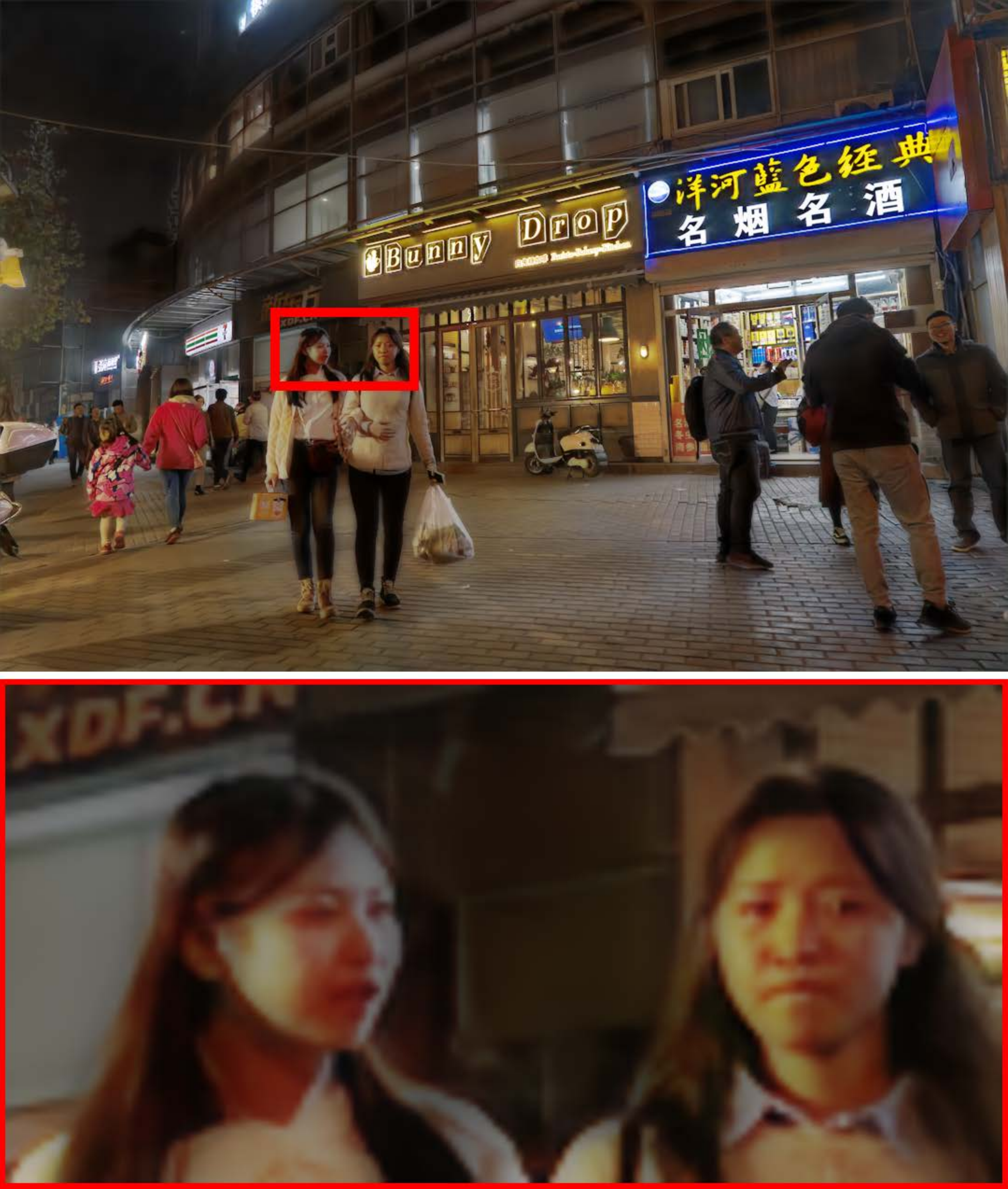}&
		\includegraphics[width=0.116\textwidth]{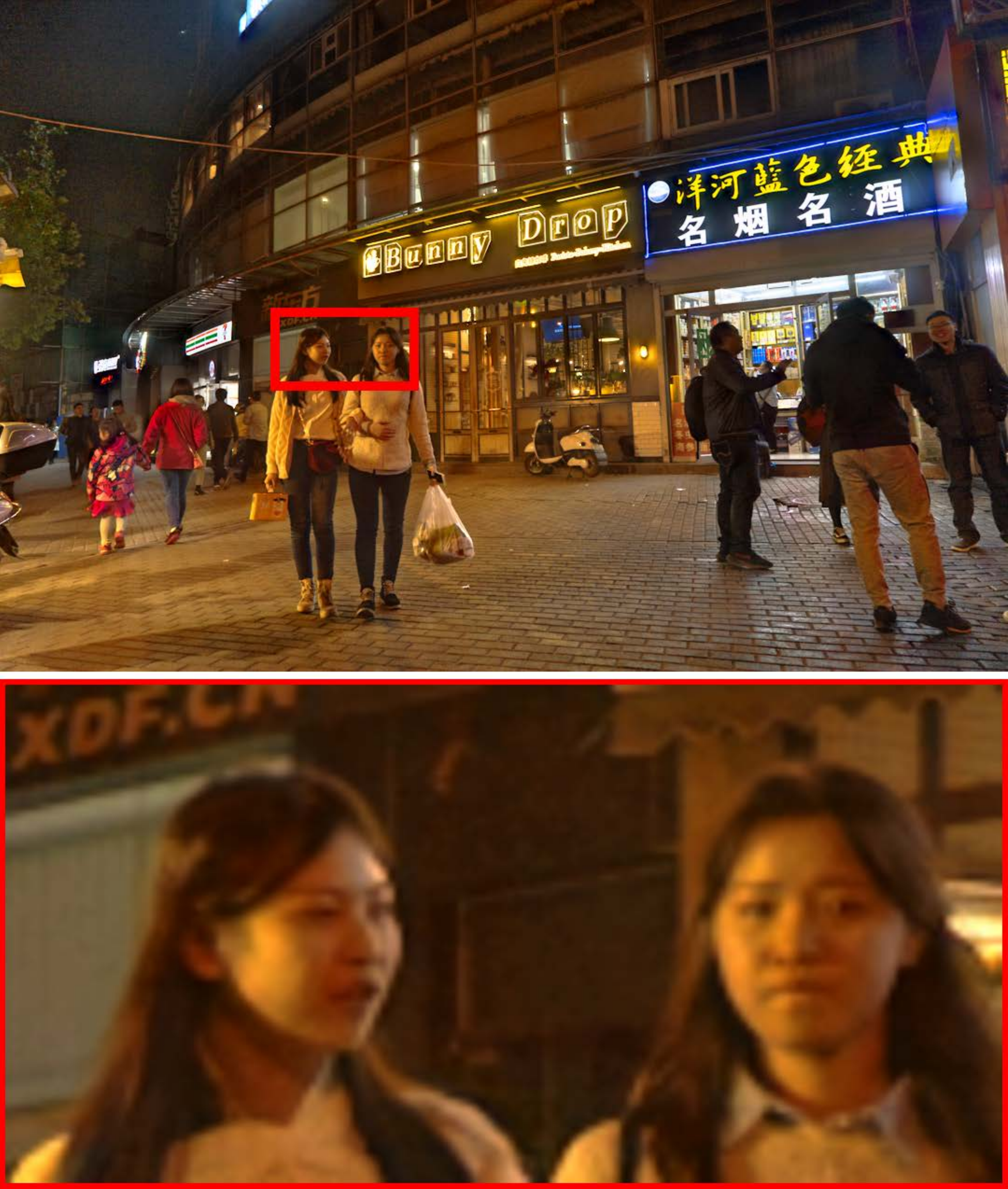}&
		\includegraphics[width=0.116\textwidth]{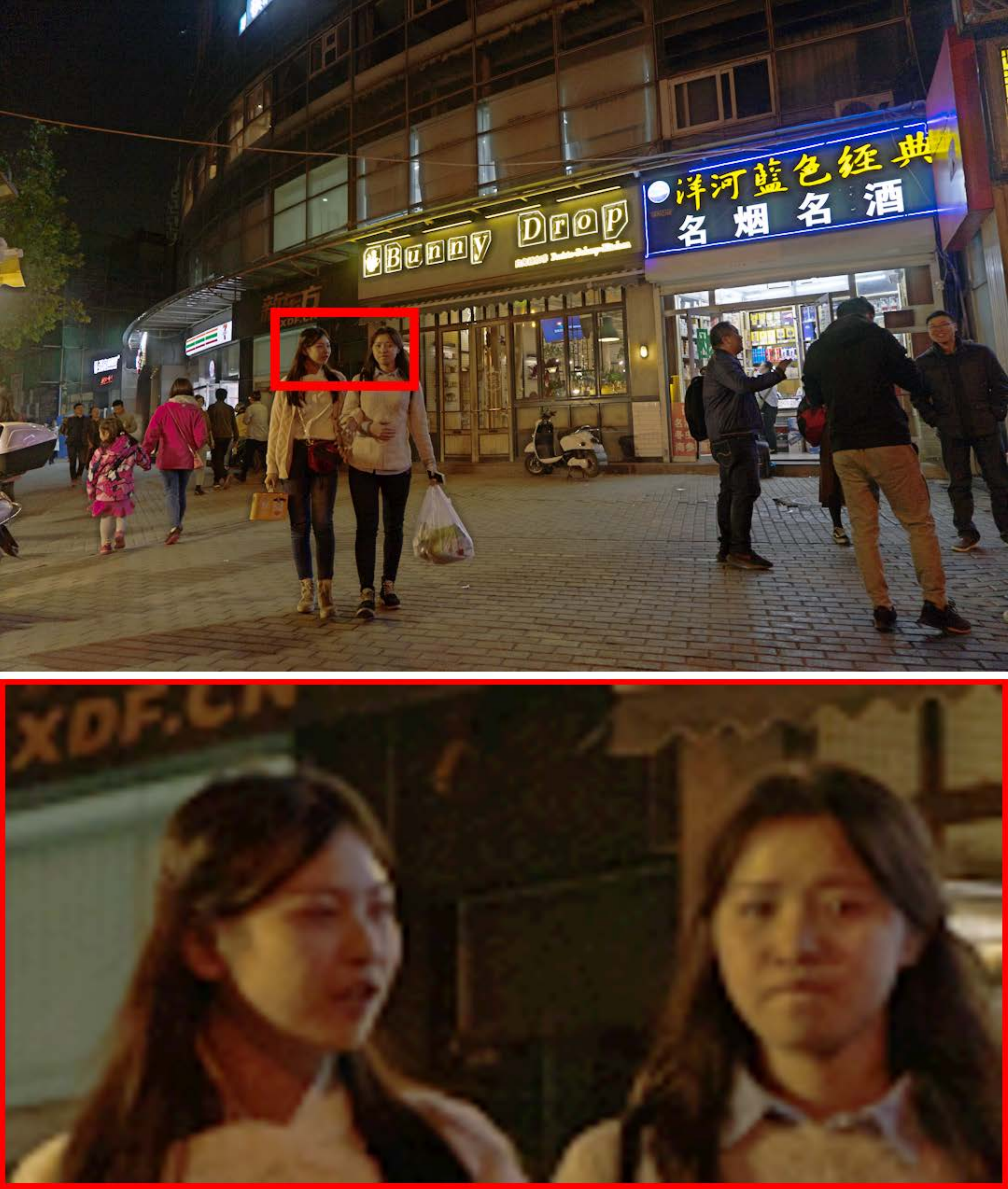}&
		\includegraphics[width=0.116\textwidth]{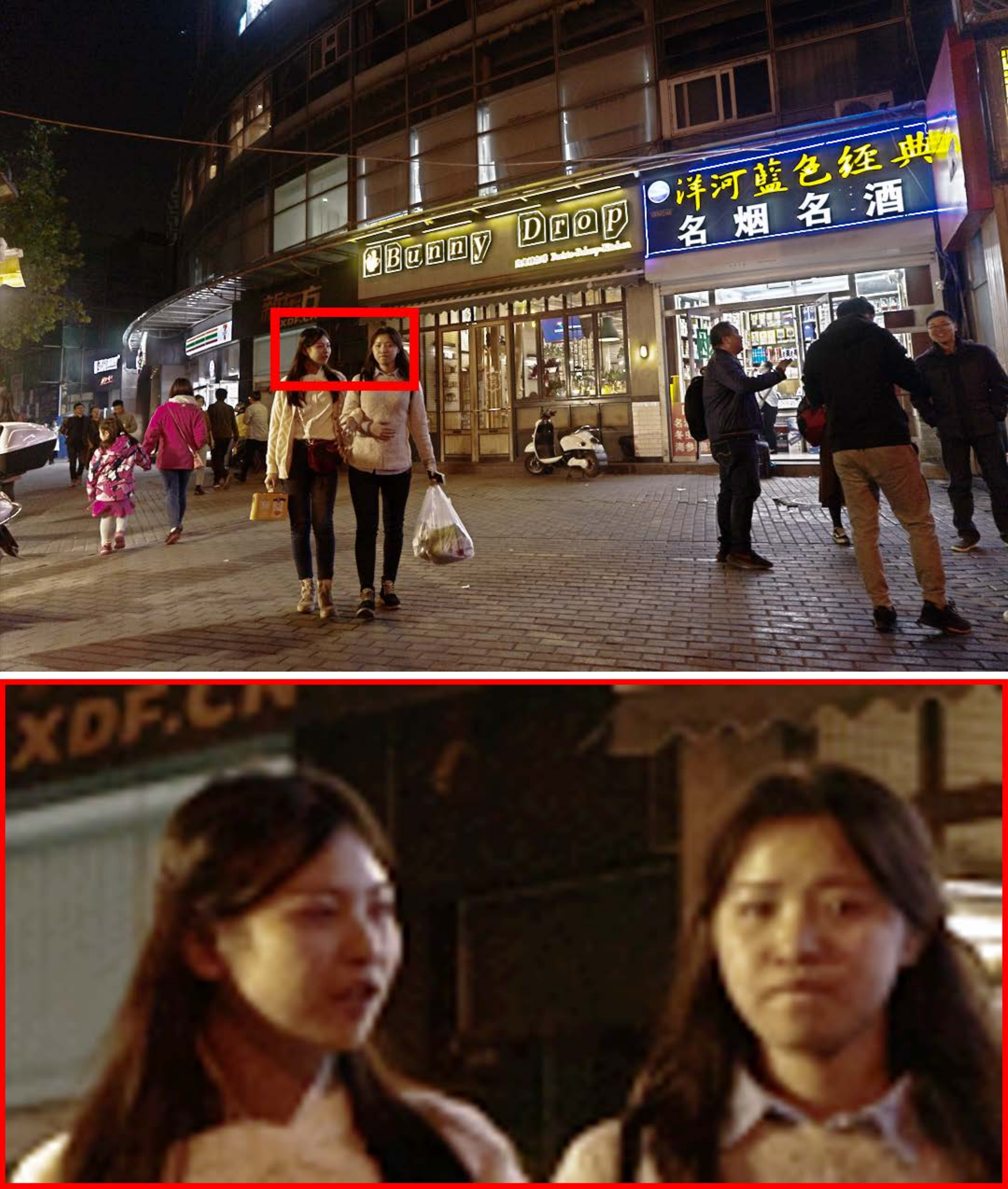}\\
		\includegraphics[width=0.116\textwidth]{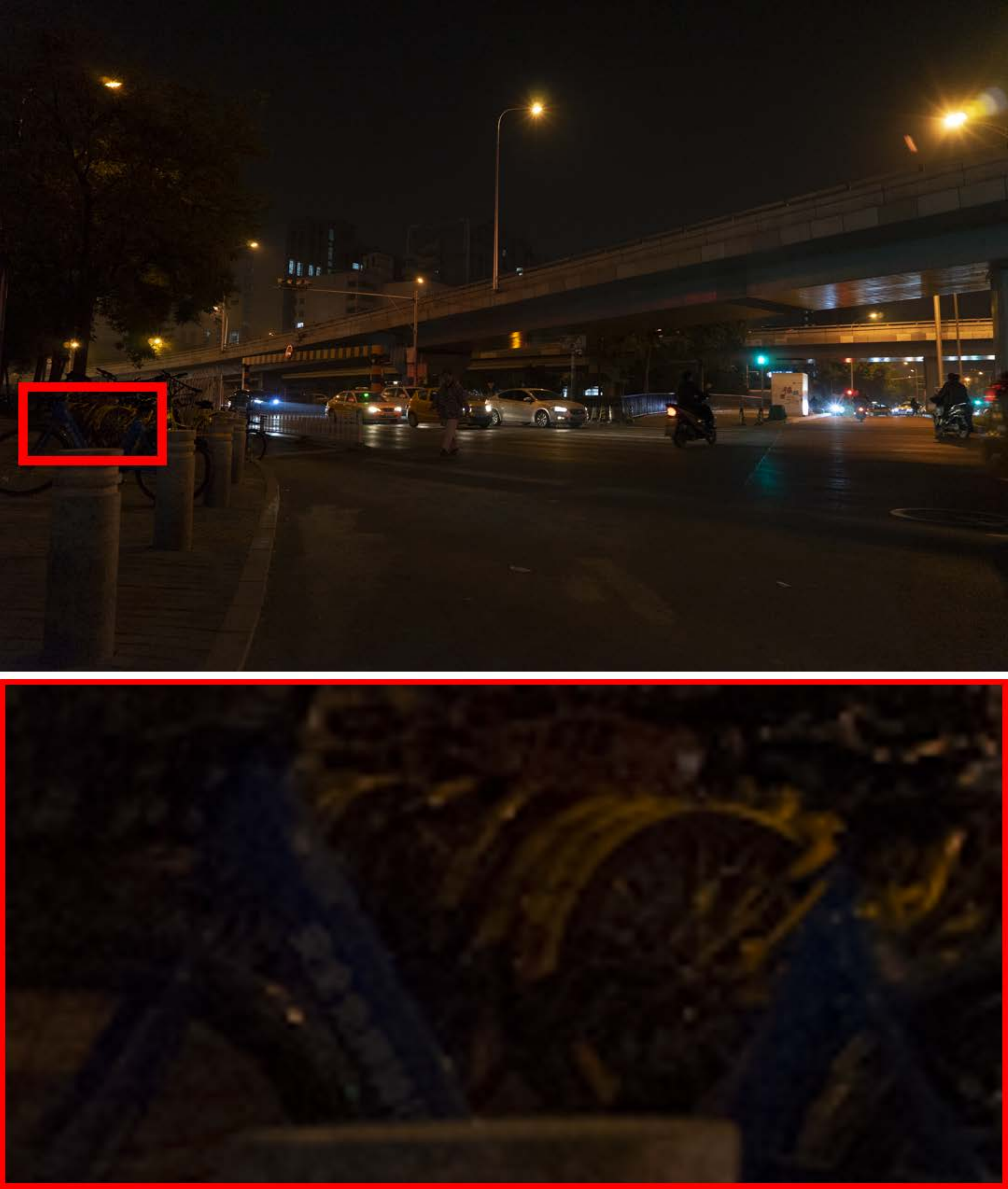}&
		\includegraphics[width=0.116\textwidth]{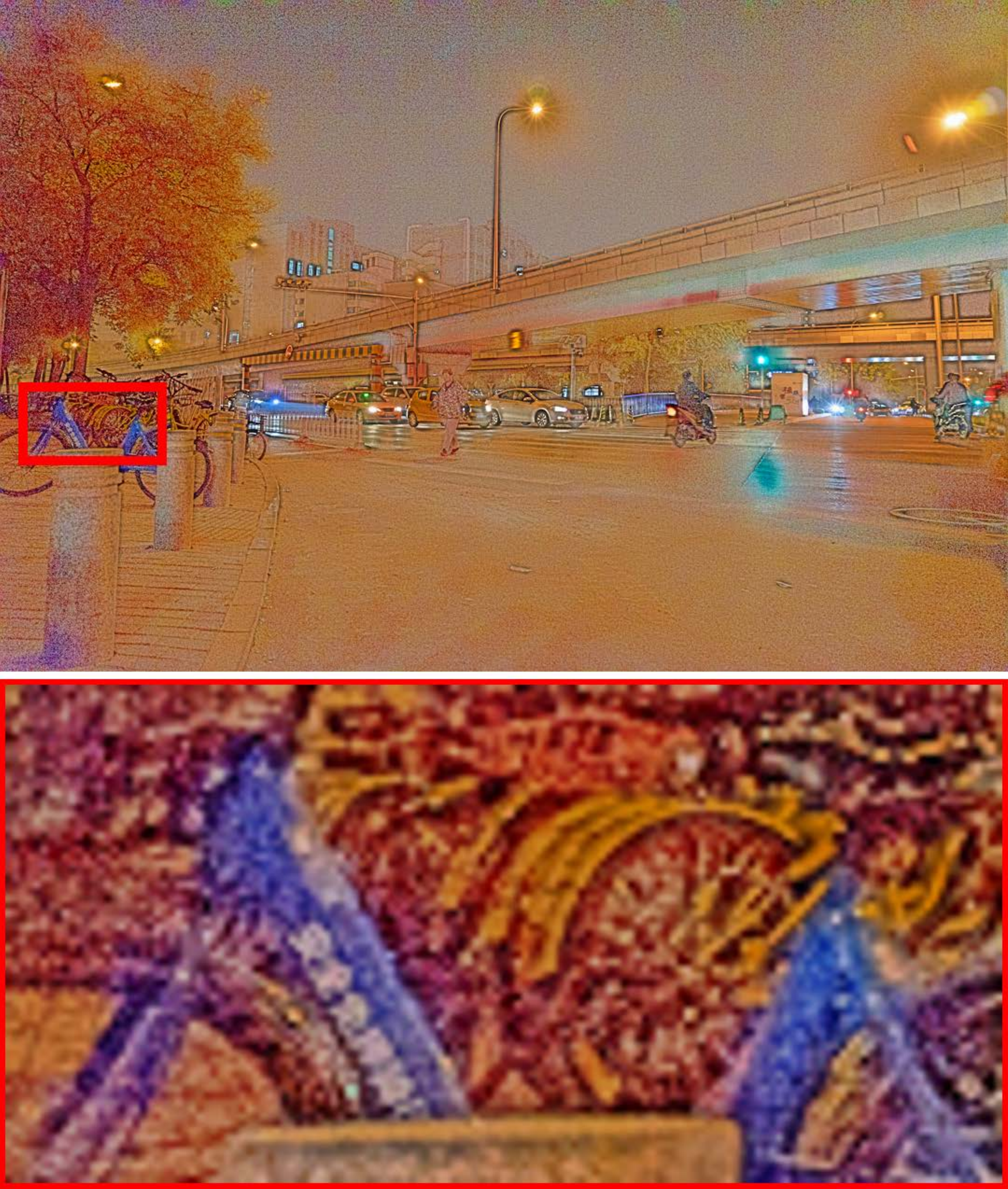}&
		\includegraphics[width=0.116\textwidth]{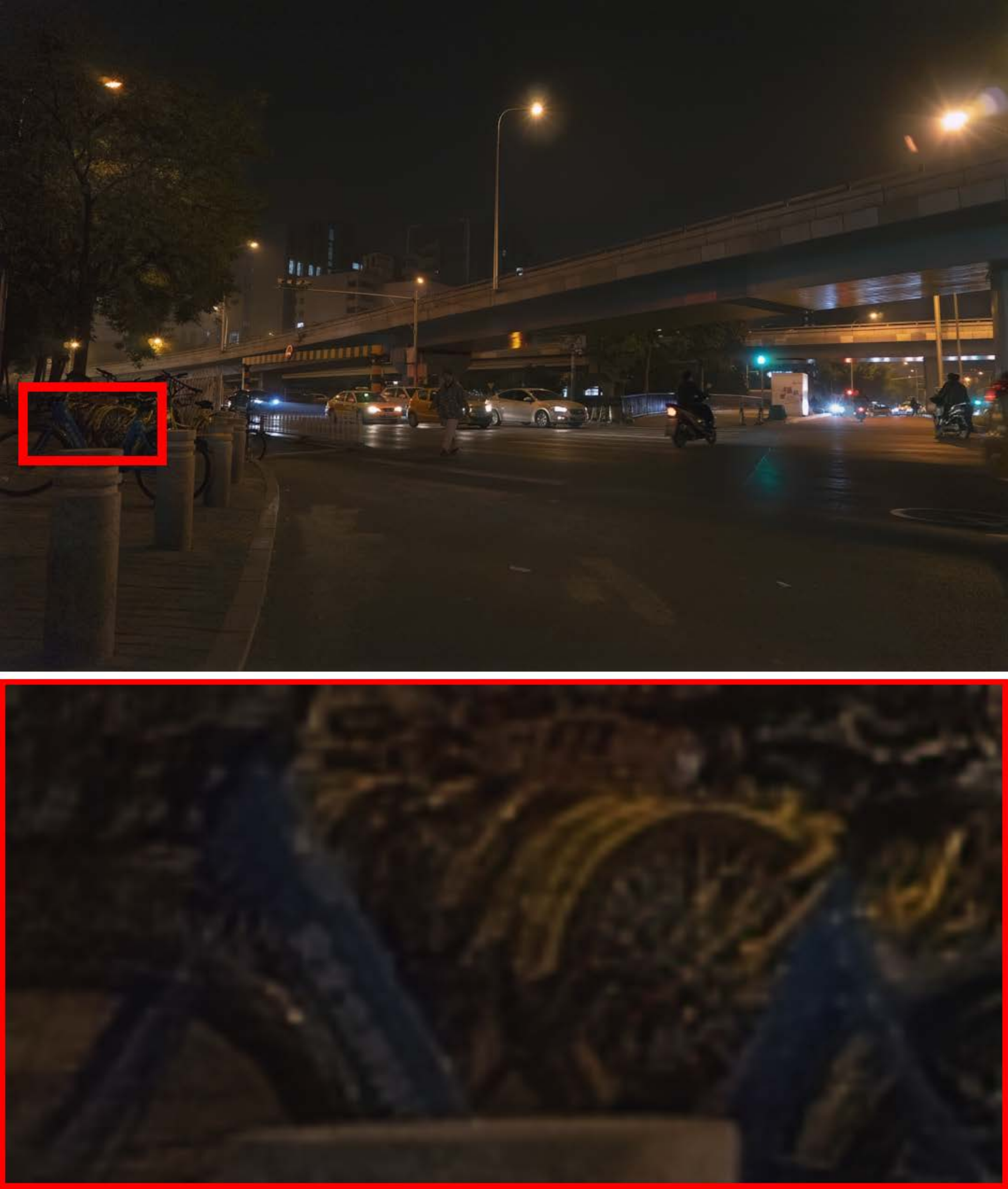}&
		\includegraphics[width=0.116\textwidth]{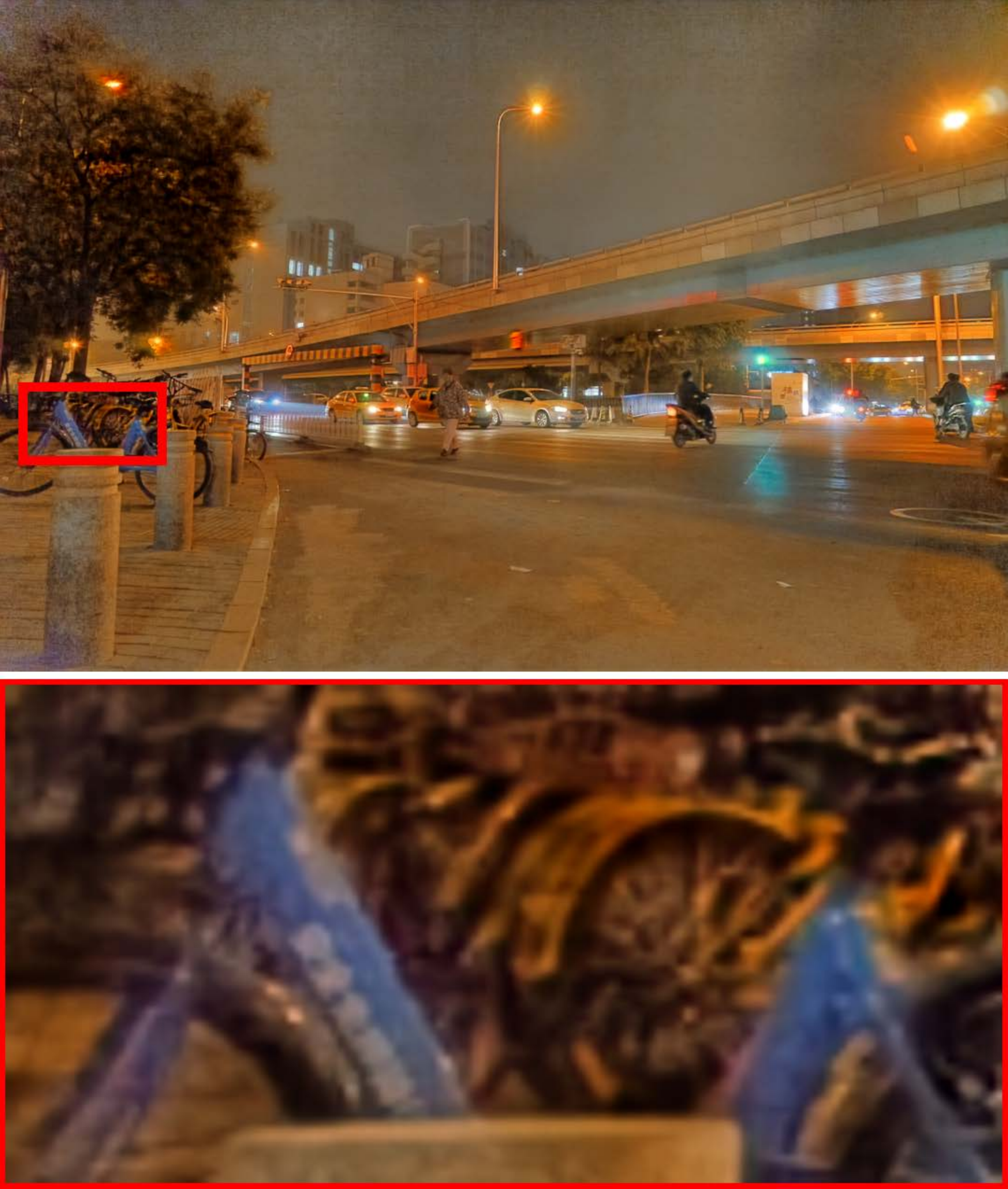}&
		\includegraphics[width=0.116\textwidth]{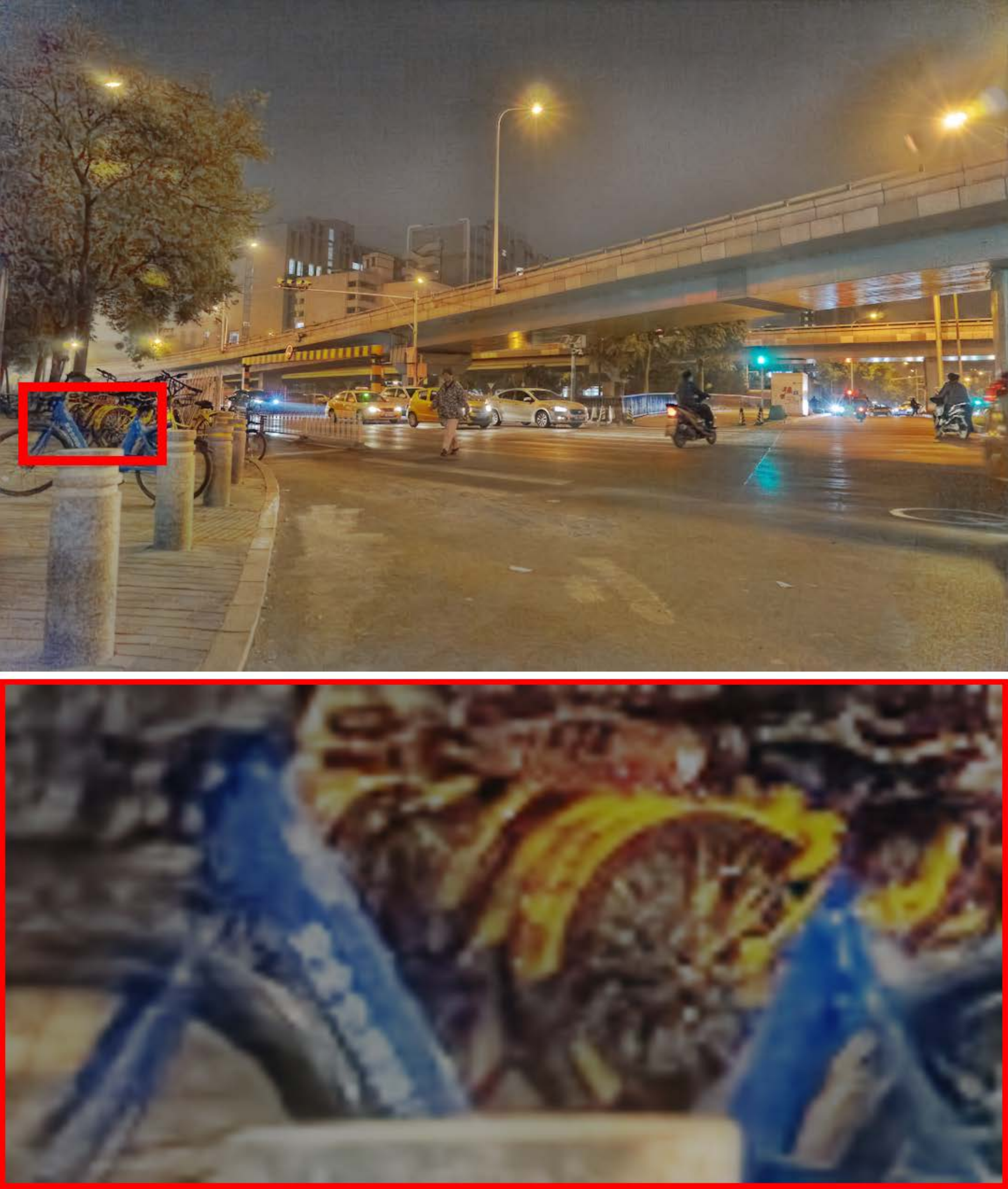}&
		\includegraphics[width=0.116\textwidth]{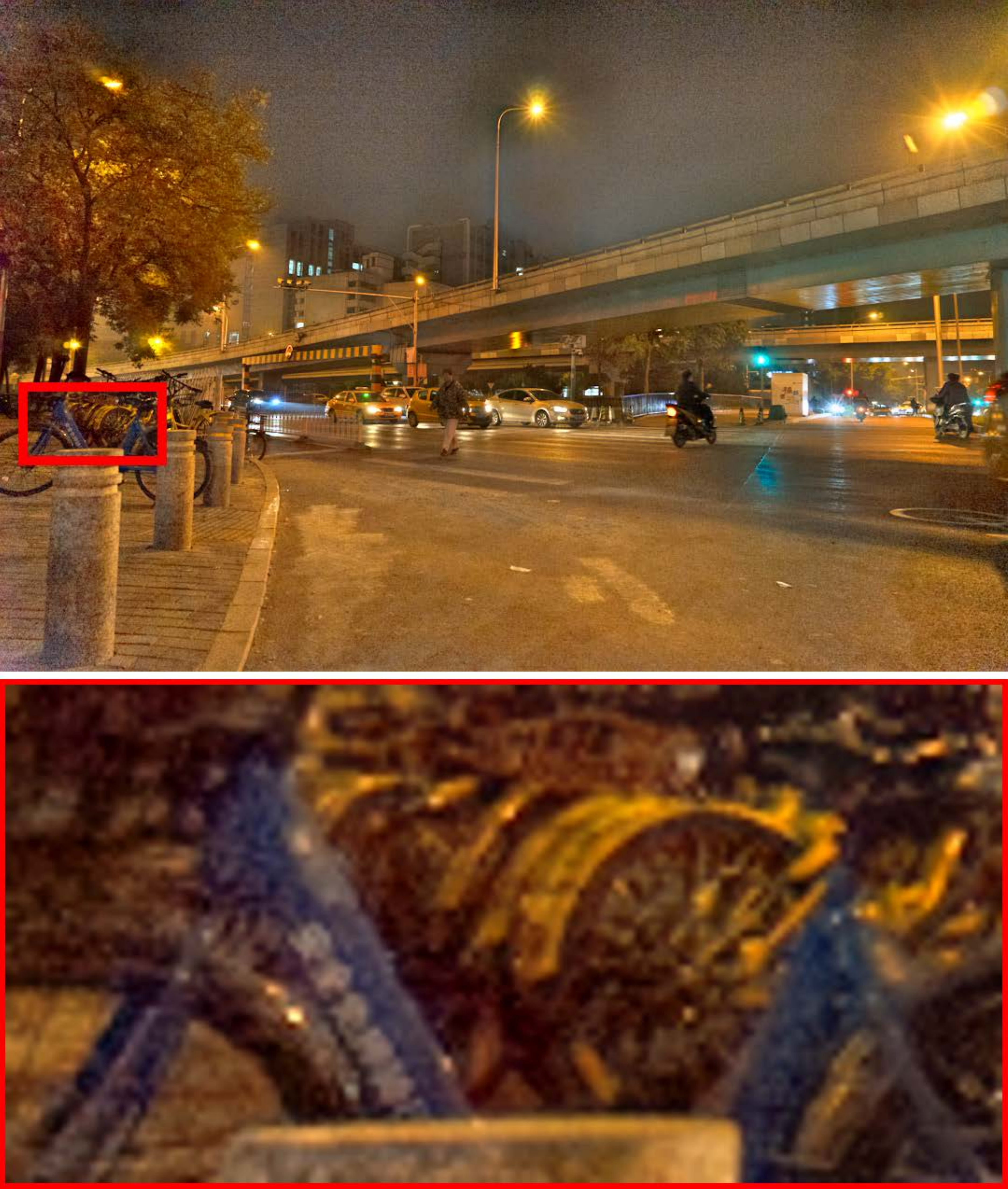}&
		\includegraphics[width=0.116\textwidth]{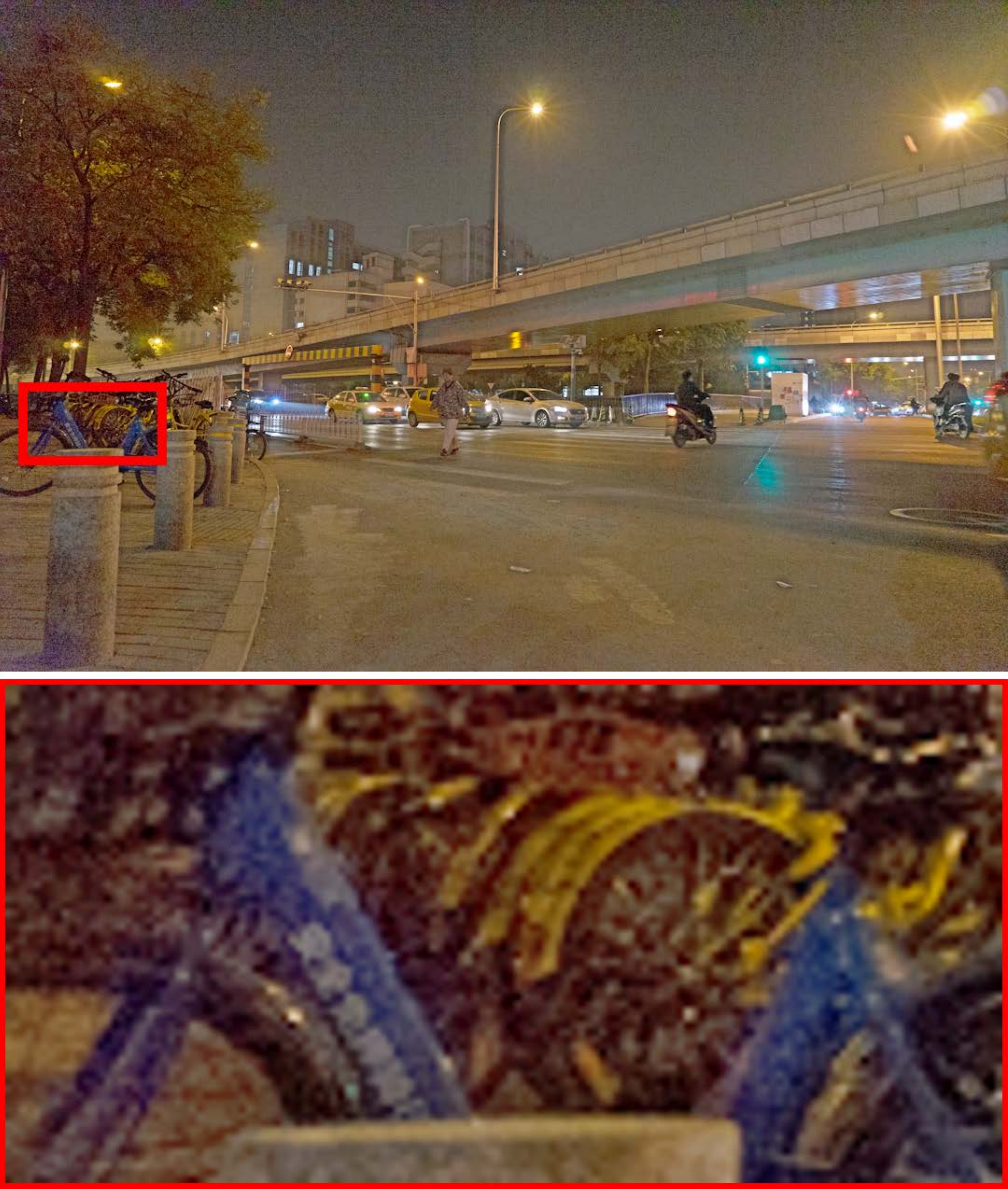}&
		\includegraphics[width=0.116\textwidth]{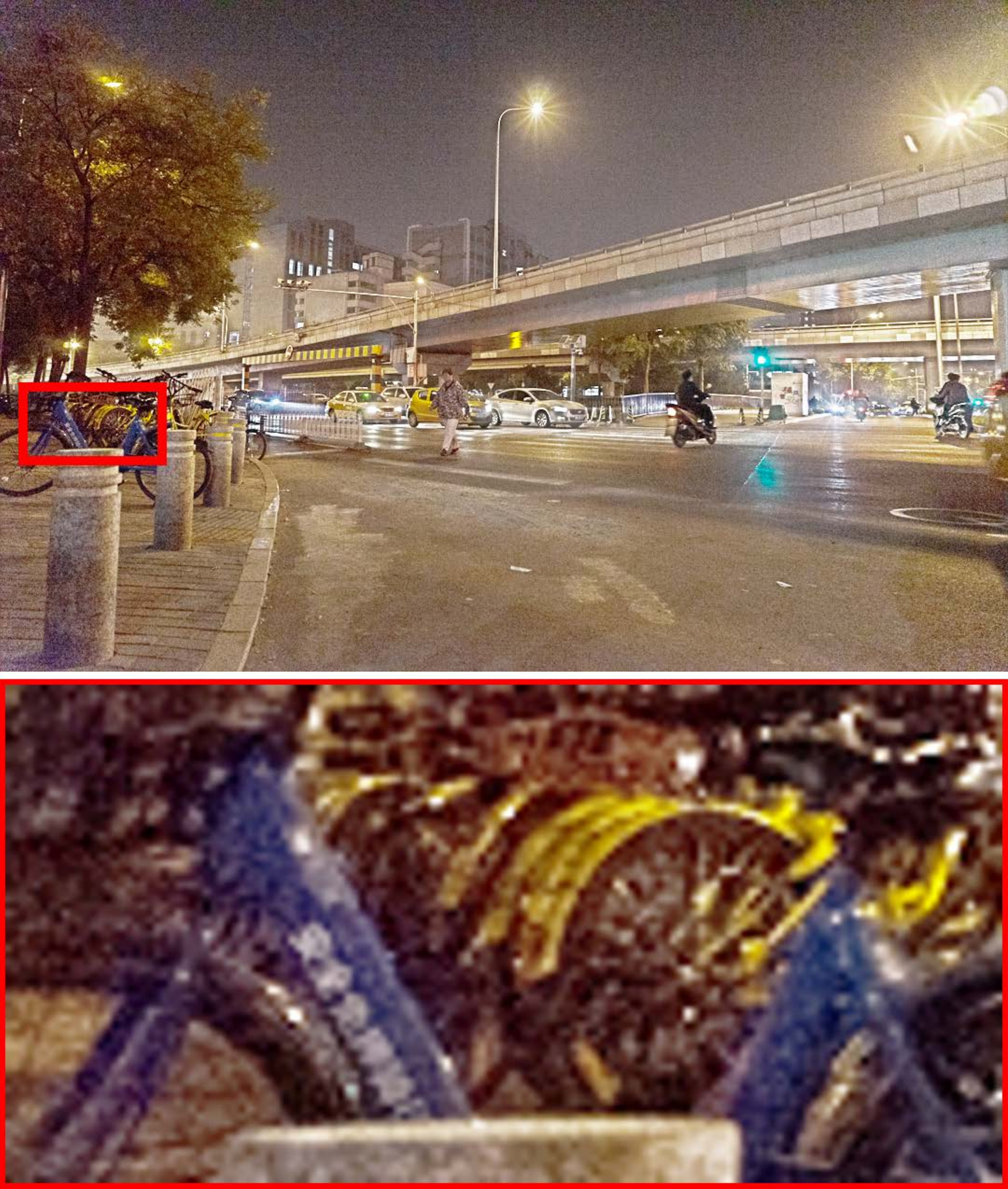}\\
		\footnotesize Input &\footnotesize RetinexNet&\footnotesize FIDE&\footnotesize DRBN&\footnotesize KinD&\footnotesize EnGAN&\footnotesize ZeroDCE&\footnotesize Ours\\
	\end{tabular}
	\vspace{-0.2cm}
	\caption{Visual comparison on some in-the-wild challenging examples. More results can be found in the Supplemental Materials. }
	\label{fig:MoreVC}
	\vspace{-0.2cm}
\end{figure*}

\subsection{Operation-Insensitive Adaptability}~\label{sec:flexibility}
In general, the operations used in network-based methods should be fixed and cannot be changed arbitrarily since these operations are acquired under the support of massive experiments. Fortunately, our proposed algorithm emerges the surprising adaptability on different exceedingly simple, even naive settings for $\mathcal{H}_{\bm{\theta}}$. As shown in Table~\ref{table:adaptability}, we can easily observe that our method acquired a stable performance among different settings (numbers of the block $3\times 3$ convolution+ReLU). Further, we provide the visual comparison in Fig.~\ref{fig:adaptability}, it can be easily observed that our SCIs with different settings all brighten the low-light observation, showing very similar enhanced results. 
Revisiting our designed framework, this property can be acquired lies in SCI not only converts the consensus for illumination (i.e., residual learning) but also integrates the physical principle (i.e., element-wise division operation). This experiment also verifies the effectiveness and correctness of our designed SCI.

	\begin{table}[t]
			\renewcommand\arraystretch{1.1}
		\centering
		\begin{tabular}{|c|c|ccc|}
			\hline
			\multicolumn{2}{|c|}{\footnotesize Method}&\footnotesize SIZE (M)&\footnotesize FLOPs (G)&\footnotesize TIME (S)\\
			\hline
			\multirow{4}{*}{\rotatebox{90}{\footnotesize Supervised$\;\;$} }&\footnotesize RetinexNet&\footnotesize 0.8383&\footnotesize 136.0151&\footnotesize 0.1192\\
			&\footnotesize FIDE&\footnotesize 8.6213&\footnotesize 57.2401&\footnotesize 0.5936\\
			&\footnotesize DRBN&\footnotesize 0.5770&\footnotesize 37.7902&\footnotesize 0.0533\\
			&\footnotesize KinD&\footnotesize 8.5402&\footnotesize 29.1303&\footnotesize 0.1814\\
			\hline
			\multirow{5}{*}{\rotatebox{90}{\footnotesize Unsupervised$\;$} }&\footnotesize EnGAN&\footnotesize 8.6360&\footnotesize 61.0102&\footnotesize 0.0097\\
			&\footnotesize SSIENet&\footnotesize 0.6824&\footnotesize 34.6070&\footnotesize 0.0272\\
			&\footnotesize ZeroDCE&\footnotesize {{0.0789}}&\footnotesize {{5.2112}}&\footnotesize \color{blue}{\textbf{0.0042}}\\
			&\footnotesize RUAS &\footnotesize \color{blue}{\textbf{0.0014}}&\footnotesize \color{blue}{\textbf{0.2813}}&\footnotesize 0.0063\\
			&\footnotesize Ours &\footnotesize \color{red}{\textbf{0.0003}}&\footnotesize \color{red}{\textbf{0.0619}}&\footnotesize \color{red}{\textbf{0.0017}}\\
			\hline
		\end{tabular}
		\vspace{-0.2cm}
		\caption{The model size, FLOPs and running time (GPU-seconds for inference) of CNN-based methods and our SCI. }
		\label{table:parameters}
		\vspace{-0.2cm}
	\end{table}

	\begin{figure}[t]
		\centering
		\begin{tabular}{c} 
			\includegraphics[width=0.46\textwidth]{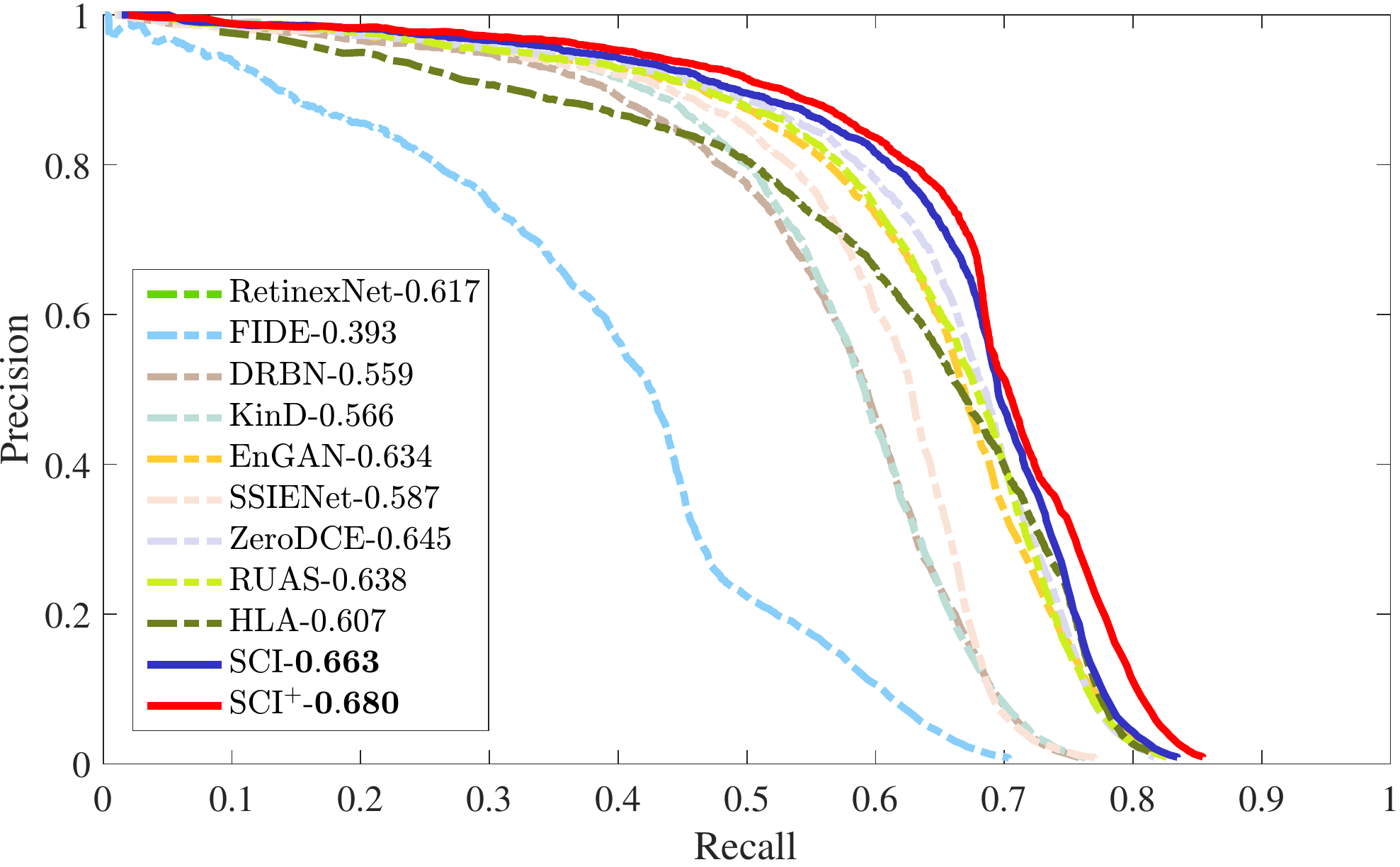}\\
		\end{tabular}
		\vspace{-0.2cm}
		\caption{Precision-Recall curve on the DARK FACE dataset. All compared methods and SCI fine-tune the detector on the enhanced results. SCI$^+$ is to jointly train the detector and SCI over the combination of losses for detection and enhancement.}
		\label{fig:PRcurve}
		\vspace{-0.2cm}
	\end{figure}

\subsection{Model-Irrelevant Generality}
Our SCI is actually a generalized learning paradigm if not limiting the task-related self-calibrated module, so ideally, it can be directly applied to existing works. Here, we take the recently-proposed representative work RUAS~\cite{liu2021retinex} as an example to make an exploration. Table~\ref{table:generality} and Fig.~\ref{fig:generality} demonstrate the quantitative and qualitative comparison before/after using our SCI to train RUAS. Obviously, although we just utilized a single block (i.e., RUAS (1)) used in the unrolling process of RUAS to evaluate our training process, the performance still attains significant improvement. More importantly, our method can remarkably suppress the overexposure that appeared in the original RUAS. This experiment reflects our learning framework is indeed flexible enough and has a strong model-irrelevant generality. Moreover, it indicates that perhaps our method can be transferred to arbitrary illumination-based low-light image enhancement works and we will try doing it in the future.

\section{Experimental Results}
In this section, we first provided all implementation details. Then we made experimental evaluations. Next, we applied enhancement methods to dark face detection and nighttime semantic segmentation. Finally, we conducted algorithmic analyses for SCI. All the experiments were performed on a PC with a single TITAN X GPU.

\begin{figure*}[t]
	\centering
	\begin{tabular}{c@{\extracolsep{0.2em}}c@{\extracolsep{0.2em}}c@{\extracolsep{0.2em}}c@{\extracolsep{0.2em}}c@{\extracolsep{0.2em}}c@{\extracolsep{0.2em}}c@{\extracolsep{0.2em}}c} 	
		\includegraphics[width=0.116\textwidth]{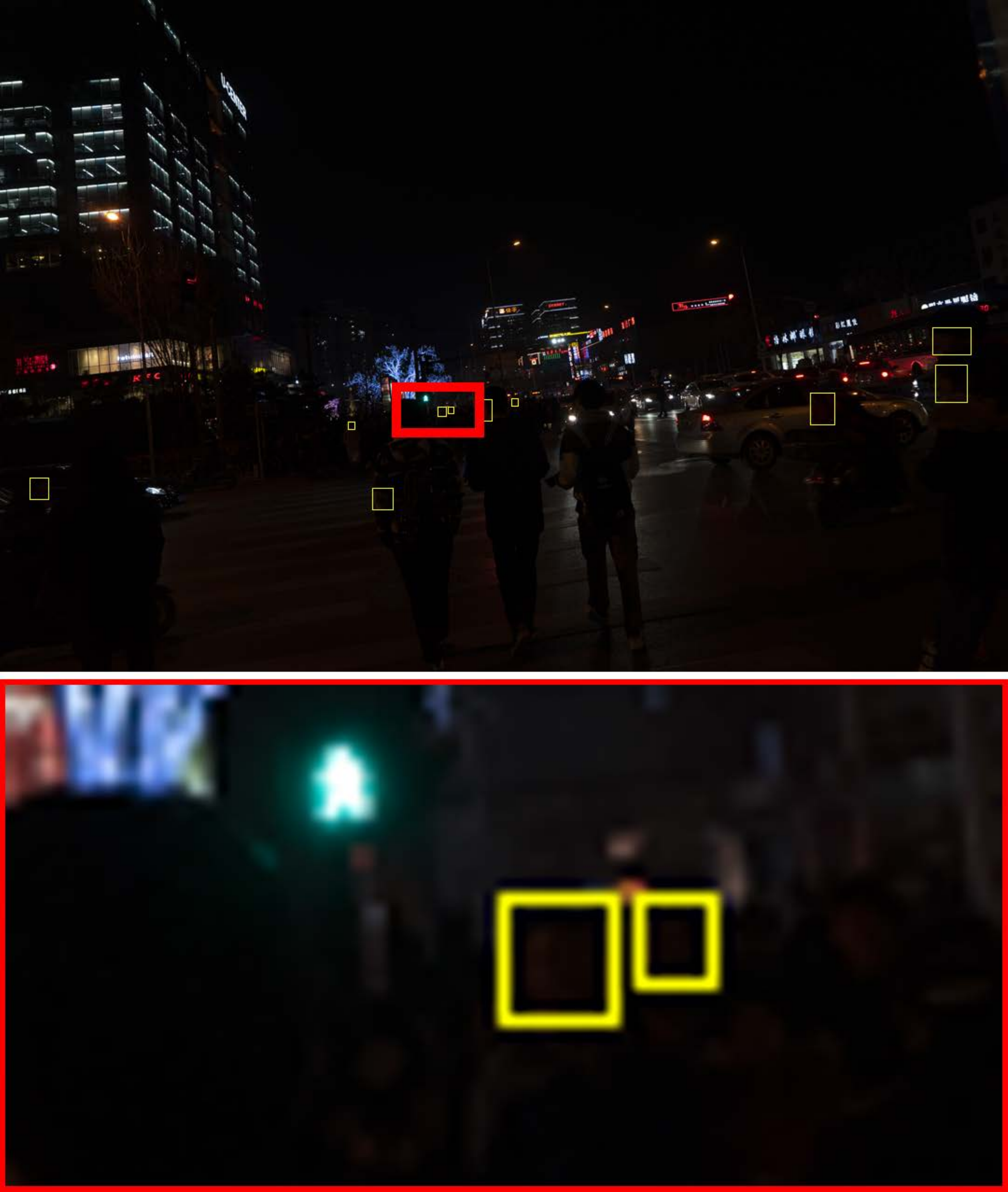}&	
		\includegraphics[width=0.116\textwidth]{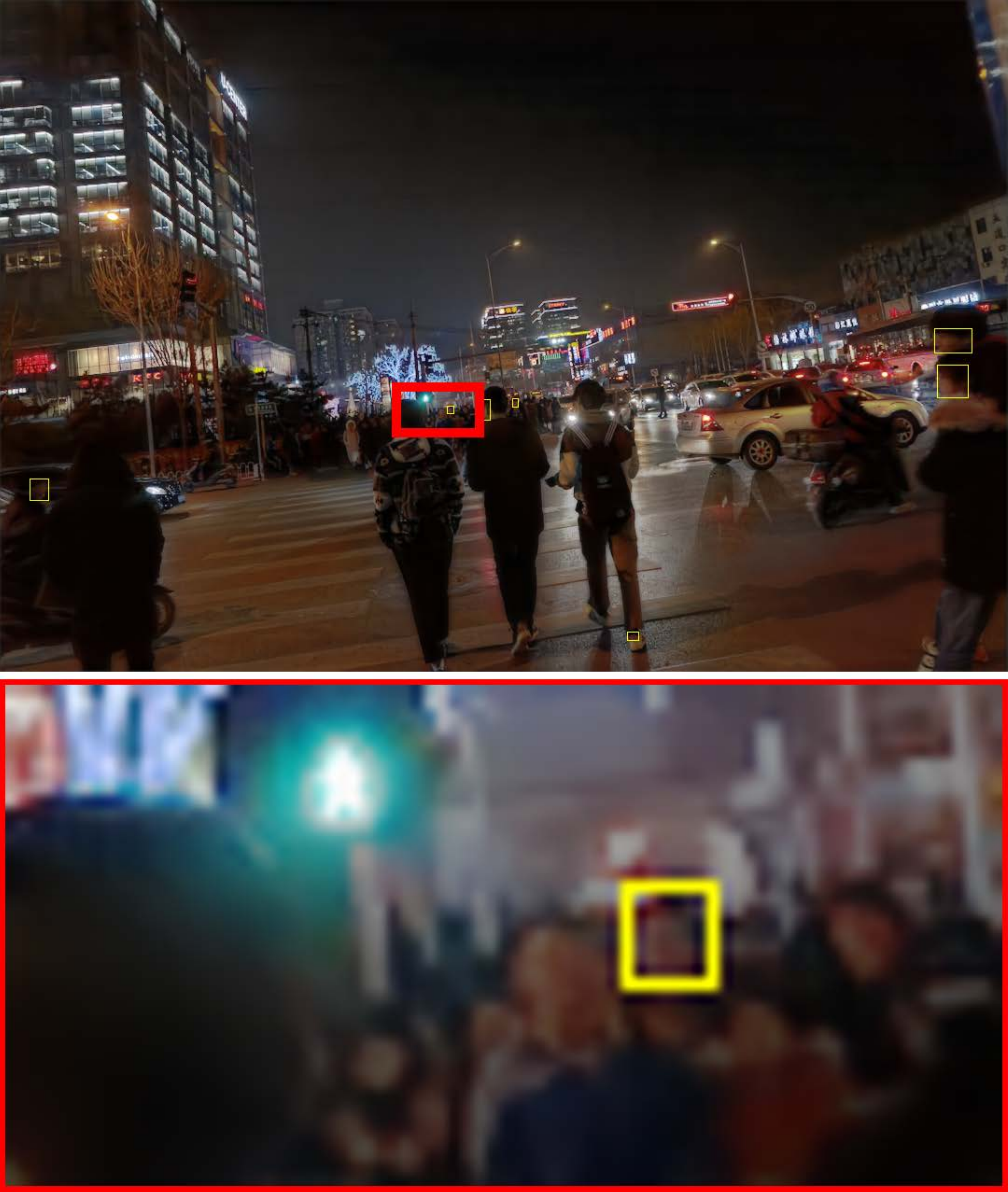}&	
		\includegraphics[width=0.116\textwidth]{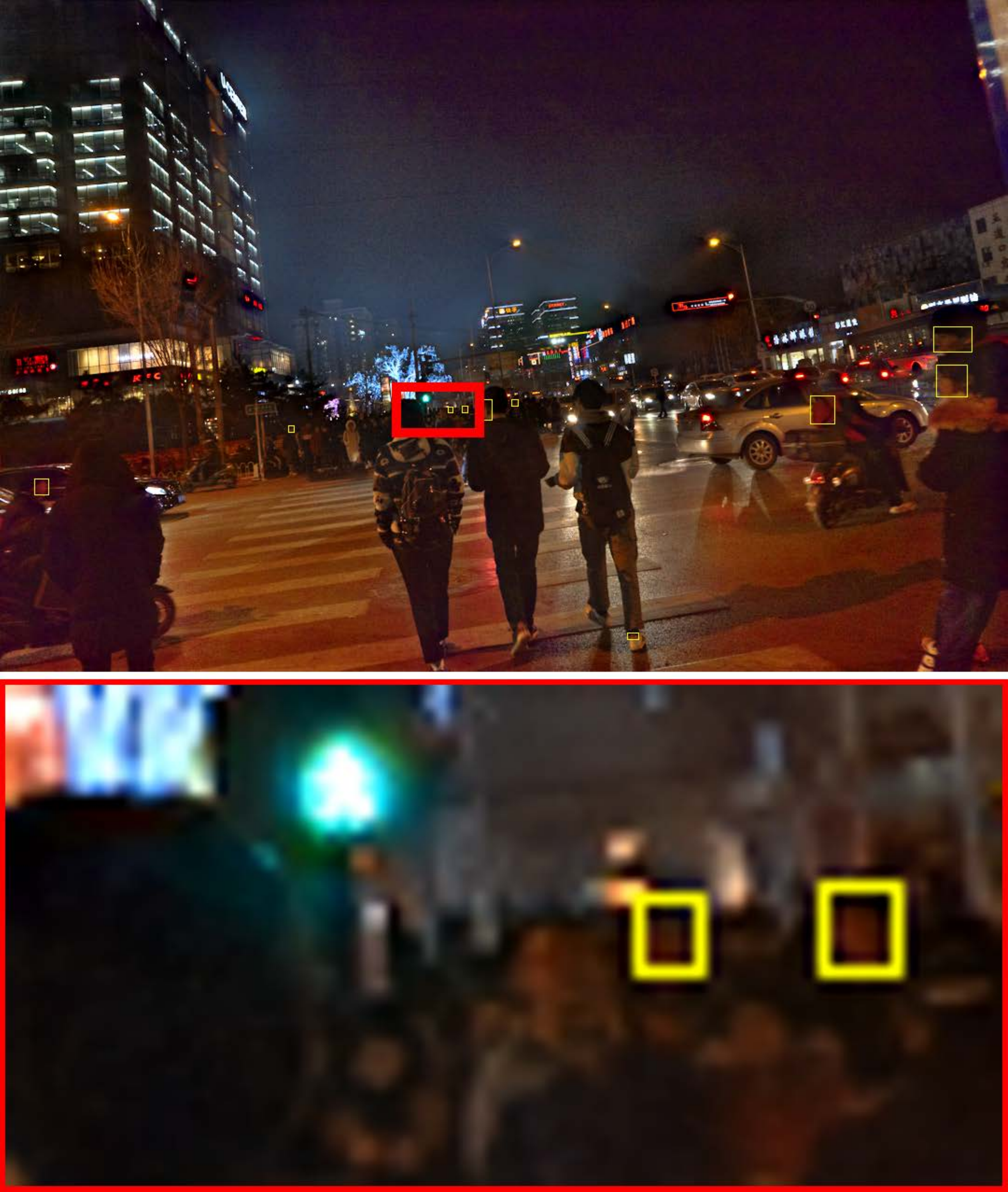}&	
		\includegraphics[width=0.116\textwidth]{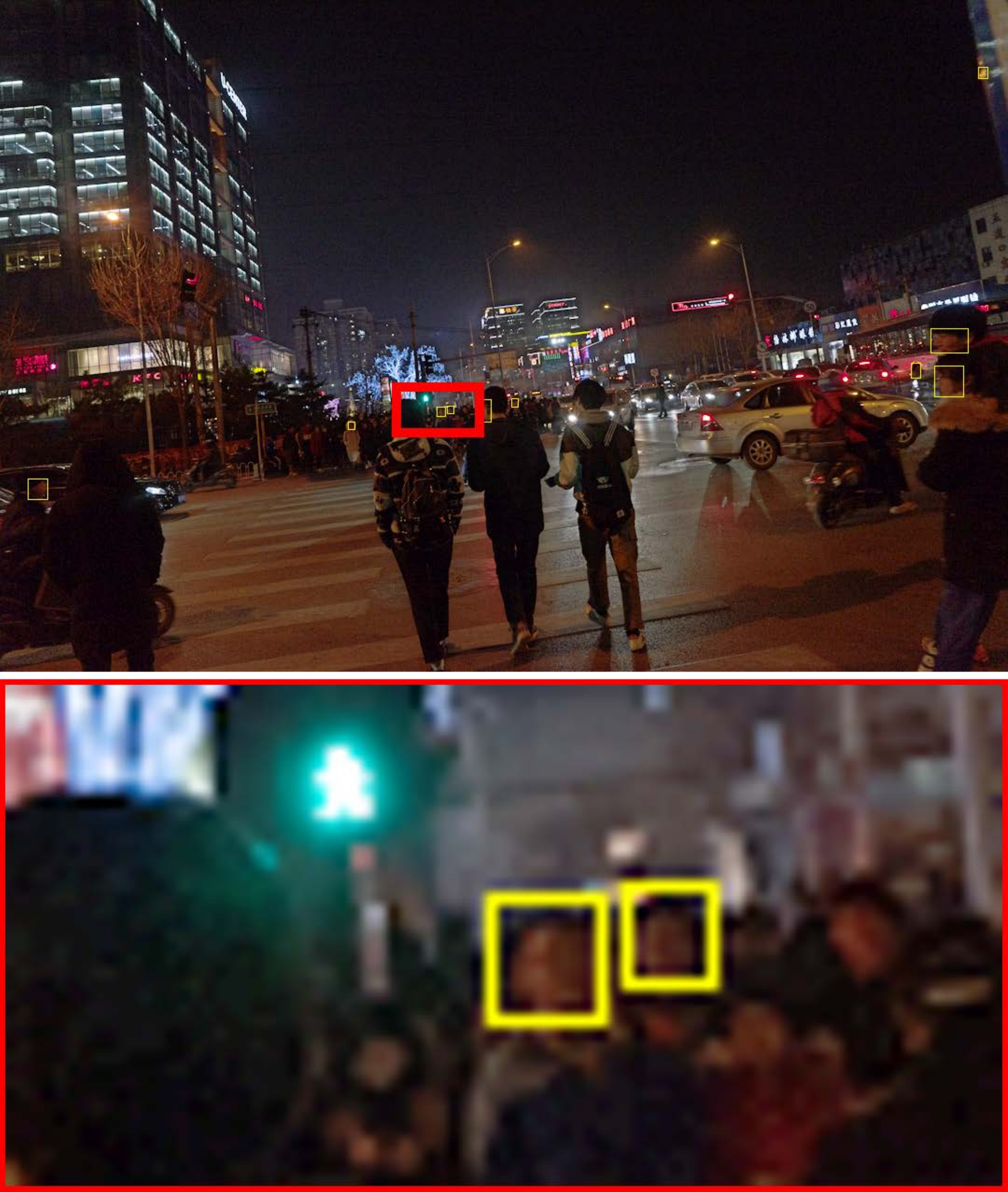}&	
		\includegraphics[width=0.116\textwidth]{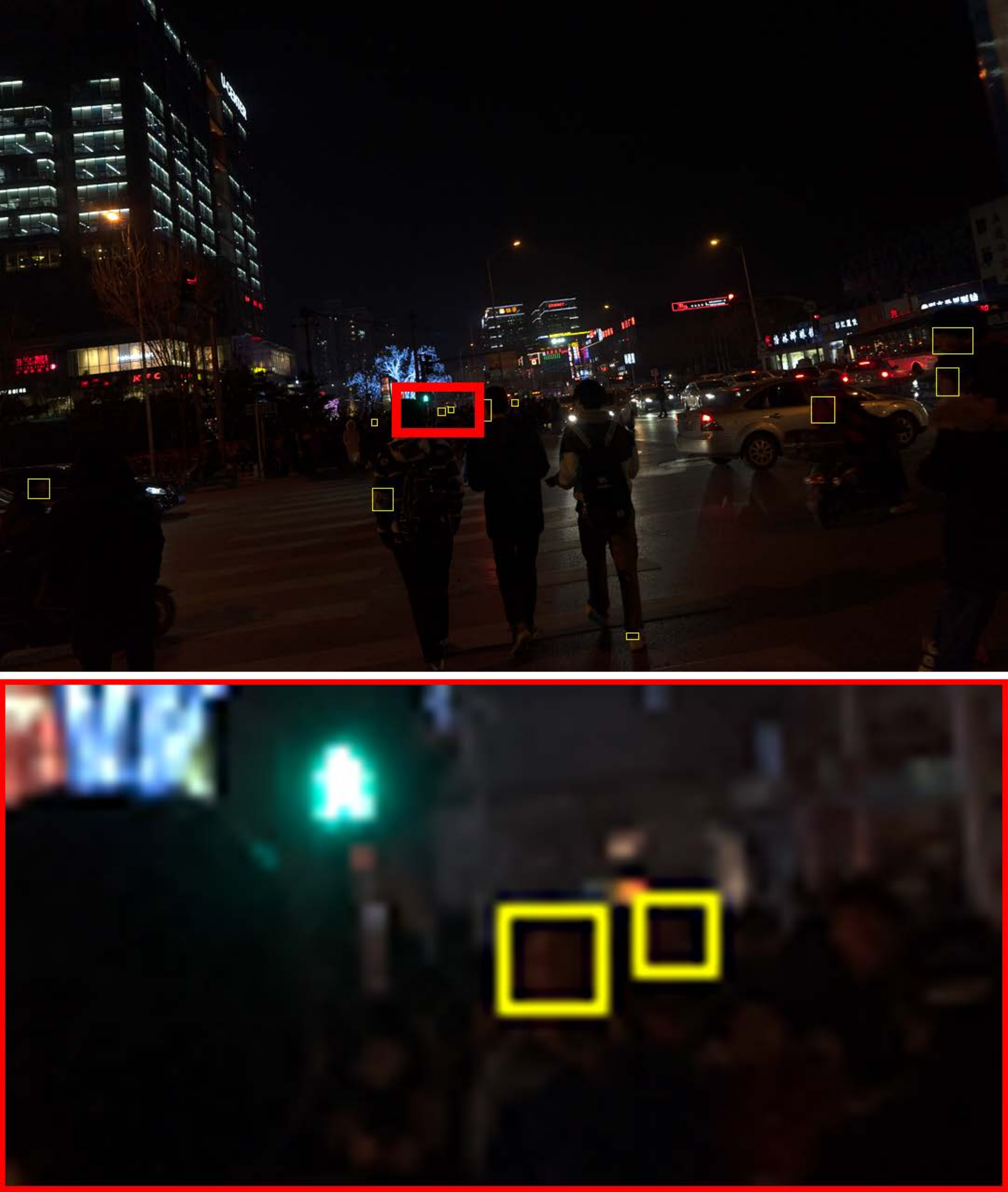}&	
		\includegraphics[width=0.116\textwidth]{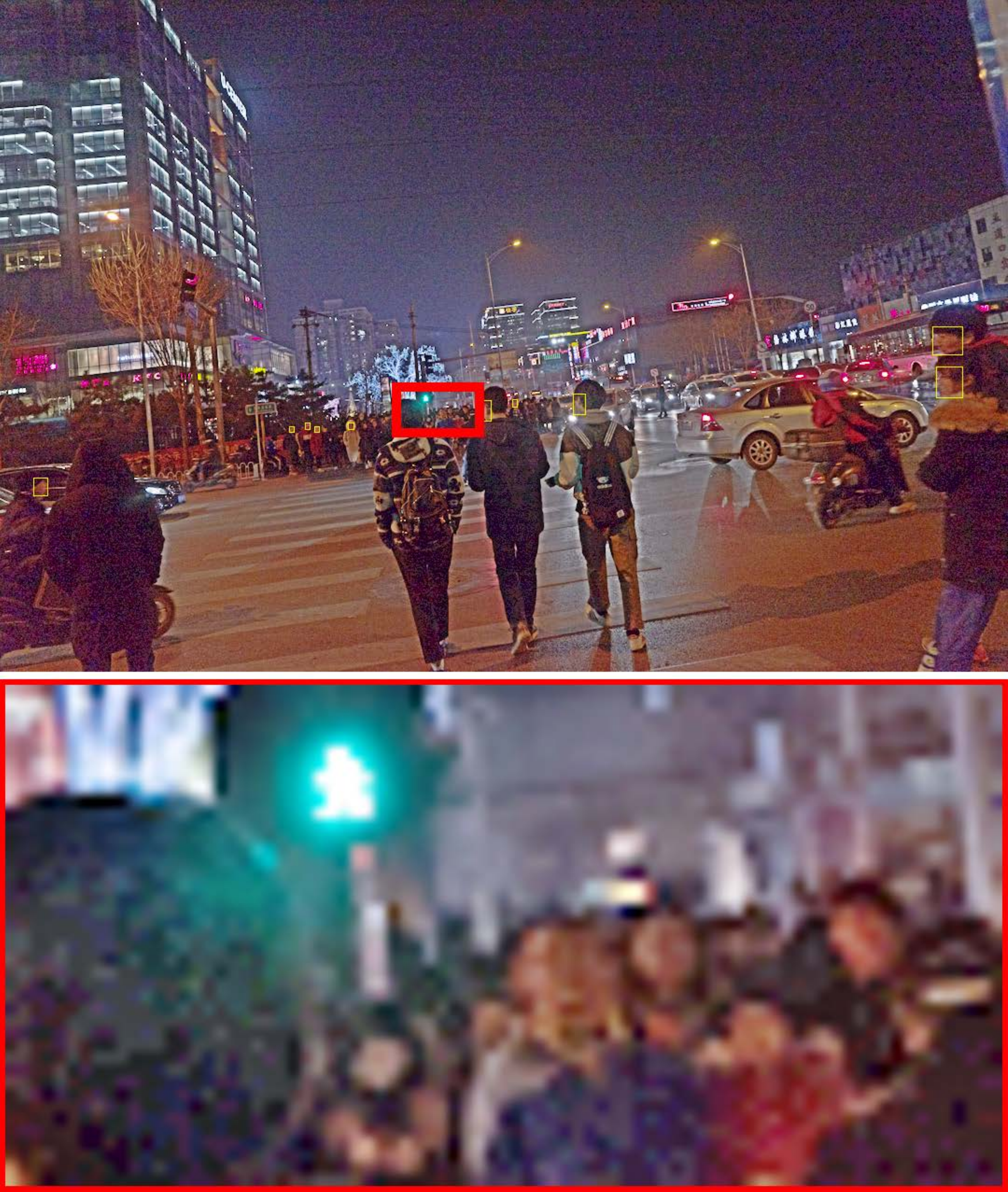}&			
		\includegraphics[width=0.116\textwidth]{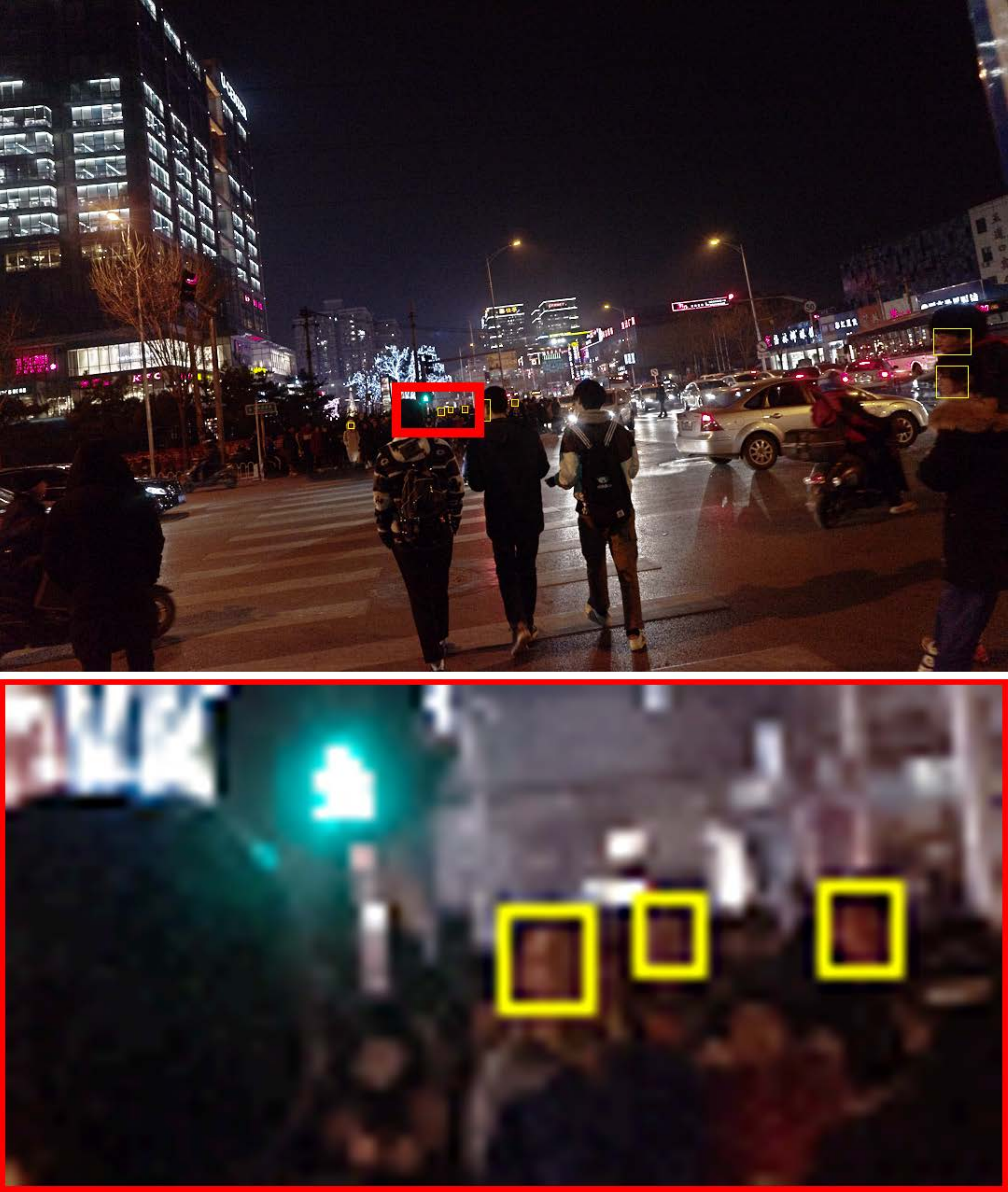}&	
		\includegraphics[width=0.116\textwidth]{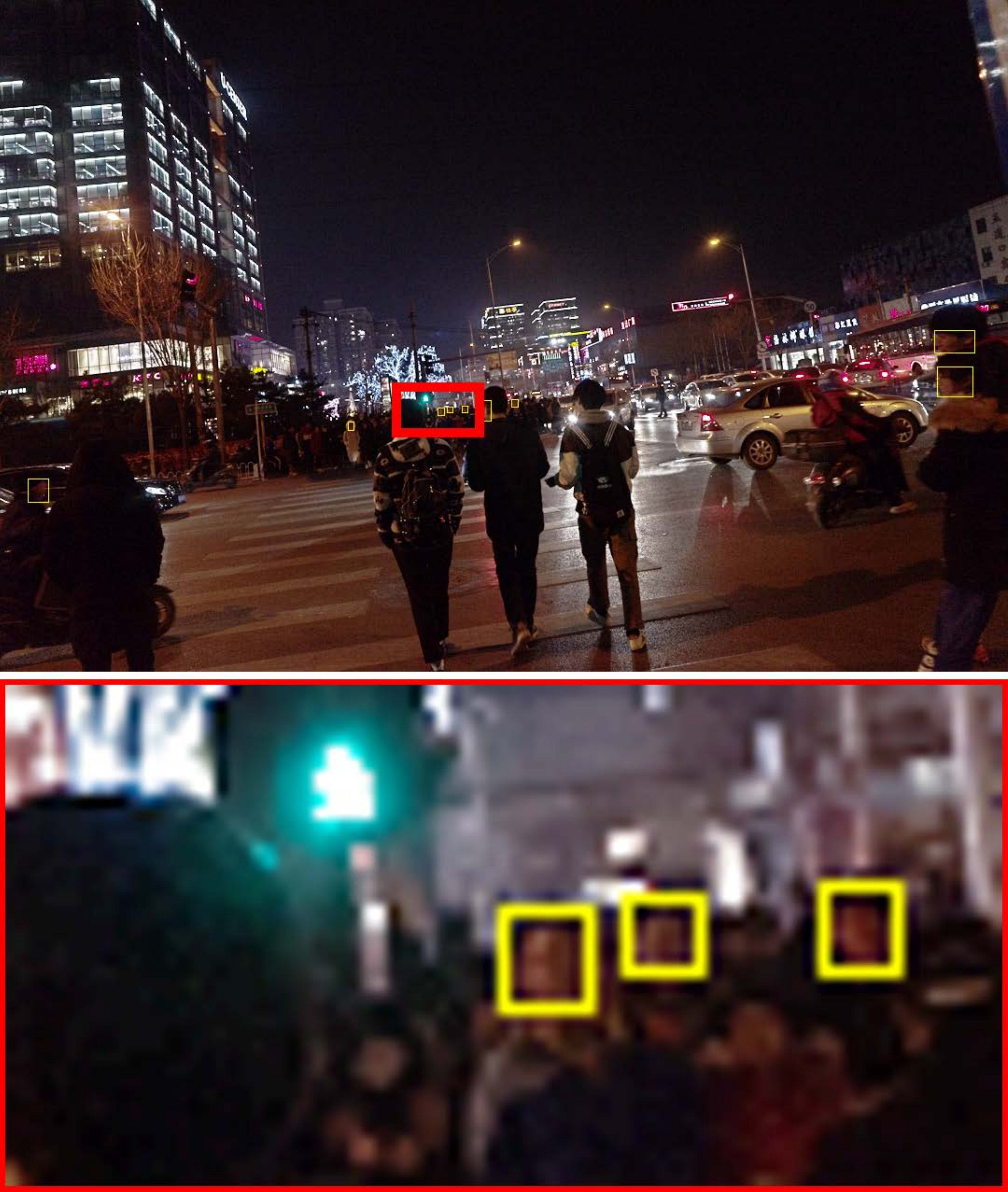}\\			
		\footnotesize Input&\footnotesize KinD&\footnotesize EnGAN &\footnotesize ZeroDCE &\footnotesize RUAS&\footnotesize HLA &\footnotesize SCI &\footnotesize SCI$^+$\\
	\end{tabular}
	\vspace{-0.2cm}
	\caption{Visual comparison of face detection on the DarkFace dataset. More results can be found in the Supplemental Materials.}
	\label{fig:DarkFace}
	\vspace{-0.2cm}
\end{figure*}

\begin{figure*}[t]
	\centering
	\begin{tabular}{c@{\extracolsep{0.25em}}c@{\extracolsep{0.25em}}c@{\extracolsep{0.25em}}c@{\extracolsep{0.25em}}c@{\extracolsep{0.25em}}c@{\extracolsep{0.25em}}c@{\extracolsep{0.25em}}c} 
		\includegraphics[width=0.116\textwidth]{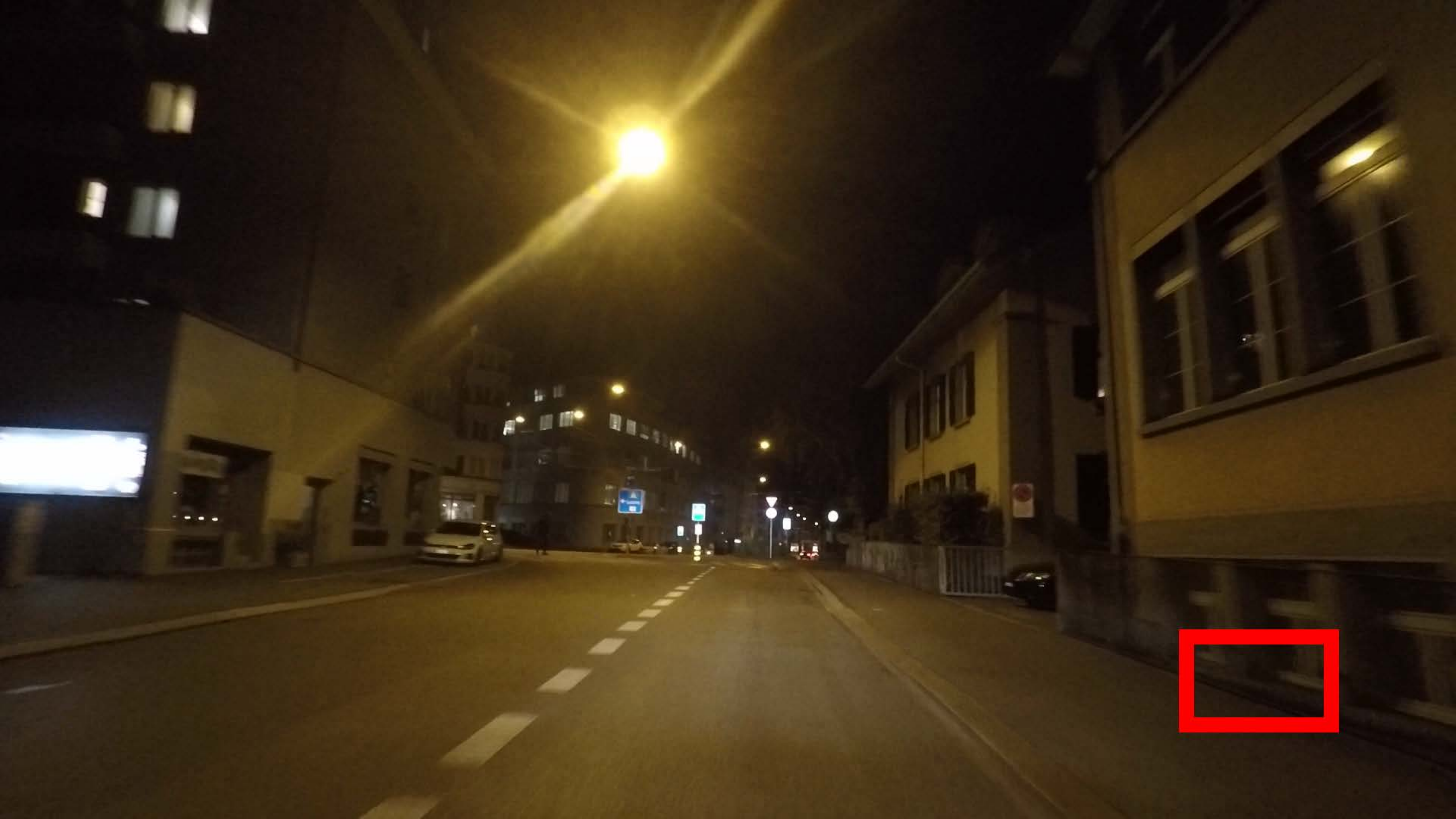}&
		\includegraphics[width=0.116\textwidth]{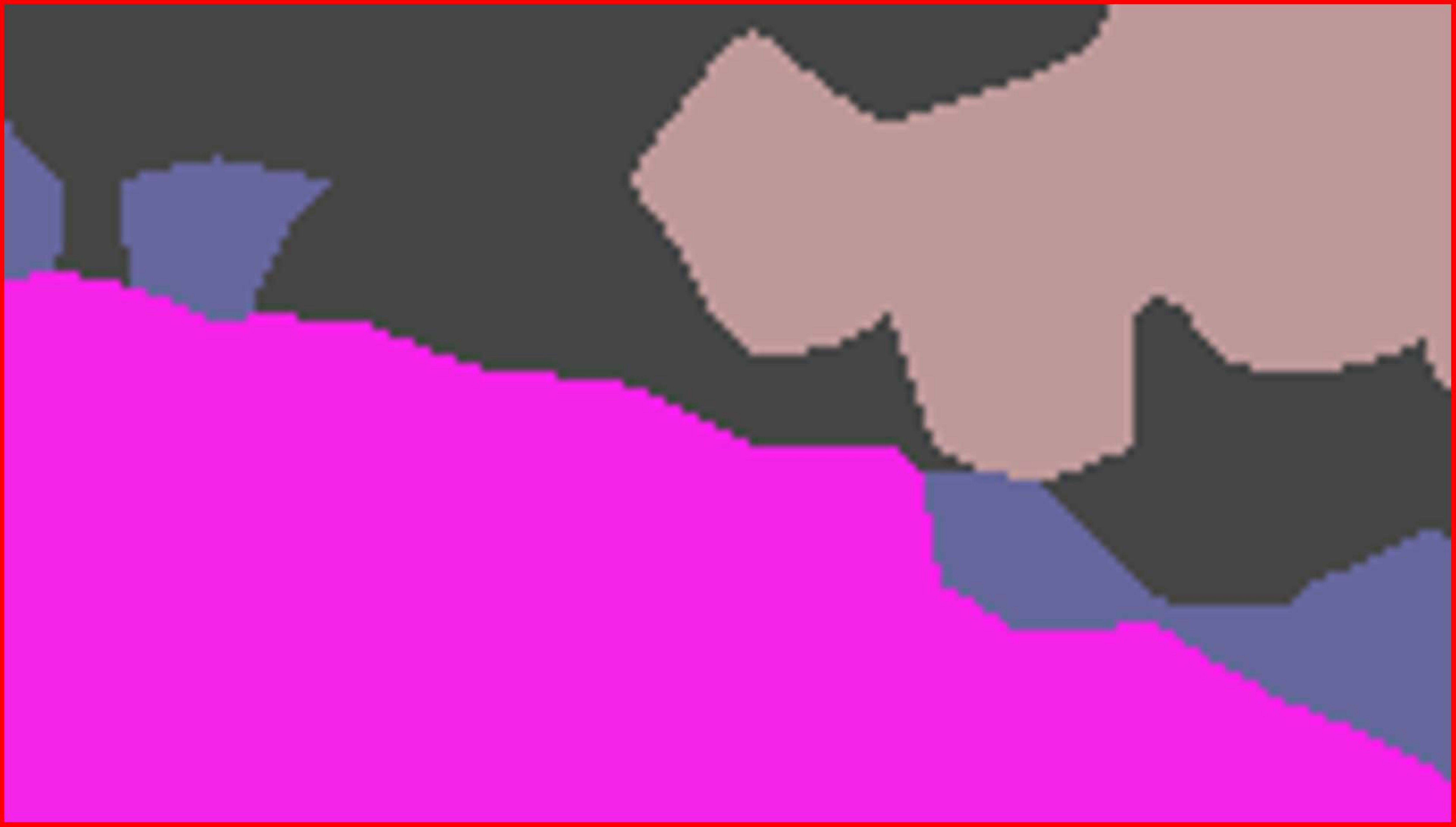}&
		\includegraphics[width=0.116\textwidth]{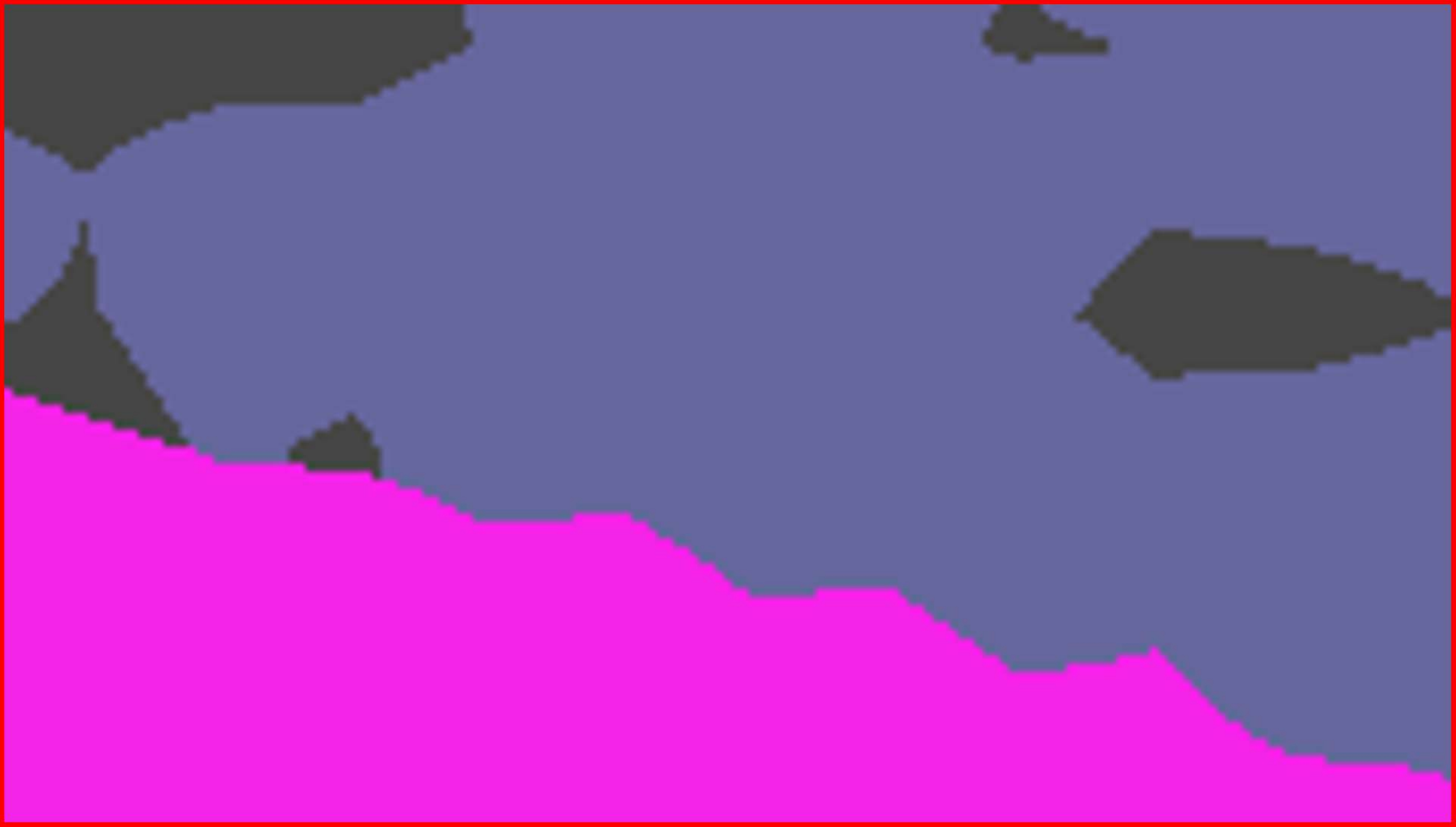}&
		\includegraphics[width=0.116\textwidth]{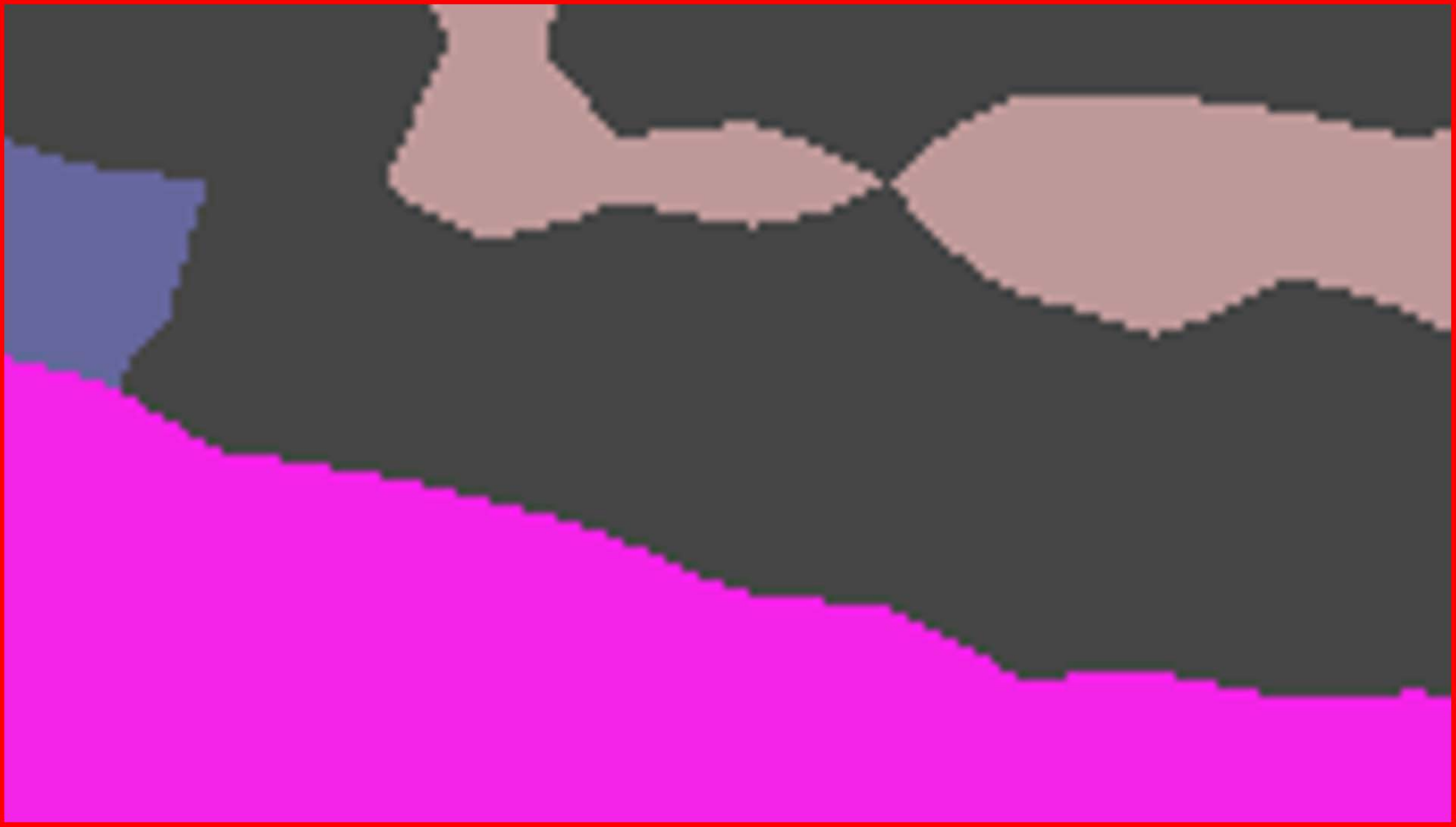}&
		\includegraphics[width=0.116\textwidth]{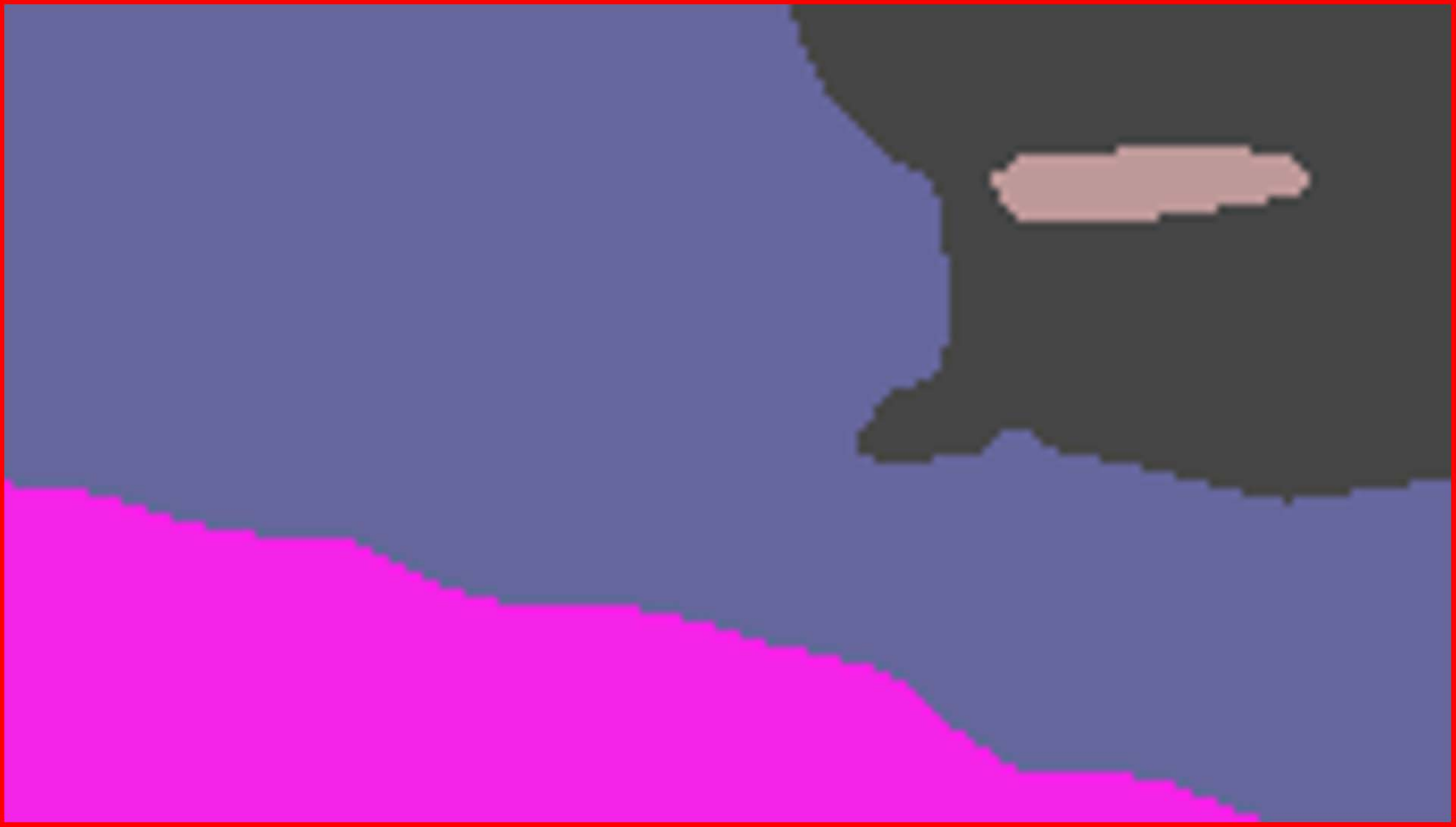}&
		\includegraphics[width=0.116\textwidth]{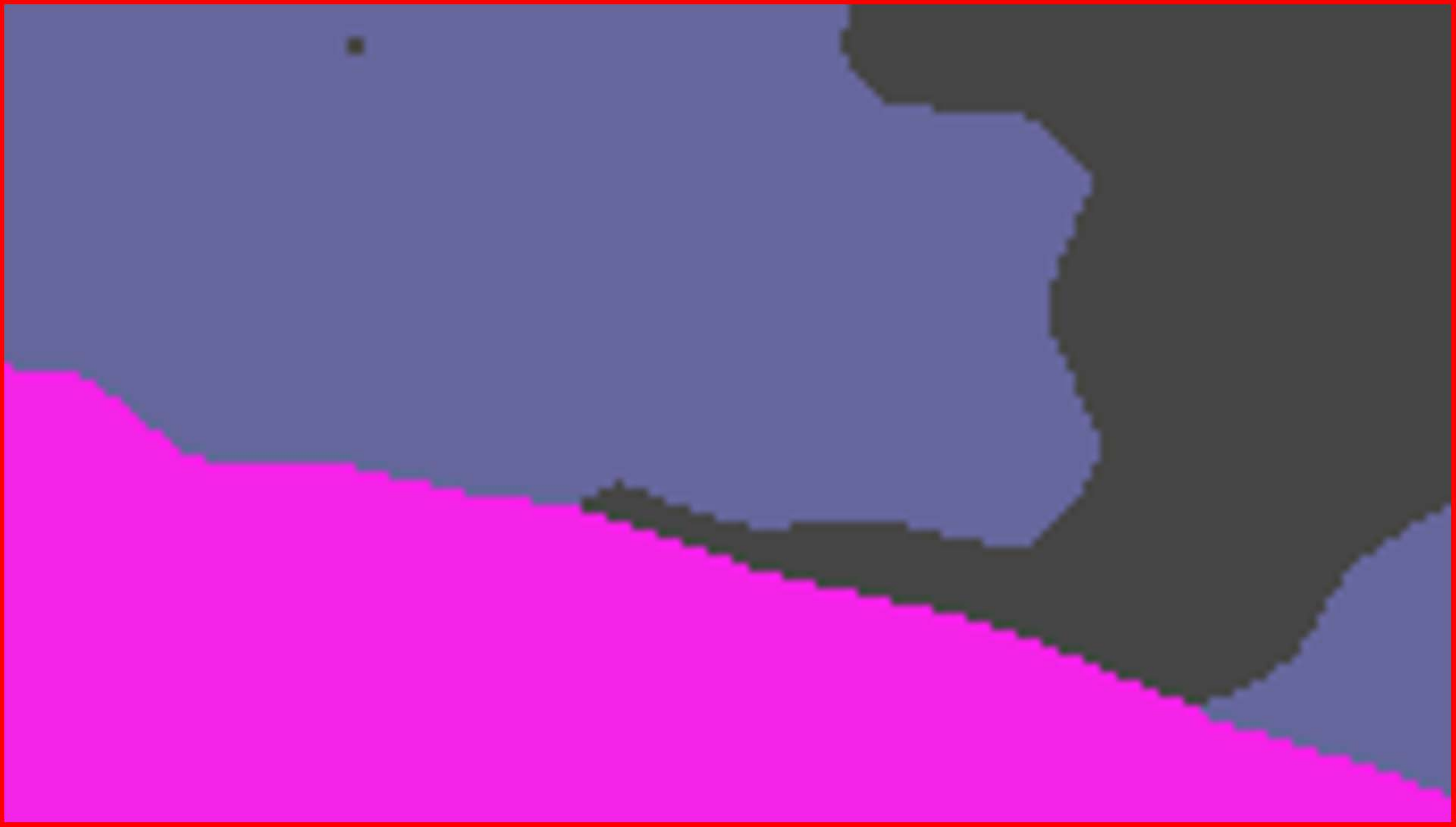}&
		\includegraphics[width=0.116\textwidth]{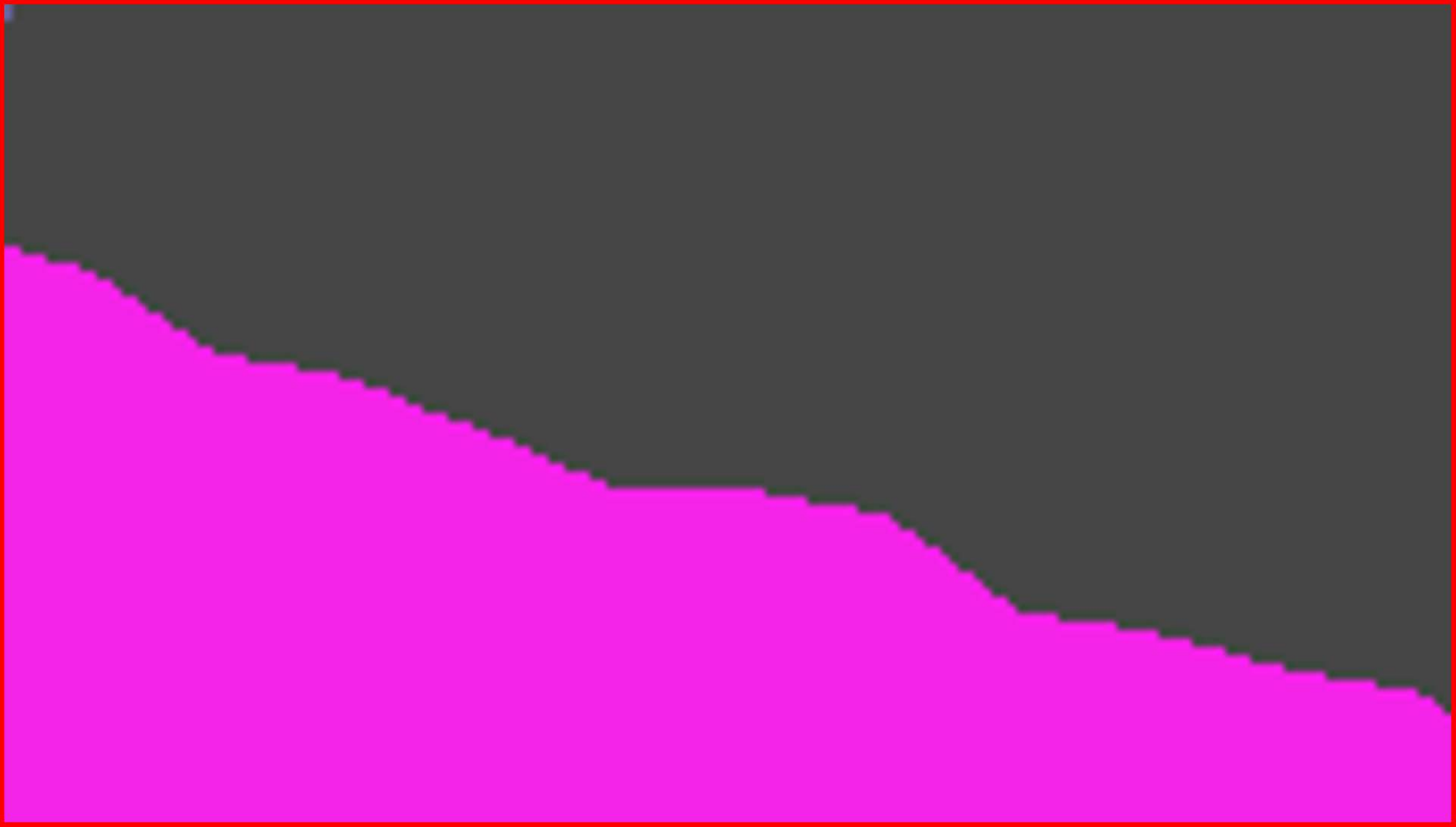}&
		\includegraphics[width=0.116\textwidth]{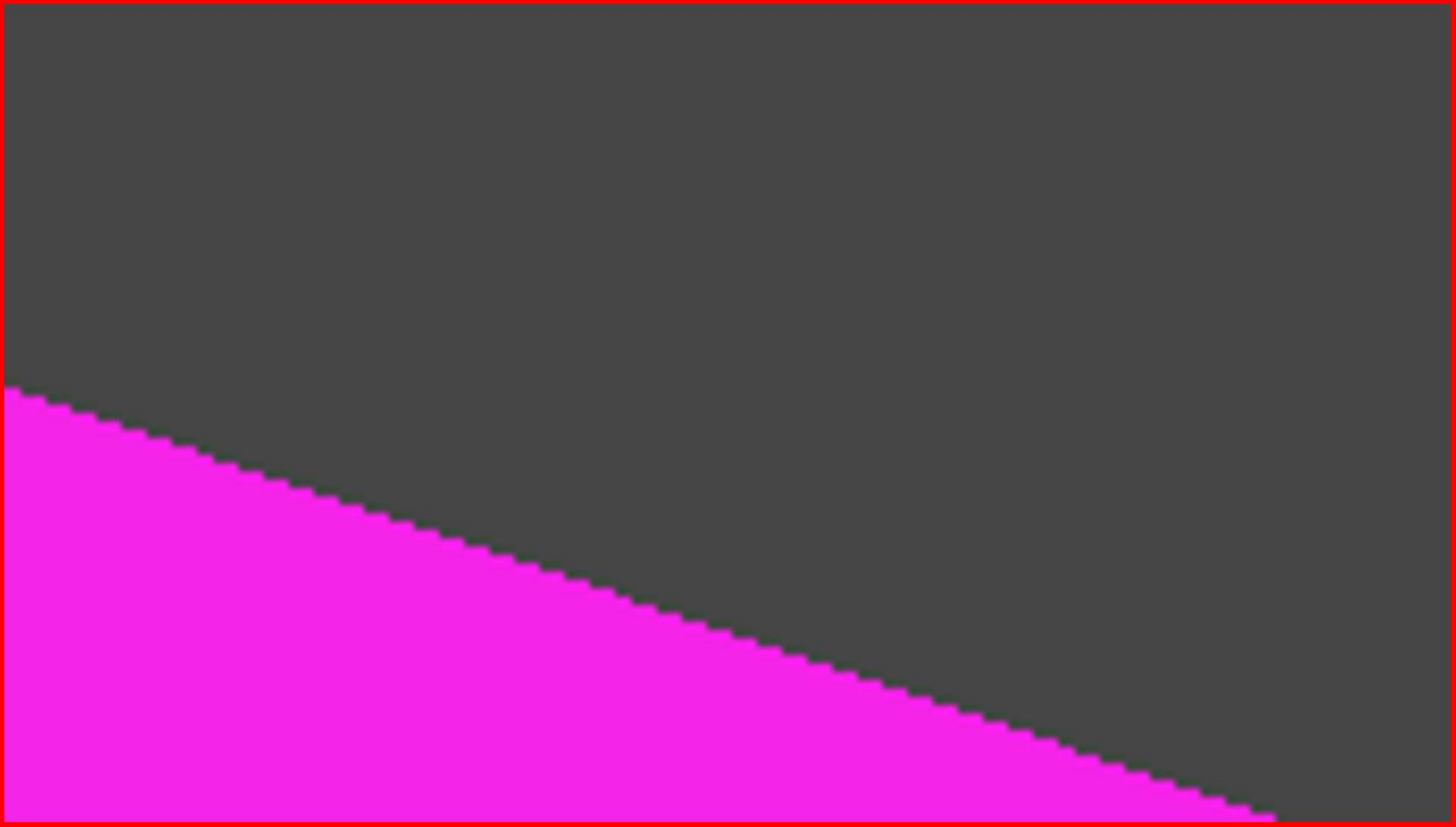}\\
		\footnotesize Input &\footnotesize DRBN&\footnotesize KinD &\footnotesize EnGAN &\footnotesize ZeroDCE &\footnotesize RUAS &\footnotesize SCI &\footnotesize GT \\
	\end{tabular}
	\vspace{-0.2cm}
	\caption{Visual results of semantic segmentation on the ACDC dataset. More results can be found in the Supplemental Materials. }
	\label{fig:segmentation}
	\vspace{-0.2cm}
\end{figure*}

\begin{table*}[t]
	\renewcommand\arraystretch{1.1} 
	\setlength{\tabcolsep}{1.8mm}
	\centering
	\begin{tabular}{|c|ccccccccccccccccc|c|}
		\hline 
		\footnotesize Method&\footnotesize RO&\footnotesize SI&\footnotesize BU&\footnotesize WA &\footnotesize FE &\footnotesize PO&\footnotesize TL&\footnotesize TS&\footnotesize VE&\footnotesize TE&\footnotesize SK&\footnotesize PE&\footnotesize RI&\footnotesize CA&\footnotesize TR&\footnotesize MO&\footnotesize BI&\footnotesize mIoU\\
		%			\hline 
		%			\footnotesize Method&\footnotesize road&\footnotesize sidewalk&\footnotesize building&\footnotesize wall &\footnotesize fence &\footnotesize pole&\footnotesize traffic light&\footnotesize traffic sign&\footnotesize vegetation&\footnotesize terrain&\footnotesize sky&\footnotesize person&\footnotesize rider&\footnotesize car&\footnotesize train&\footnotesize motorcycle&\footnotesize bicycle&\footnotesize mIoU\\
		\hline 
		\footnotesize RetinexNet&\footnotesize 90.6 &\footnotesize 67.2 &\footnotesize 74.3 &\footnotesize 35.5 &\footnotesize 37.4 &\footnotesize 41.0 &\footnotesize 36.4 &\footnotesize 31.9 &\footnotesize 67.3 &\footnotesize  13.0 &\footnotesize 79.4 &\footnotesize 30.7 &\footnotesize 4.1 &\footnotesize 0.3&\footnotesize \textcolor{red}{\textbf{65.3}} &\footnotesize 5.9 &\footnotesize 37.1 &\footnotesize 42.1 \\
		\hline 
		\footnotesize FIDE&\footnotesize 90.4 &\footnotesize 68.8 &\footnotesize \textcolor{blue}{\textbf{77.0}} &\footnotesize \textcolor{blue}{\textbf{40.1}} &\footnotesize 35.4 &\footnotesize  42.0 &\footnotesize 42.1 &\footnotesize 42.4 &\footnotesize 68.0 &\footnotesize 16.4 &\footnotesize 81.4 &\footnotesize \textcolor{blue}{\textbf{39.6}} &\footnotesize 8.7 &\footnotesize \textcolor{red}{\textbf{2.4}}  &\footnotesize 55.0 &\footnotesize 11.0 &\footnotesize 40.8 &\footnotesize 44.8 \\
		\hline
		\footnotesize DRBN&\footnotesize \textcolor{blue}{\textbf{91.8}} &\footnotesize \textcolor{blue}{\textbf{69.0}} &\footnotesize 76.9 &\footnotesize 39.5 &\footnotesize 38.3 &\footnotesize  \textcolor{red}{\textbf{43.3}} &\footnotesize 41.8 &\footnotesize \textcolor{blue}{\textbf{43.9}} &\footnotesize 68.1 &\footnotesize 16.5 &\footnotesize 80.1 &\footnotesize 36.6 &\footnotesize 7.6 &\footnotesize \textcolor{blue}{\textbf{1.7}}  &\footnotesize 64.2 &\footnotesize \textcolor{blue}{\textbf{12.1}} &\footnotesize 44.2 &\footnotesize \textcolor{blue}{\textbf{45.6}} \\
		\hline
		\footnotesize KinD&\footnotesize 89.5 &\footnotesize 66.7 &\footnotesize 75.3 &\footnotesize 35.6 &\footnotesize 37.5 &\footnotesize  \textcolor{blue}{\textbf{43.1}} &\footnotesize \textcolor{blue}{\textbf{46.3}} &\footnotesize 38.9 &\footnotesize 68.4 &\footnotesize 16.1 &\footnotesize 81.3 &\footnotesize \textcolor{red}{\textbf{40.4}} &\footnotesize \textcolor{red}{\textbf{10.5}} &\footnotesize 0.8  &\footnotesize 48.9 &\footnotesize 3.9 &\footnotesize 43.9 &\footnotesize 44.5 \\
		\hline
		\footnotesize EnGAN&\footnotesize 89.7 &\footnotesize 67.8 &\footnotesize  76.8 &\footnotesize 39.0 &\footnotesize \textcolor{blue}{\textbf{38.8}} &\footnotesize \textcolor{blue}{\textbf{43.1}} &\footnotesize  40.8 &\footnotesize  42.2&\footnotesize 68.8 &\footnotesize 18.4&\footnotesize 81.2&\footnotesize 39.4 &\footnotesize 8.0 &\footnotesize  0.3&\footnotesize 46.0 &\footnotesize  9.7 &\footnotesize 46.1 &\footnotesize  44.5\\
		\hline 
		\footnotesize SSIENet&\footnotesize 89.0 &\footnotesize  65.4 &\footnotesize 76.4 &\footnotesize 36.7 &\footnotesize 38.6 &\footnotesize 40.9 &\footnotesize 41.8 &\footnotesize 40.5 &\footnotesize \textcolor{blue}{\textbf{69.2}} &\footnotesize \textcolor{blue}{\textbf{20.2}} &\footnotesize 81.6 &\footnotesize 34.6 &\footnotesize 8.0 &\footnotesize 1.3&\footnotesize 46.5 &\footnotesize 10.4 &\footnotesize 39.9 &\footnotesize 43.6\\
		\hline 
		\footnotesize ZeroDCE&\footnotesize 90.1 &\footnotesize 67.2 &\footnotesize 77.3 &\footnotesize \textcolor{red}{\textbf{40.2}} &\footnotesize 37.8 &\footnotesize 41.9 &\footnotesize 42.2 &\footnotesize 41.9 &\footnotesize 69.1 &\footnotesize \textcolor{red}{\textbf{22.3}}&\footnotesize \textcolor{red}{\textbf{81.9}} &\footnotesize 36.3 &\footnotesize 7.0 &\footnotesize 0.3 &\footnotesize 54.3 &\footnotesize 11.8 &\footnotesize \textcolor{red}{\textbf{47.3}} &\footnotesize 45.2\\
		\hline
		\footnotesize RUAS&\footnotesize 90.4 &\footnotesize 66.7 &\footnotesize 76.2 &\footnotesize 37.6 &\footnotesize \textcolor{blue}{\textbf{38.8}} &\footnotesize 42.9 &\footnotesize 39.6 &\footnotesize 40.9 &\footnotesize 69.0 &\footnotesize 18.6&\footnotesize 81.5 &\footnotesize \textcolor{blue}{\textbf{39.6}} &\footnotesize \textcolor{blue}{\textbf{9.5}} &\footnotesize 0.6 &\footnotesize 49.6 &\footnotesize \textcolor{red}{\textbf{13.5}} &\footnotesize \textcolor{blue}{\textbf{46.6}} &\footnotesize 44.8\\
		\hline
		\footnotesize Ours&\footnotesize \textcolor{red}{\textbf{92.1}} &\footnotesize \textcolor{red}{\textbf{70.0}} &\footnotesize  \textcolor{red}{\textbf{78.3}} &\footnotesize 39.3  &\footnotesize \textcolor{red}{\textbf{39.8}}  &\footnotesize 43.0&\footnotesize \textcolor{red}{\textbf{46.6}} &\footnotesize \textcolor{red}{\textbf{44.2}} &\footnotesize \textcolor{red}{\textbf{69.9}} &\footnotesize 19.7&\footnotesize \textcolor{blue}{\textbf{81.8}} &\footnotesize 38.7  &\footnotesize 6.5 &\footnotesize 0.7   &\footnotesize \textcolor{blue}{\textbf{64.9}} &\footnotesize 9.0 &\footnotesize 42.4 &\footnotesize \textcolor{red}{\textbf{46.3}} \\
		\hline 
	\end{tabular}
	\vspace{-0.2cm}
	\caption{Quantitative results of nighttime semantic segmentation on the ACDC dataset. The symbol set \{RO, SI, BU, WA, FE, PO, TL, TS, VE, TE, SK, PE, RI, CA, TR, MO, BI\} represents \{road, sidewalk, building, wall, fence, pole, traffic light, traffic sign, vegetation, terrain, sky, person, rider, car, train, motorcycle, bicycle\}. Notice that we retrained the segmentation model on the enhanced results that were generated by all the compared methods. The best result is in {\textcolor{red}{\textbf{red}}} whereas the second best one is in {\textcolor{blue}{\textbf{blue}}}.}
	\label{tab: Segmentation}
	\vspace{-0.2cm}
\end{table*}

\subsection{Implementation Details}
\noindent\textbf{Parameter Settings.}$\;$In the training process, we used the ADAM optimizer~\cite{kingma2014adam} with parameters $\beta_1=0.9$, $\beta_2=0.999$, and $\epsilon=10^{-8}$. 
The minibatch size was set to 8. The learning rate was initialized to $10^{-4}$. The training epoch number was set to 1000.
We adopt 3 convolution + ReLU with 3 channels as our default setting for $\mathcal{H}_{\bm{\theta}}$ in our all experiments according to conclusion in Sec.~\ref{sec:flexibility}. 
Self-calibrated module contains four convolution layers, which ensures the lightweight of the training process. In fact, the form of the network may not be fixed, and we have done experiments to verify it in the Supplementary Materials.

\noindent\textbf{Compared Methods.}$\;$As for low-light image enhancement, we compared our SCI with four recently-proposed model-based methods (including LECARM~\cite{ren2018lecarm}, SDD~\cite{hao2020low}, STAR~\cite{xu2020star}), four advanced supervised learning methods (including RetinexNet~\cite{Chen2018Retinex}, KinD~\cite{zhang2019kindling}, FIDE~\cite{xu2020learning}, DRBN~\cite{yang2020fidelity}), and four unsupervised learning methods (including EnGAN~\cite{jiang2019EnGAN}, SSIENet~\cite{zhang2020self}, ZeroDCE~\cite{guo2020zero}, and RUAS~\cite{liu2021retinex}). As for dark face detection, except for performing the above-mentioned network-based enhancement works before the detector, we also compared the recently-proposed dark face detection method HLA~\cite{wang2021hla}. 

\noindent\textbf{Benchmarks Description and Metrics.}$\;$As for low-light image enhancement, we randomly sampled 100 images from MIT dataset~\cite{fivek} and 50 testing image from LSRW dataset~\cite{hai2021r2rnet} for testing. We used two full-reference metrics including PSNR and SSIM, five no-reference metrics including DE~\cite{shannon1948mathematical}, EME~\cite{agaian2007transform}, LOE~\cite{wang2013naturalness} and NIQE~\cite{wang2013naturalness}. As for dark face detection, we utilized the DARK FACE dataset~\cite{yang2020advancing} that consisted of 1000 challenging testing images that randomly sampled from the sub-challenge of UG2+ PRIZE CHALLENGE held at CVPR 2021. We considered the detection accuracy precision and recall rate as the evaluated metric. As for nighttime semantic segmentation, we utilized 400 images in ACDC~\cite{sakaridis2021acdc} for training and the remaining 106 images as the evaluated dataset. The evaluated metrics were defined as IoU and mIoU.

\subsection{Experimental Evaluation on Benchmarks}
\noindent\textbf{Performance Evaluation.} As shown in Table~\ref{table:LOLQuan}, our SCI achieved competitive performance, especially in no-reference metrics. 
As shown in Fig.~\ref{fig:MIT}-\ref{fig:LSRW}, advanced deep networks generated the unknown veils, leading to the inconspicuous details and unnatural colors. By comparison, our SCI achieved the best visual quality with vivid colors and prominent textures.
More visual comparisons can be found in the Supplemental Materials. 

\noindent\textbf{Computational Efficiency.} Further, we reported the model size, FLOPs, and running time (GPU-seconds) of some recently-proposed CNN-based methods in Table~\ref{table:parameters}. Obviously, our proposed SCI is the most lightweight compared with other networks, and significantly superior to others.

\subsection{In-the-Wild Experimental Evaluation}
Low-light image enhancement in  the wild scenarios is extremely challenging. The control of the partial overexposure information of the image, the correction of the overall color, and the preservation of image details are all problems that need to be solved urgently. Here, we tested lots of challenging in-the-wild examples from the DARK FACE~\cite{yang2020advancing} and ExDark~\cite{loh2019getting} datasets. As demonstrated in Fig.~\ref{fig:MoreVC}, through a large number of experiments, it can be seen that our method achieved more satisfactory visualization results than others, especially in the exposure level, structure depict, color presentation. Limited to the space, we provided more comparisons in the Supplemental Materials.

\subsection{Dark Face Detection}
We utilized the S3FD~\cite{zhang2017s3fd}, a well-known face detection algorithm to evaluate the dark face detection performance. Note that the S3FD was trained with the WIDER FACE dataset~\cite{yang2016wider} as presented in the original S3FD, and we used the pre-trained model of S3FD to fine-tune the images enhanced by various methods. 

At the same time, we performed a new method named SCI$^+$ which embed our SCI as a basic module into the front of S3FD for joint training over the combination of losses for task and enhancement. As reported in Fig.~\ref{fig:PRcurve}, our methods (SCI and SCI$^+$) realized the best scores among all compared method, and the reinforced version acquired the better performance than the fine-tune version. Fig.~\ref{fig:DarkFace} further demonstrated the visual comparison. It can be easily observed that with applying our SCI, the smaller objects can also be detected, while other methods failed to do so, as shown in the zoomed-in regions.

\subsection{Nighttime Semantic Segmentation}
Here we adopted the PSPNet~\cite{zhao2017pyramid} as the baseline to evaluate the segmentation performance on the pattern ``pre-train + fine-tune'' (similar to the version of SCI in dark face detection) for all methods. 
Table~\ref{tab: Segmentation} and Fig.~\ref{fig:segmentation} demonstrated the results of quantitative and qualitative comparison among different methods. Our performance is significantly superior to other state-of-the-art methods. As shown in the zoomed-in regions in Fig.~\ref{fig:segmentation}, all compared methods produced some unknown artifacts to damage the quality of the generated segmentation map.

\begin{figure}[t]
	\centering
	\begin{tabular}{c@{\extracolsep{0.25em}}c@{\extracolsep{0.25em}}c@{\extracolsep{0.25em}}c} 
		\includegraphics[width=0.108\textwidth]{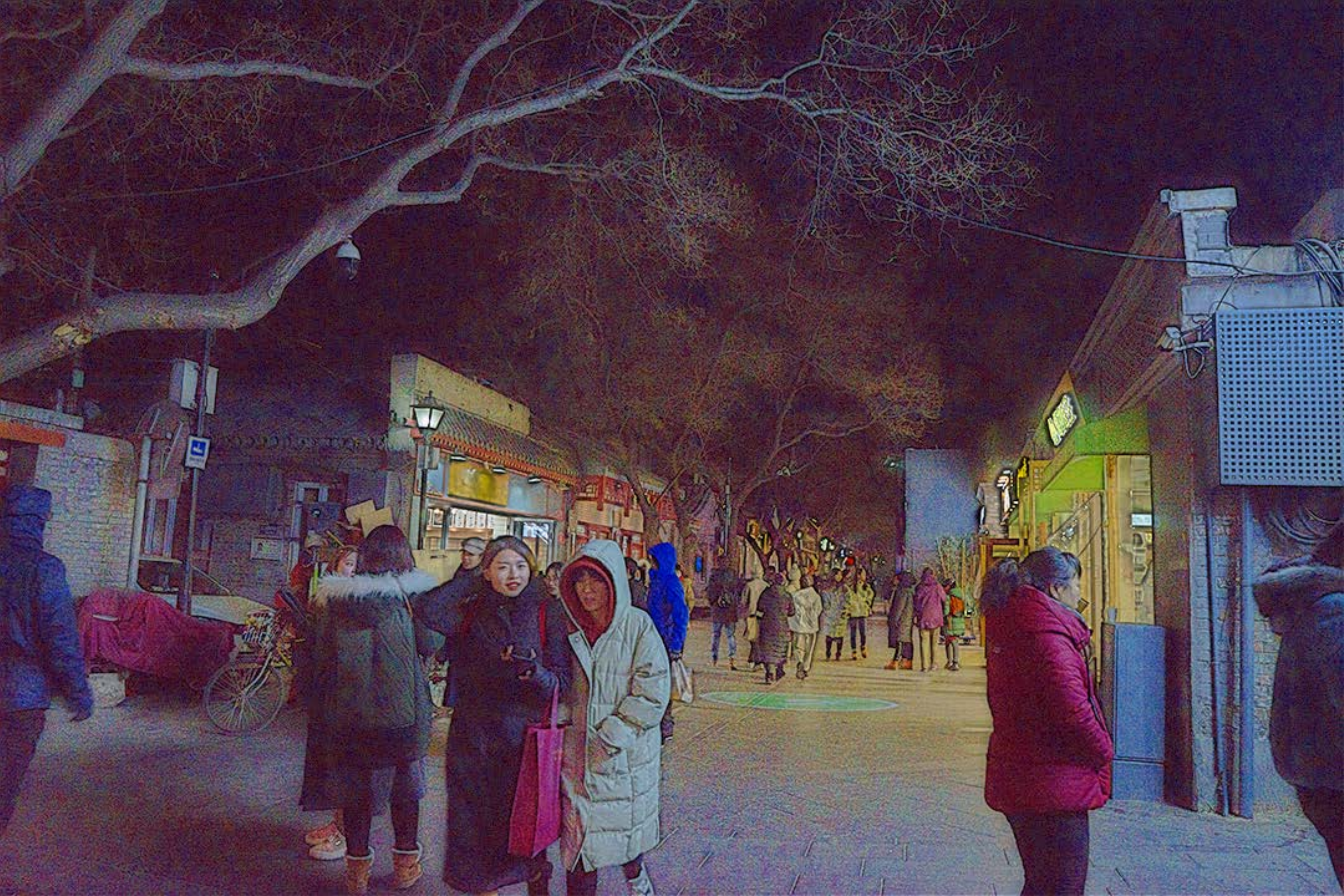}&
		\includegraphics[width=0.108\textwidth]{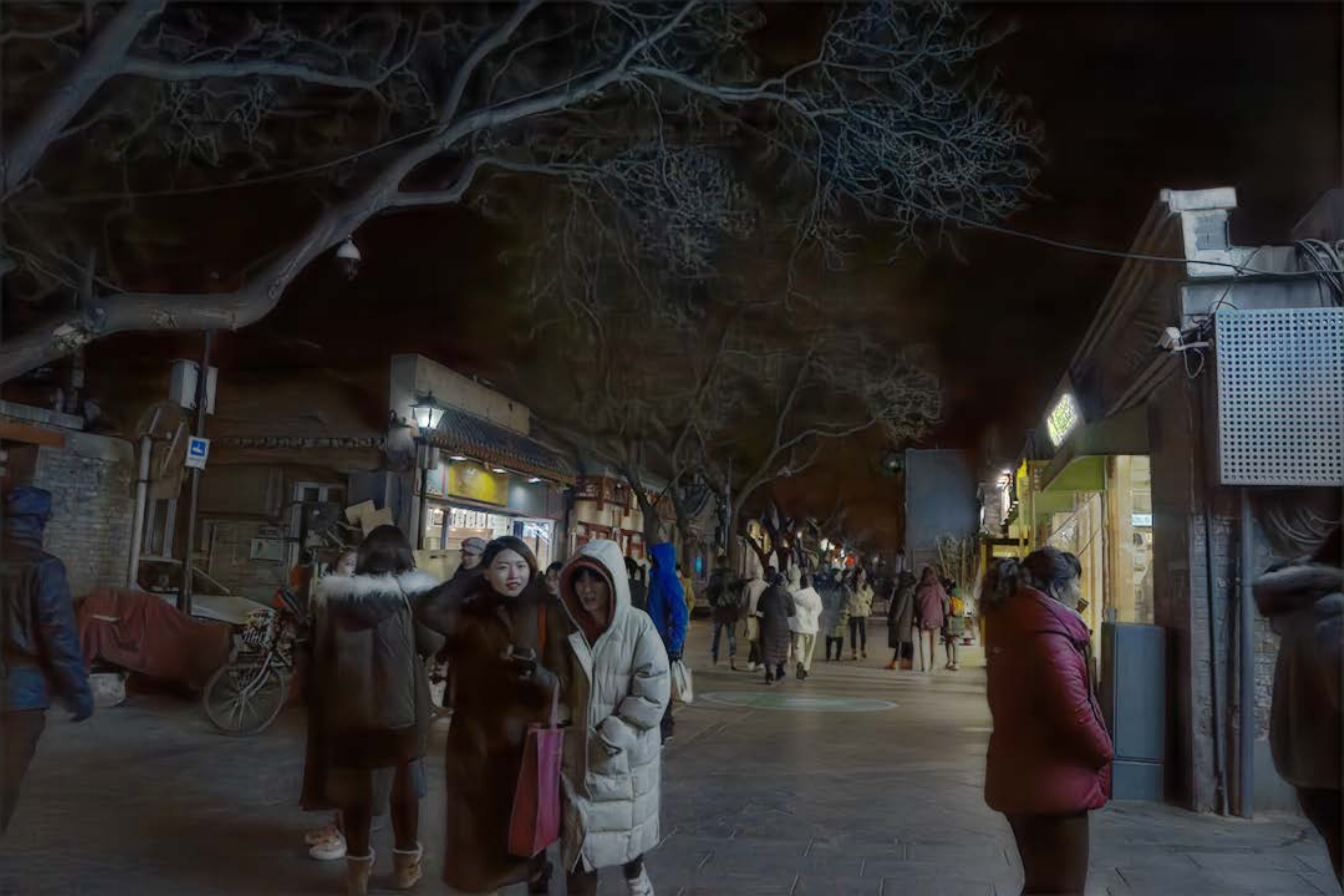}&
		\includegraphics[width=0.108\textwidth]{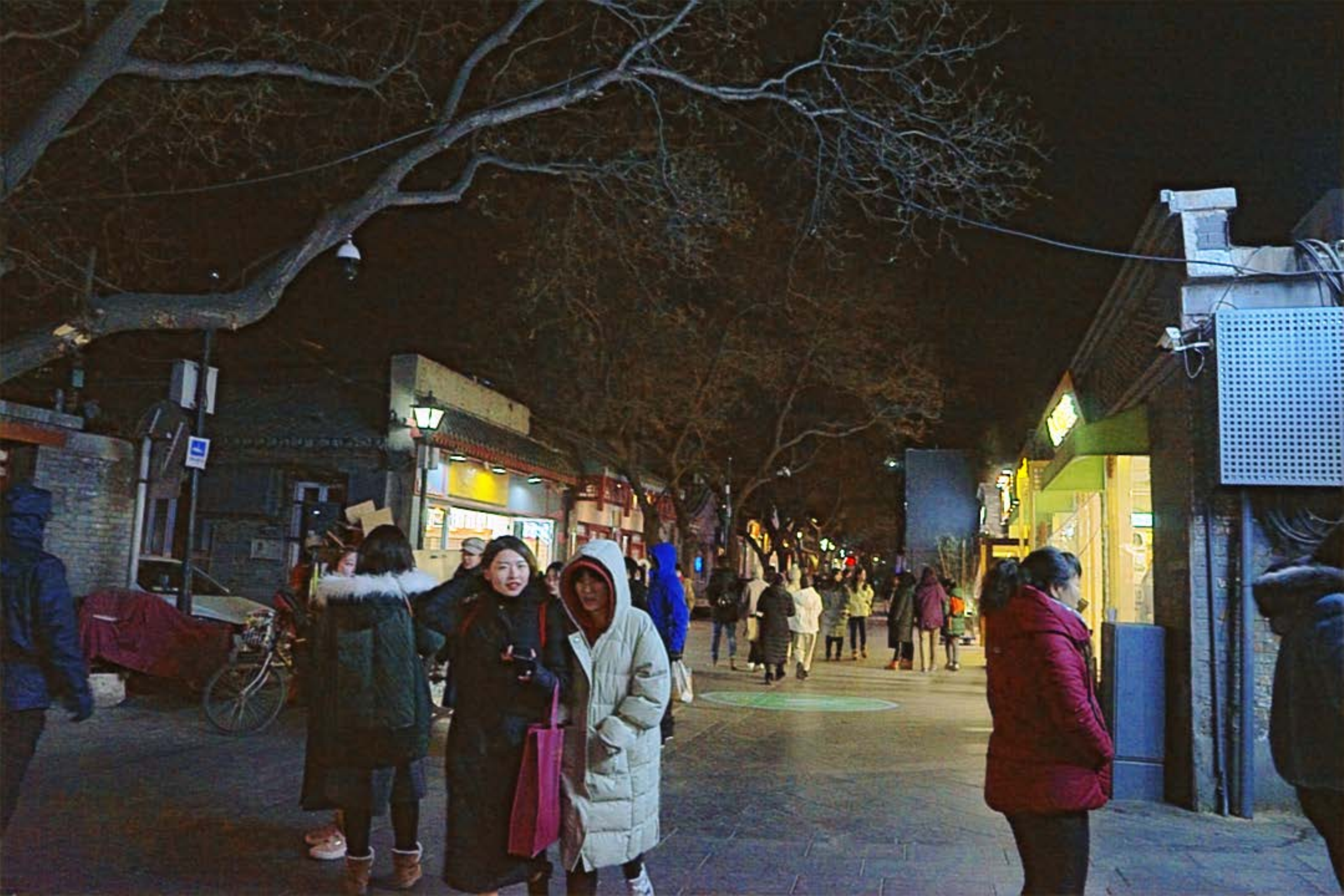}&
		\includegraphics[width=0.108\textwidth]{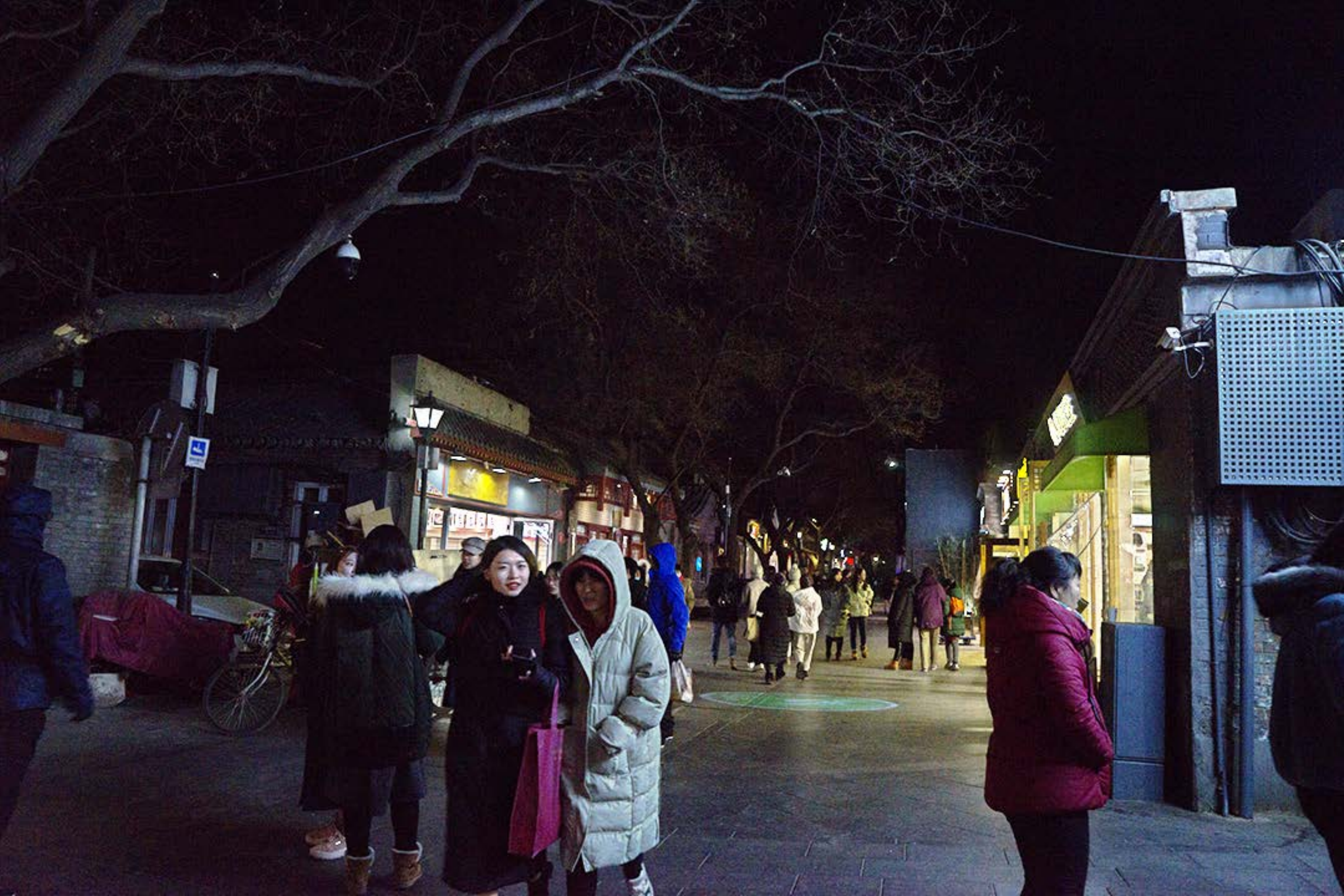}\\
		\includegraphics[width=0.108\textwidth]{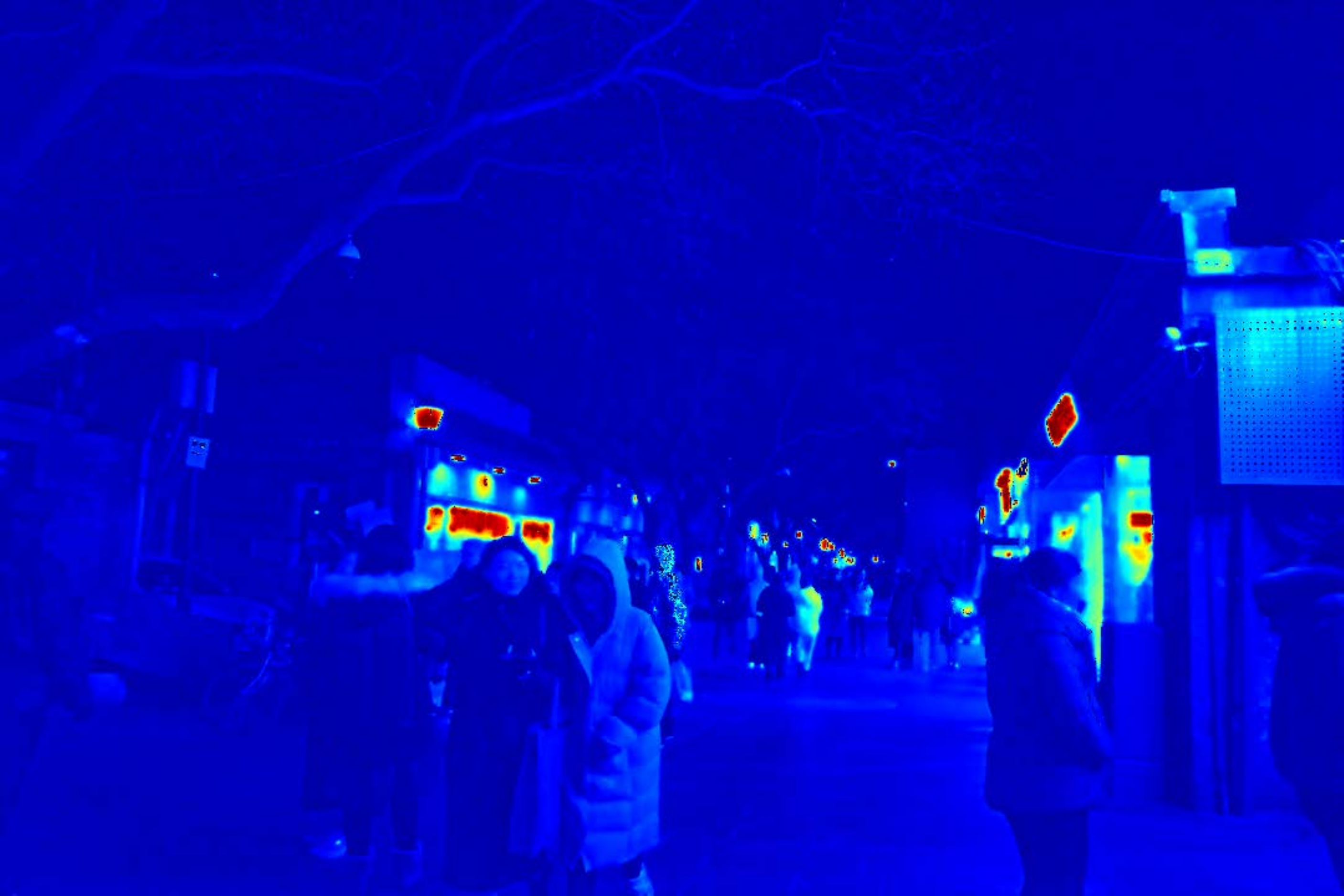}&
		\includegraphics[width=0.108\textwidth]{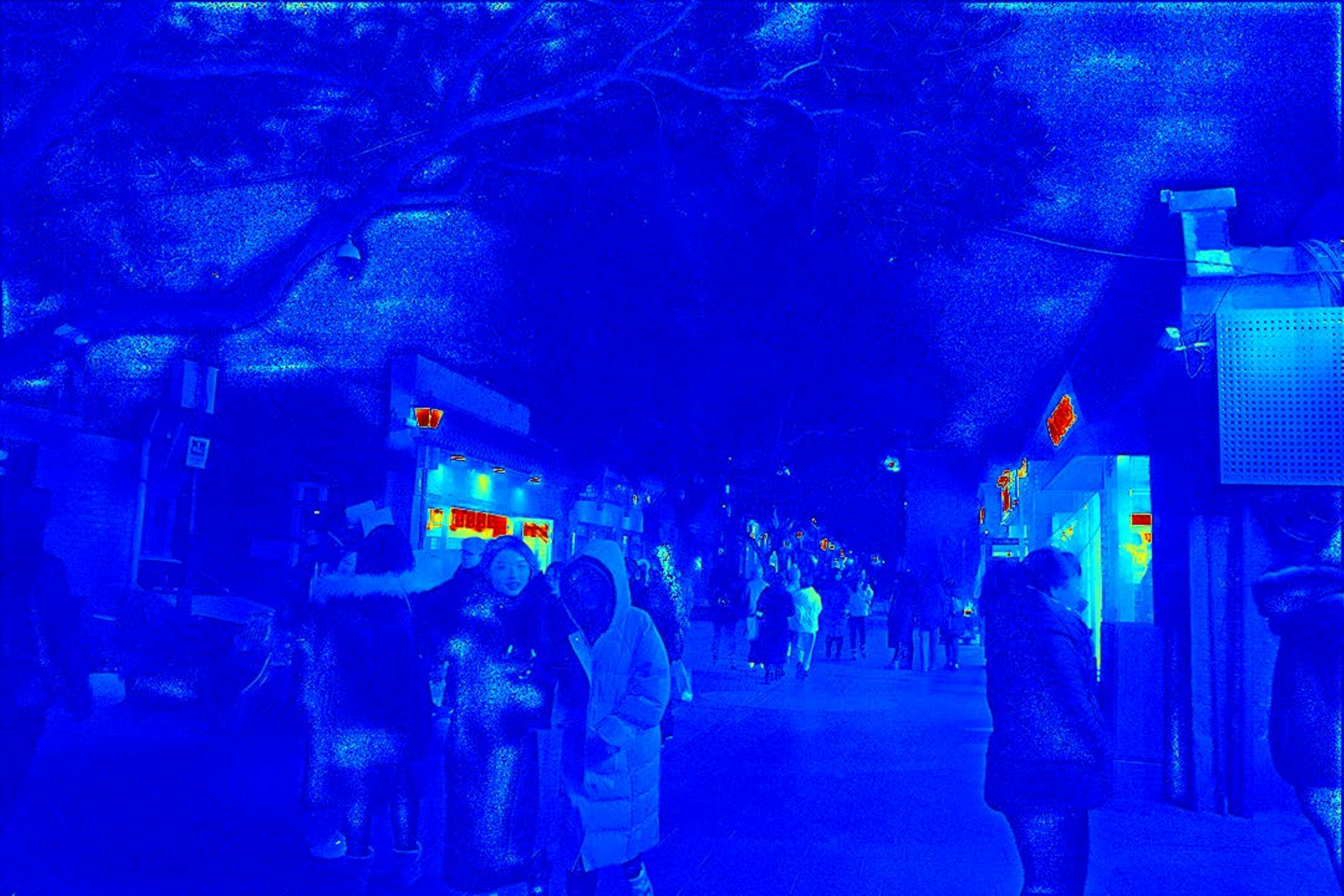}&
		\includegraphics[width=0.108\textwidth]{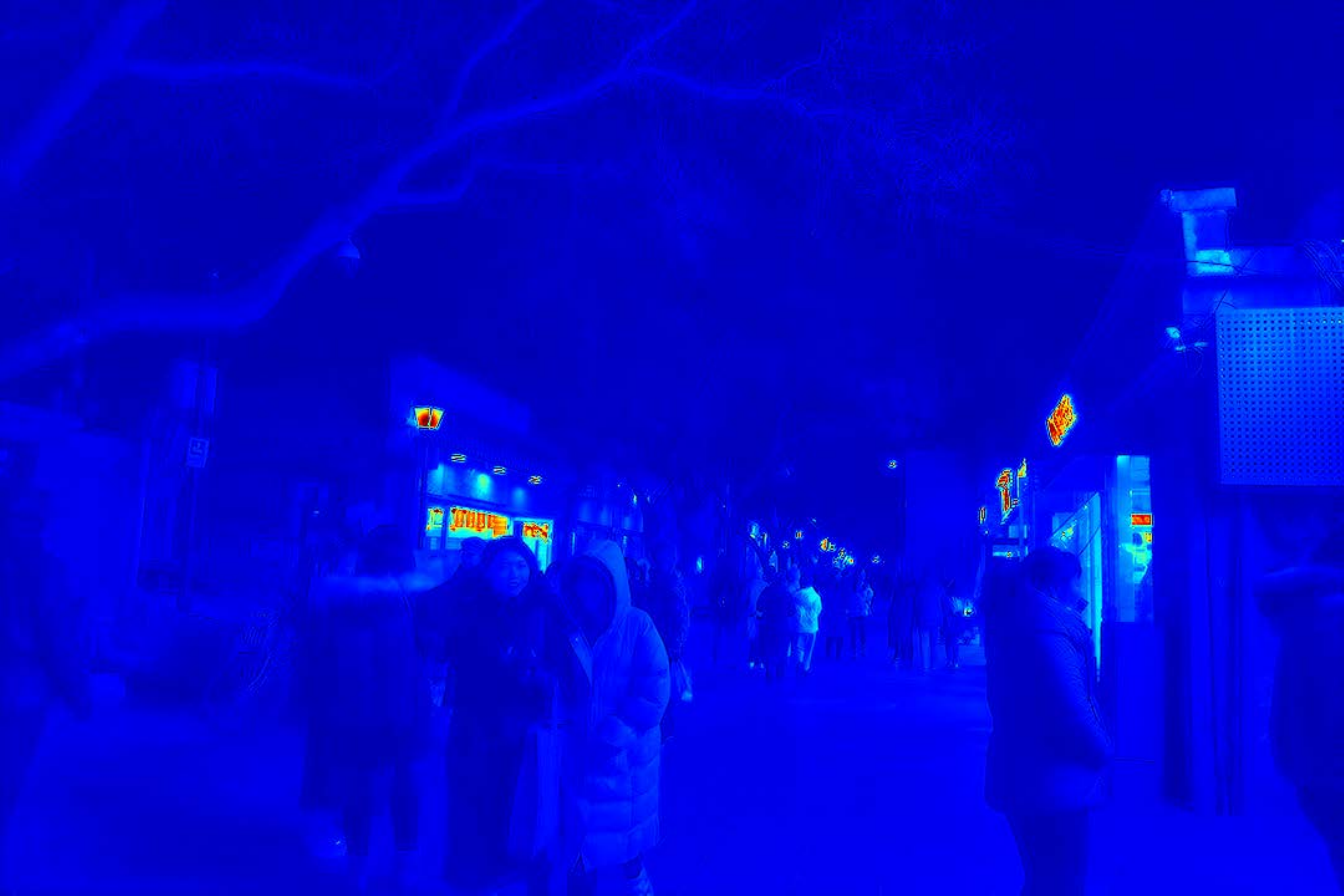}&
		\includegraphics[width=0.108\textwidth]{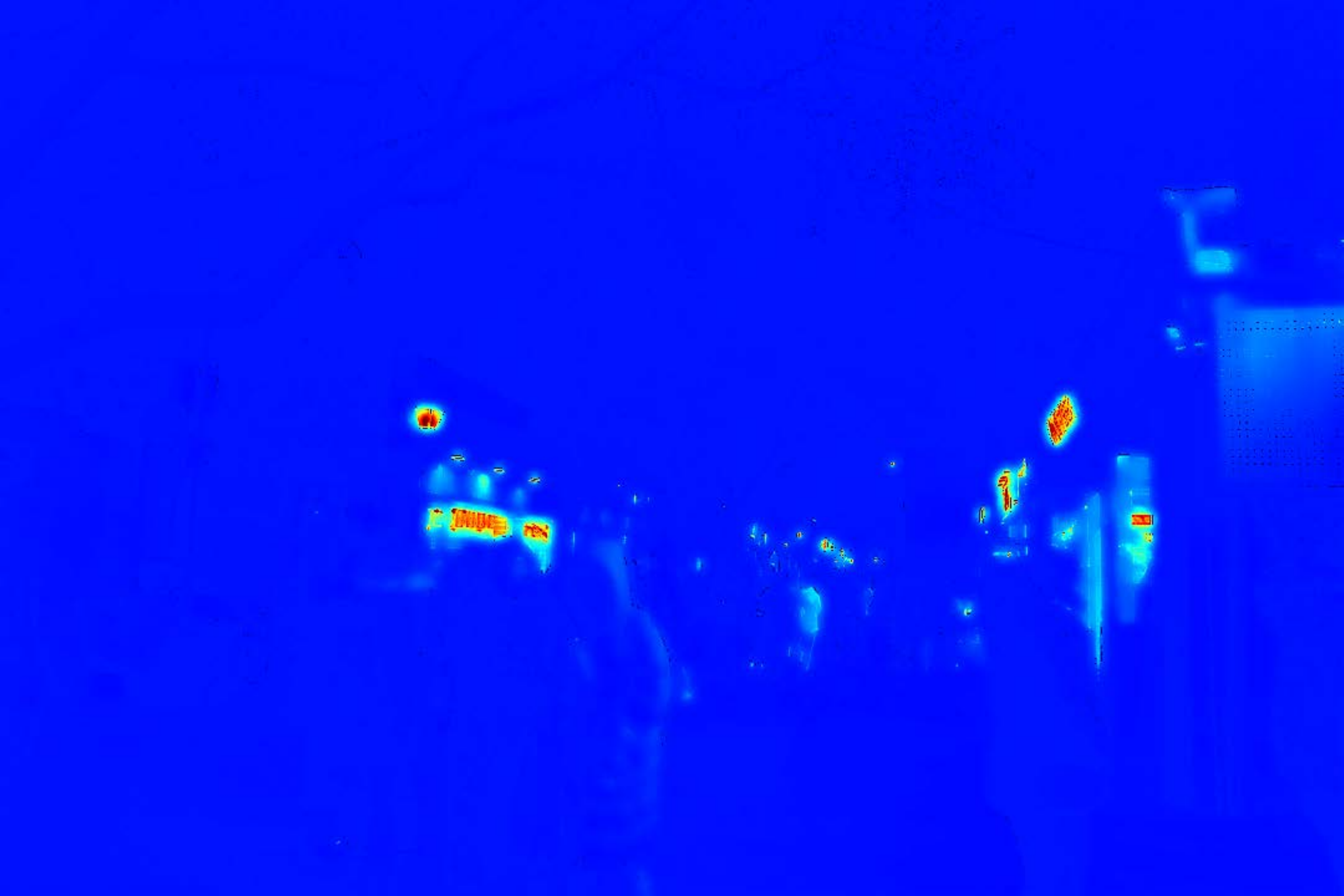}\\
		\footnotesize  RetinexNet &\footnotesize KinD &\footnotesize  SSIENet &\footnotesize  Ours\\
	\end{tabular}
	\vspace{-0.3cm}
	\caption{Comparing the decomposed components among different illumination-based networks. \textit{Zoom in for best view.}}
	\label{fig:Illumination}
	\vspace{-0.2cm}
\end{figure}

\subsection{Algorithmic Analyses}\label{sec:ablation}
\noindent\textbf{Comparing Decomposed Components.} Actually, our SCI belongs to illumination-based learning methods, the enhanced visual quality heavily depends on the estimated illumination. Here we compared our SCI with three representative illumination-based learning approaches, including RetinexNet, KinD, and SSIENet. As demonstrated in Fig.~\ref{fig:Illumination}, we can easily see that our estimated illumination kept an excellent smoothness property. It ensured our generated reflectance more visually friendly. 

\begin{figure}[t]
	\centering
	\begin{tabular}{c@{\extracolsep{0.25em}}c@{\extracolsep{0.25em}}c@{\extracolsep{0.25em}}c} 
		\includegraphics[width=0.108\textwidth]{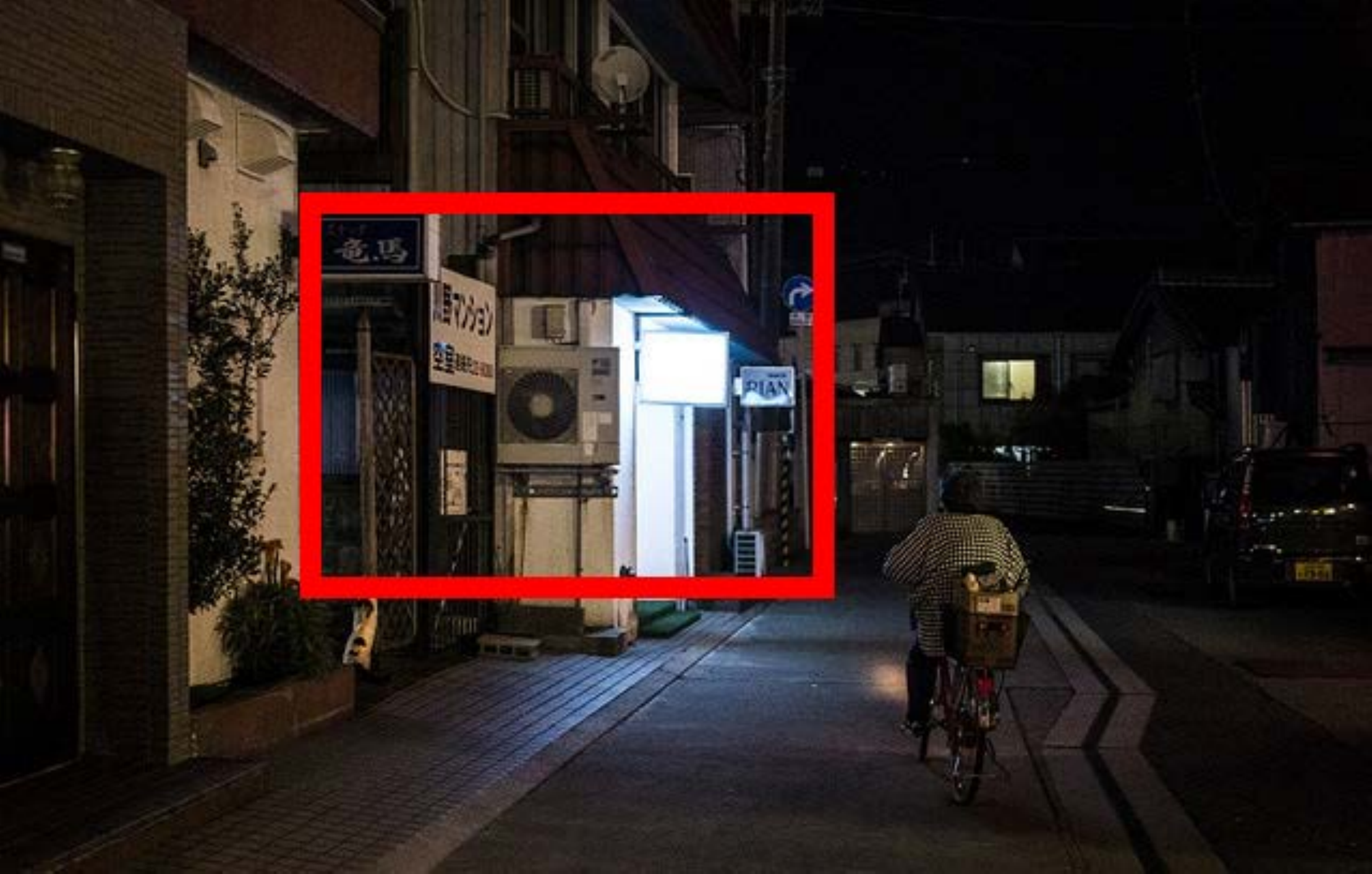}&
		\includegraphics[width=0.108\textwidth]{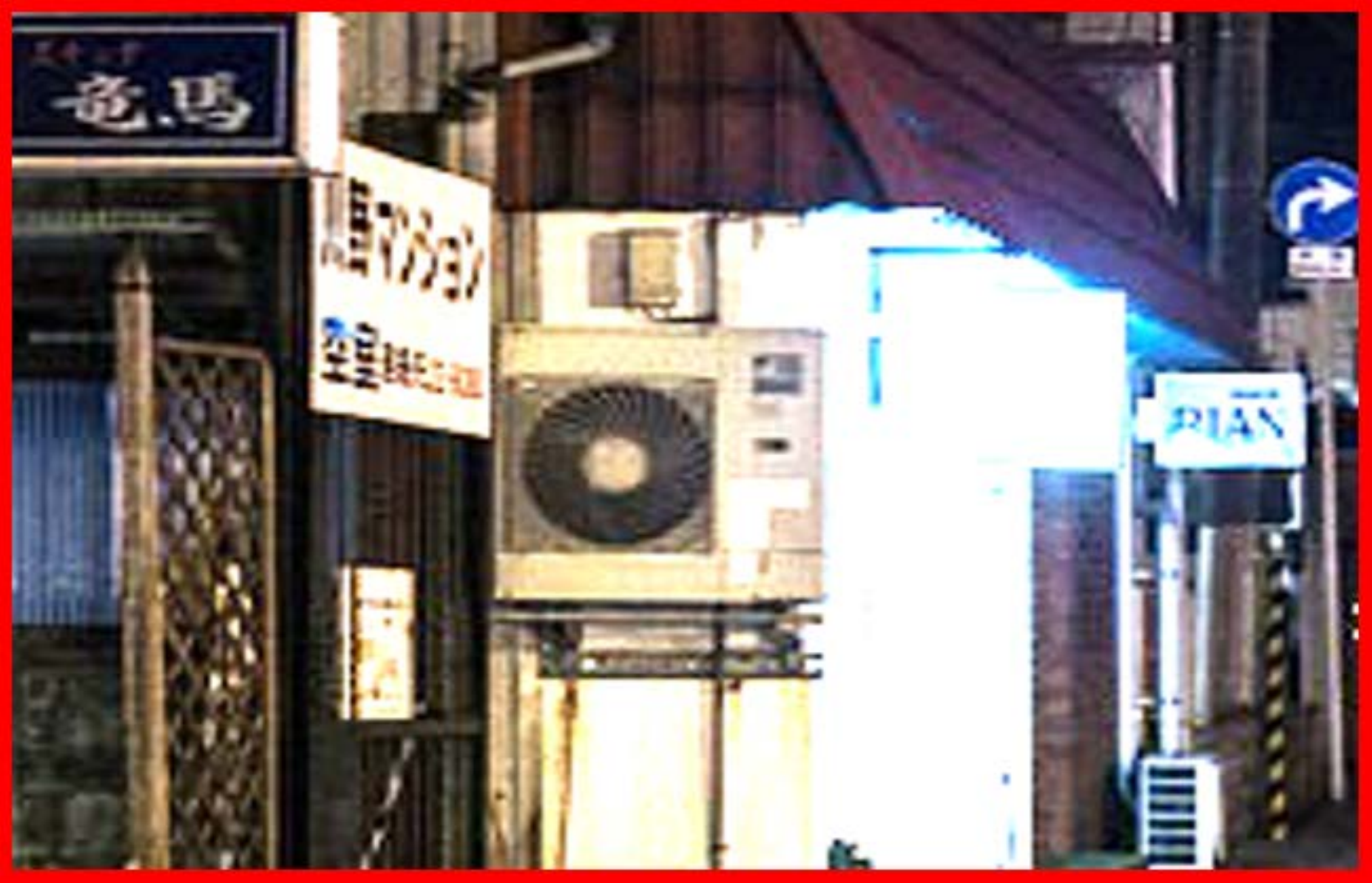}&
		\includegraphics[width=0.108\textwidth]{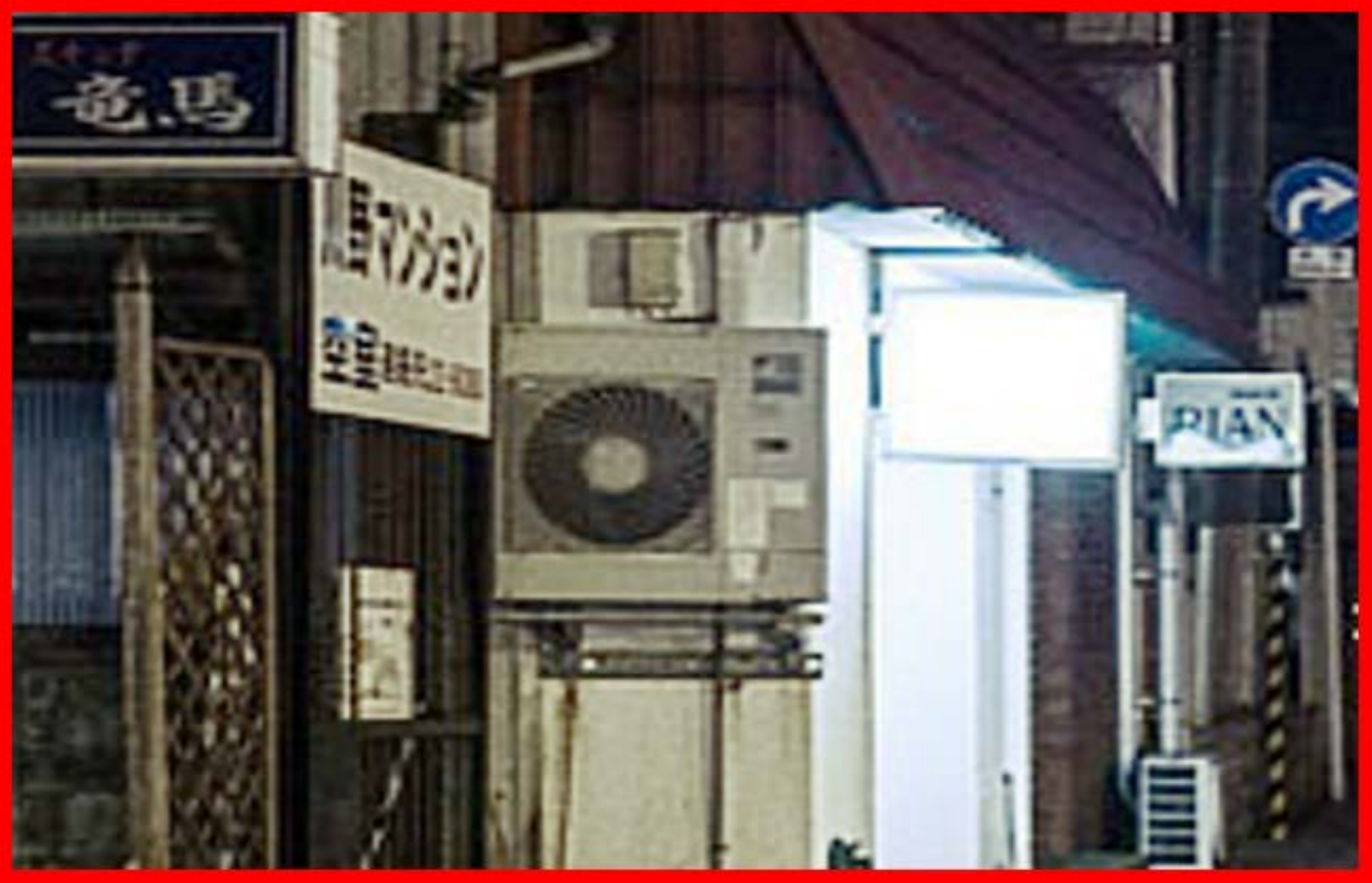}&
		\includegraphics[width=0.108\textwidth]{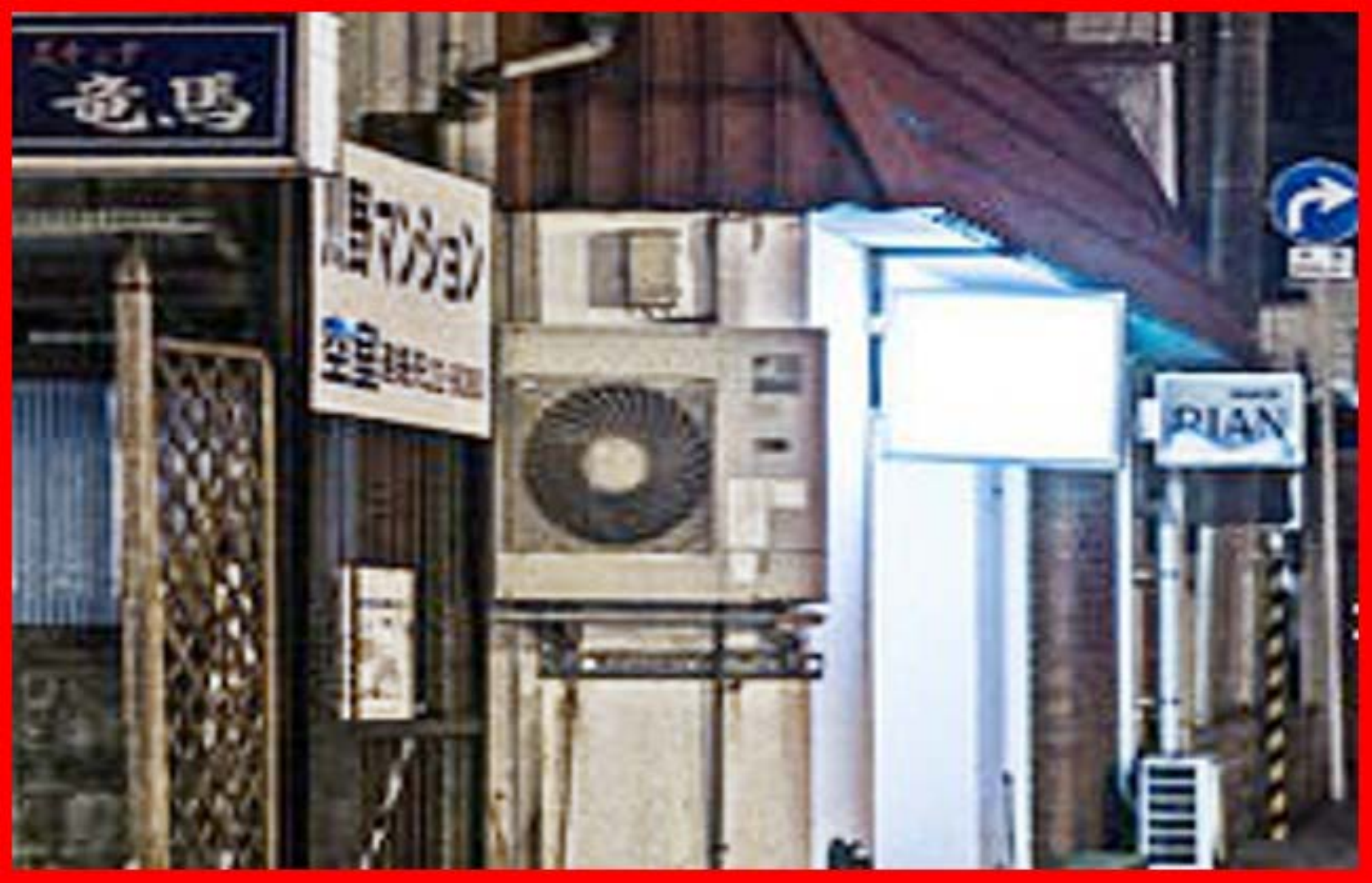}\\
		\footnotesize  Input &\footnotesize w/o residual &\footnotesize  w/ residual &\footnotesize  Ours\\
	\end{tabular}
	\vspace{-0.3cm}
	\caption{Analyze different modes in the illumination learning. }
	\label{fig:ablation}
	\vspace{-0.3cm}
\end{figure}

\noindent\textbf{Ablation Study.}
We compared the performance of different modes in Fig.~\ref{fig:ablation}. Learning the illumination directly will cause the image to be overexposed. The process of learning residuals between the illumination and the input indeed suppressed the overexposure, but the overall image quality is still not high, especially for the grasp of details. By comparison, the enhanced results using our method not only suppress the overexposure but also enrich image structures.

\section{Concluding Remarks}
In this paper, we successfully established a lightweight yet effective framework, Self-Calibrated Illumination (SCI) for low-light image enhancement toward different real-world scenarios. We not only made a thorough exploration to take on the excellent properties of SCI, but also we performed extensive experiments to indicate our effectiveness and superiority in low-light image enhancement, dark face detection, and nighttime semantic segmentation. 

\textbf{Broader Impacts.}
From the task's perspective, SCI provides an efficient and effective learning framework and has received extremely superior performance in both image quality and inference speed. Maybe it will be a brace to enter a new high-speed and high-quality era for low-light image enhancement. 
As for the method design, SCI opens a new perspective (i.e., introducing the auxiliary process for boosting the model capability of the basic unit in the training phase) to improve the practicability toward real-world scenarios for other low-level vision problems. 
%%%%%%%%% REFERENCES

\vspace{0.1cm}
\noindent\textbf{Acknowledgements:} This work is supported by the National Key R\&D Program of China (2020YFB1313503), the National Natural Science Foundation of China (Nos. 61922019, 61733002 and 62027826), and the Fundamental Research Funds for the Central Universities.

{\small
\bibliographystyle{ieee_fullname}
\bibliography{egbib}
}

\end{document}